\documentclass{article}

\usepackage[numbers, sort&compress]{natbib}
\usepackage[preprint]{neurips_2026}

\usepackage[utf8]{inputenc}
\usepackage[T1]{fontenc}
\usepackage{hyperref}
\usepackage{url}
\usepackage{booktabs}
\usepackage{amsfonts}
\usepackage{nicefrac}
\usepackage{microtype}
\usepackage{xcolor}
\usepackage{amsmath}
\usepackage{graphicx}
\usepackage{wrapfig}
\usepackage{amssymb}
\usepackage{extarrows}
\usepackage{mathtools}
\usepackage{algorithm}
\usepackage{algorithmic}
\usepackage{tikz}
\usetikzlibrary{automata, positioning}

\definecolor{blackcherry}{HTML}{510D0A}
\definecolor{oldgold}{HTML}{A29F15}

\title{Spiking Sensory Networks}

\author{%
	Lleyton A.~Ariton \\
	Cerebral Systems Inc. (Cerybra.) \\
	\texttt{lleyton.ariton@cerybra.com} \\
}

\begin{document}
	
	\addtocontents{toc}{\protect\setcounter{tocdepth}{-1}}
	
	\maketitle
	
	\begin{abstract}
		Effective sensory reception is a critical requirement for an organism's survival, yet the computational mechanisms behind how peripheral sensory neurons (SNs) transduce external signals into spiking representations usable by downstream brain circuits remain to be fully understood. Expectedly, numerous theoretical frameworks have been proposed to model sensory processing, of which Bayesian approaches are among the most prominent. However, existing machine learning-based attempts to apply Bayesian theory to spiking neurons remain largely phenomenological, disconnected from empirical biology, and poorly validated on non-stationary temporal data. Conversely, on the neuroscientific side, models often rely on strong inductive experimental biases or explicit kinetic dynamics fit \emph{post hoc} to match existing spiking data. Here, we introduce a class of recurrent Poisson VAE models, termed Spiking Sensory Networks (SSNs), designed to bridge these two paradigms by reconciling variational inference with biologically inspired spike transduction for time-varying stimuli. When evaluated on real-world odor plume recordings from locust olfactory receptor neurons (ORNs), SSNs match or exceed the reconstruction fidelity of Spiking Neural Network (SNN) and GLM baselines while exhibiting superior representational characteristics. Furthermore, we show that SSNs are equally well-suited for neuroscientific investigation as they can learn empirical \textit{in vivo} ORN response motifs and exhibit emergent spike-frequency adaptation under unsupervised reconstruction without manual parameter tuning. Moreover, we theoretically link the recurrent decoding objective directly to frameworks of mutual information maximization, and derive a formal thermodynamic lower bound connecting the Poisson Kullback--Leibler regularizer to the excess non-equilibrium metabolic power of biophysical homeostasis. Ultimately, our work simultaneously synthesizes principles from traditional machine learning theory and empirical neuroscience, thereby providing a viable bridge that concretizes Bayesian models with biological mechanisms.
	\end{abstract}
	
	\section{Introduction}
	
	Alongside learning and inference, sensory reception is arguably the most important mediator of an organism's survival within an environment. Evidenced by this, primary peripheral sensory neurons (SNs) were part of the first neuronal-type class to be developed by evolution \cite{arendtElementaryNervousSystems2021, arendtOriginEvolutionCell2016}, and are likewise amongst the earliest to be differentiated during the first neurogenic wave in the mammalian embryo \cite{marmigereSpecificationConnectivityNeuronal2007}. Consequently, understanding the computational principles of this ancient and fundamental processing step, from a theoretical standpoint, may garner deep insights into the neural code or cross over to improve methodologies for artificial, embodied agents.
	
	Interestingly, many early-stage sensory perception architectures across and within species exhibit analogous structure, even when taking into account different modalities \cite{shepherdEvolutionaryMicrocircuitApproach2021, byrneOxfordHandbookInvertebrate2019, gireTemporalProcessingOlfactory2013, kauppOlfactorySignallingVertebrates2010, strausfeldOlfactorySystemsCommon1999, hildebrandMechanismsOlfactoryDiscrimination1997}. For instance, in mammals, the olfactory and visual pathways are strikingly similar in construction, albeit with certain customizations pertaining to the corresponding sense \cite{gireTemporalProcessingOlfactory2013, shepherdEvolutionaryMicrocircuitApproach2021}. Fundamentally, then, various theories have regarded the overarching problem of sensory transduction to be one of mutual information maximization subject to metabolic constraints \cite{linskerSelforganizationPerceptualNetwork1988, laughlinMetabolicCostNeural1998, stemmlerInformationMaximizationSingle1998, hasenstaubMetabolicCostUnifying2010, wangEfficientNeuralCodes, chalkUnifiedTheoryEfficient2018}. Hence, the primary consideration of nature for evolving an SN is to optimize this objective---for which statistical learning theory stands to provide an excellent framing for the analysis of such systems. Indeed, this approach has an extremely rich and successful history dating back over a century \cite{helmholtzHandbuchPhysiologischenOptik1867, fechnerElementePsychophysik1860}. In particular, the ``Bayesian brain'' is an exceptionally popular theory of perceptual inference that formalizes cognitive processing through probabilistic modelling \cite{knillBayesianBrainRole2004, doyaBayesianBrainProbabilistic2011, fristonFreeenergyPrincipleUnified2010, aitchisonSamplingbasedProbabilisticInference2018}, though many such models skew towards a phenomenological side; in contrast, there is a large space of so-called spiking neural networks (SNNs) which offer a closer grounding in biology, though they are, for the most part, entirely deterministic and do not necessarily capture the priors established by empirical data. Moreover, standard SNNs are not inherently generative models, as the Bayesian brain or alternative predictive coding hypotheses would require. At the opposite end, neuroscience fits generalized linear models (GLMs) to reproduce observed SN spiking rates \cite{pillowSpatiotemporalCorrelationsVisual2008, caponeSpontaneousActivityEmerging2018}, but this is unsuitable for much else, nor do these models stem from a prescriptive learning theory; rather, a dual model is needed: one that captures key canonical properties of peripheral SNs, accounts for the inherent stochasticity of real-world neurons, adheres to biology, and is interpretable, all while simultaneously being capable of mapping onto current theories. Additionally, strong performance on time-varying signal processing tasks is desirable so that the model can extend beyond neuroscientific investigation into practical machine learning applications. Inspired by this, recent work has mapped the VAE formulation onto neuroscientifically inspired sensory perception, demonstrating biological alignment with learned sparsity and receptive fields for static natural images \cite{vafaiiPoissonVariationalAutoencoder2024, vafaiiBrainlikeVariationalInference2025, bellIndependentComponentsNatural1997}. However, these Poisson VAE ($\mathcal{P}$-VAE) approaches have yet to consider truly continuous, time-varying stimuli as encountered by peripheral receptors in the physical world.
	
	\paragraph{Contributions.} In this work, we propose a recurrent, biologically motivated VAE architecture suited for temporal signal processing and spike transduction: first, we show how the recently introduced $\mathcal{P}$-VAE can be extended through a recurrent VAE framework while still retaining a close biomimetic capacity; second, we evaluate the model on synthetic and naturalistic chemosensory tasks, demonstrating competitive reconstruction and superior population coding properties compared to existing spiking baselines; third, and most markedly, the model is used to draw information-theoretic and thermodynamic formalizations for the empirical behaviours commonly found in receptor neurons. Collectively, we believe that Spiking Sensory Networks (SSNs) have promise in providing a principled foundation for modelling time-varying sensory transduction in peripheral receptor neurons. Specifically, we highlight the following:
	\begin{itemize}
		\item The SSN architecture natively leverages distinct, learnable excitatory and inhibitory spatio-temporal filter banks alongside a leaky integration backbone and refractory history kernel. Consequently, when validated on encoding naturalistic odor plumes in a strictly causal manner akin to an organism navigating an environment, SSNs match or exceed the reconstruction performance of SNNs while producing higher-dimensional, decorrelated population codes.
		\item SSNs serve as a natural extension to both Poisson GLMs and the spatial $\mathcal{P}$-VAE, possessing stronger non-linear encoding capabilities than the former while extending the latter to continuous temporal signal modelling; additionally, by joining Poisson reparametrization with leaky integrator dynamics and a homeostatic prior, SSNs trained solely on unsupervised stimulus reconstruction organically learn spike-frequency adaptation profiles without explicit adaptive variables.
		\item Because SSNs essentially scaffold variational inference over biologically interpretable components, they provide a tractable abstraction for real-world sensory neurons; as such, this allows empirical receptor behaviours to be analyzed through statistical learning theory. Indeed, we link the recurrent ELBO objective to evolutionary mutual information maximization, and derive a formal lower bound via stochastic thermodynamics that connects the Poisson Kullback--Leibler regularizer to the excess non-equilibrium metabolic cost of cellular homeostasis.
	\end{itemize}
	
	\section{Background and Preliminaries}
	
	Taken abstractly, many sensory conversion problems can be distilled as some non-linear model receiving input from a set of spatio-temporal filters that selectively integrates a stimulus and converts it into spike-space; it is also required, though, depending on the sense, species, or stage of lifecycle, that the filters themselves be adaptable \cite{fritzRapidTaskrelatedPlasticity2003, francoOlfactoryReceptorsNeural2024, hirataniRapidBayesianLearning2020, suOlfactoryPerceptionReceptors2009}. Alternatively, we can view the filter learning process itself as an evolutionary force that eventually settles onto a set of fixed, optimal filters for a given task.
	
	Let $\mathbf{x} = (x_{1}, x_{2}, \cdots, x_{d}) \in \mathbb{R}^{d}$ be a vector of natural stimuli; since the world varies with time, the components of $\mathbf{x}$ are themselves inherently time-varying, so we take $\mathbf{x}_{t} = (x_{1, t}, x_{2, t}, \cdots, x_{d, t})$. The underlying representation of each of the entries of $\mathbf{x}_{t}$ can be almost anything; for instance, $x_{i, t}$ might be the concentration of the $i$th chemical compound to which the organism is sensitive, the amplitude of the $i$th frequency band of light, or proportional to the mechanical stress of the $i$th sensory hair. Under the information maximization framework, the evolutionary goal for a sensory neuron or population (acting as an encoder parametrized by $\phi$) is to maximize the mutual information (MI) between its transduced representations $\mathbf{z} \in \mathbb{R}^{n}$ and the stimulus vector $\mathbf{x} \in \mathbb{R}^{d}$ subject to metabolic penalties. This can be formally expressed as:
	\begin{equation}\label{eq:evolutionaryLoss}
		\min_{\phi} \mathcal{L}(\phi) = \min_{\phi} \left\{ -I(\mathbf{x}; \mathbf{z}) + \mathbb{E}[\mathcal{C}(\mathbf{z})] \right\},
	\end{equation}
	where $\mathcal{C}(\mathbf{z})$ is a cost function representing the energy expenditure of producing or maintaining the latent state $\mathbf{z}$. We assume that, in the context of spikes, $\mathcal{C}(\mathbf{z})$ captures the required metabolic energy for the neuron (or population) to fire at a given rate $\boldsymbol{\lambda}$ \cite{laughlinMetabolicCostNeural1998, hasenstaubMetabolicCostUnifying2010, vafaiiMetabolicCostInformation2026, balduzziMetabolicCostOrganizing2013}.
	
	Since the organism is unaware of the system governing the generation of $\mathbf{x}$, it views a given $\mathbf{x}_{t}$ as a sample of some process with underlying distribution $p^{*}(\mathbf{x}_{1}, \mathbf{x}_{2}, \cdots, \mathbf{x}_{t})$. To this end, we can map the problem of eq.~\eqref{eq:evolutionaryLoss} directly onto that solved by a variational autoencoder (VAE).
	
	\paragraph{VAE objective.} In standard VAE theory, one seeks to learn a generative model of data $\mathbf{x}$ via some latents $\mathbf{z}$ such that 
	\begin{equation}\label{eq:generativeModel}
		p_{\theta}(\mathbf{x}) = \int p_{\theta}(\mathbf{x}, \mathbf{z})\,\mathrm{d}\mathbf{z} = \int p_{\theta}(\mathbf{x} \mid \mathbf{z})p(\mathbf{z})\,\mathrm{d}\mathbf{z}.
	\end{equation}
	It is assumed that $\mathbf{x}$ carries the true underlying distribution $p^{*}(\mathbf{x})$, so the goal of the model is to minimize the discrepancy, i.e., the Kullback--Leibler (KL) divergence, between these two distributions. The latent variables themselves are endowed with a prior $p(\mathbf{z})$ and sampled from the approximate posterior distribution $q_{\phi}(\mathbf{z} \mid \mathbf{x})$, parametrized by $\phi$, which serves as an \emph{encoder}. The original data is then reconstructed with a \emph{decoder} by sampling from the likelihood $p_{\theta}(\mathbf{x} \mid \mathbf{z})$; subsequent derivations find that optimizing $\theta$ and $\phi$ for this reconstruction can be performed by maximizing the so-called evidence lower bound, or ELBO \cite{kingmaAutoEncodingVariationalBayes2022, rezendeStochasticBackpropagationApproximate}, defined as:
	\begin{equation}\label{eq:evidenceLowerBound}
		\mathcal{L}_{\text{ELBO}}(\theta, \phi; \mathbf{x}) = \mathbb{E}_{q_{\phi}(\mathbf{z} \mid \mathbf{x})}\!\left[\ln p_{\theta}(\mathbf{x} \mid \mathbf{z})\right] - D_{\text{KL}}\!\left(q_{\phi}(\mathbf{z} \mid \mathbf{x}) \,\|\, p(\mathbf{z})\right).
	\end{equation}
	The term \emph{variational} comes from the variational Bayes approach used in deriving eq.~\eqref{eq:evidenceLowerBound}. When considered for real-valued data with normal errors, the total loss function becomes:
	\begin{equation}
		\mathcal{L}_{\text{VAE}}(\theta, \phi; \mathbf{x}) = \|\mathbf{x} - \hat{\mathbf{x}}\|^{2}_{2} + \beta_{\text{KL}} D_{\text{KL}}\!\left(q_{\phi}(\mathbf{z} \mid \mathbf{x}) \,\|\, p(\mathbf{z})\right),
	\end{equation}
	where $\hat{\mathbf{x}}$ denotes the reconstruction produced by the decoder from latent sample $\mathbf{z} \sim q_{\phi}(\mathbf{z} \mid \mathbf{x})$, replacing $\ln p_{\theta}(\mathbf{x} \mid \mathbf{z})$ with a squared error term, and $\beta_{\text{KL}}$ controls the degree of regularization performed by the KL term. Usually, $\beta_{\text{KL}}$ is warmed-up throughout training to avoid posterior collapse \cite{bowmanGeneratingSentencesContinuous2016, sonderbyHowTrainDeep2016}. Additionally, the choices of both $p_{\theta}$ and $q_{\phi}$ are theoretically free, having already been chosen to span a number of distributions \cite{parkVariationalLaplaceAutoencoders2022, davidsonHypersphericalVariationalAutoEncoders2022, kamataFullySpikingVariational2022, vafaiiPoissonVariationalAutoencoder2024, sablicaSphericalCauchyVariational2026}.
	
	\paragraph{Temporal extension and recurrent VAEs.} The original VAE formulation operates on static, independent input snapshots with no temporal structure. In contrast, multiple extensions endowing the VAE formalism with recurrent processing capabilities have been presented \cite{girinDynamicalVariationalAutoencoders2022}; notable examples include the 
	Variational Recurrent Neural Network (VRNN) \cite{chungRecurrentLatentVariable2016}, the 
	Variational Recurrent Auto-Encoder (VRAE) \cite{fabiusVariationalRecurrentAutoEncoders2015}, 
	and Dynamical VAEs (DVAE) \cite{girinDynamicalVariationalAutoencoders2022}. The Deep Kalman 
	Filter \cite{krishnanDeepKalmanFilters2015} is also a member of this family.
	
	A recurrent VAE operates by considering a time-dependent variable, $\mathbf{u}_{t}$, hereafter referred to as the \emph{integration backbone}, which is responsible for conditioning the stochastic latent $\mathbf{z}_{t}$ for that given moment in time. When taken with the now also time-dependent input sequence $\mathbf{x}_{1:T} = (\mathbf{x}_{1}, \dots, \mathbf{x}_{T})$, the generative modelling problem from eq.~\eqref{eq:generativeModel} becomes:
	\begin{equation}\label{eq:recurrentGenerativeModel}
		p_{\theta}(\mathbf{x}_{1:T}) = \int p_{\theta}(\mathbf{x}_{1:T}, \mathbf{z}_{1:T})\,\mathrm{d}\mathbf{z}_{1:T} = \int \prod_{t = 1}^{T} p_{\theta}(\mathbf{x}_{t} \mid \mathbf{z}_{t}, \mathbf{u}_{t})\,p_{\theta}(\mathbf{z}_{t} \mid \mathbf{u}_t)\,\mathrm{d}\mathbf{z}_{1:T}.
	\end{equation}
	The choices for the integration backbone are highly diverse; commonly, they are standard recurrent neural networks (RNNs) such as the GRU \cite{choPropertiesNeuralMachine2014} or LSTM \cite{hochreiterLongShortTermMemory1997}, meaning that $\mathbf{u}_{t}$ admits the following formulation:
	\begin{equation}
		\mathbf{u}_{t} = f_{\theta}(\mathbf{u}_{t - 1}, \mathbf{x}_{t - 1}, \mathbf{z}_{t - 1}).
	\end{equation}
	If the integration backbone has no temporal dependencies, then the original VAE paradigm is recovered, wherein each time step is treated as an independent sample and the joint distribution captures the probability of observing the $\mathbf{x}_{1:T}$ values together. For a comprehensive review of recurrent and dynamical VAE architectures, see \cite{girinDynamicalVariationalAutoencoders2022}.
	
	\paragraph{Considering the evolutionary objective as a variational problem.} We connect the evolutionary objective eq.~\eqref{eq:evolutionaryLoss} to the recurrent VAE optimization problem eq.~\eqref{eq:recurrentGenerativeModel} through the Barber--Agakov variational lower bound \cite{barberIMAlgorithmVariational, brekelmansImprovingMutualInformation2023}. Expanding the trajectory MI between the stimulus $\mathbf{x}_{1:T} \in \mathbb{R}^{d \times T}$ and the latent spike sequence $\mathbf{z}_{1:T}$, we have:
	\begin{align}\label{eq:trajectoryMutualInformation}
		I(\mathbf{x}_{1:T}; \mathbf{z}_{1:T}) &= h(\mathbf{x}_{1:T}) + \mathbb{E}_{p(\mathbf{x}_{1:T}, \mathbf{z}_{1:T})}\!\left[ \ln p(\mathbf{x}_{1:T} \mid \mathbf{z}_{1:T}) \right] \nonumber \\
		&\ge h(\mathbf{x}_{1:T}) + \mathbb{E}_{p^{*}(\mathbf{x}_{1:T}) q_{\phi}(\mathbf{z}_{1:T} \mid \mathbf{x}_{1:T})}\!\left[ \ln p_{\theta}(\mathbf{x}_{1:T} \mid \mathbf{z}_{1:T}) \right].
	\end{align}
	Assuming an isotropic Gaussian decoding likelihood $p_{\theta}(\mathbf{x}_{1:T} \mid \mathbf{z}_{1:T}) = \mathcal{N}\left(\hat{\mathbf{x}}_\theta(\mathbf{z}_{1:T}), \sigma^{2} \mathbf{I}_{dT}\right)$, the expected log-likelihood evaluates directly to:
	\begin{equation}
		\mathbb{E}_{p^{*}(\mathbf{x}_{1:T}) q_{\phi}(\mathbf{z}_{1:T} \mid \mathbf{x}_{1:T})}\!\left[\ln p_{\theta}(\mathbf{x}_{1:T} \mid \mathbf{z}_{1:T})\right] = -\frac{dT}{2} \ln\left(2 \pi \sigma^{2}\right) - \frac{dT}{2 \sigma^{2}} \operatorname{MSE}(\phi, \theta),
	\end{equation}
	where $\operatorname{MSE}(\phi, \theta) = \frac{1}{dT}\mathbb{E}_{p^{*}(\mathbf{x}_{1:T}) q_{\phi}(\mathbf{z}_{1:T} \mid \mathbf{x}_{1:T})}\!\left[\|\mathbf{x}_{1:T} - \hat{\mathbf{x}}_\theta(\mathbf{z}_{1:T})\|_{2}^{2}\right]$. Maximizing this bound analytically with respect to the decoding noise variance $\sigma^{2}$ yields:
	\begin{equation}\label{eq:variationalMSEBound}
		I(\mathbf{x}_{1:T}; \mathbf{z}_{1:T}) \ge h(\mathbf{x}_{1:T}) - \frac{dT}{2} \ln \left( 2 \pi e \operatorname{MSE}(\phi, \theta) \right).
	\end{equation}
	Clearly, because $-\ln\left(2\pi e\operatorname{MSE}(\phi, \theta)\right)$ is strictly monotonically decreasing with respect to $\operatorname{MSE}(\phi, \theta)$, improving reconstruction fidelity increases the MI lower bound \cite{prasadCertainRelationsMutual2010}. For a detailed exposition including all intermediate steps, refer to Appendix~\ref{section:stimulusReconstructionTheory}.
	
	\paragraph{Homeostatic prior and thermodynamic metabolic cost.} The static prior $p(\mathbf{z}_t) = \operatorname{Poisson}(\mathbf{z}_{t}; \boldsymbol{\lambda}_{\text{ref}})$ is set to reflect the spontaneous baseline firing observed ubiquitously in biological receptor neurons even in the absence of stimuli ($\sim$3--6~Hz in insect ORNs \cite{kimOlfactoryReceptorNeurons2023} and $\sim$0.5--3~Hz in mammals \cite{duchamp-viretOdorResponseProperties1999, reisertOriginBasalActivity2010}). Since the KL-divergence subsequently penalizes firing rates away from the reference, it acts, and can be intuitively interpreted, as a regularization term mimicking homeostatic metabolic cost \cite{vafaiiPoissonVariationalAutoencoder2024, vafaiiMetabolicCostInformation2026}. Beyond intuition, the functional form of the Poisson KL-divergence (eq.~\eqref{eq:poissonKL}) admits a formal physical grounding in cellular non-equilibrium thermodynamics. In excitable cells, maintaining firing rates away from the resting homeostatic baseline, i.e., $\boldsymbol{\lambda}_{t} \neq \boldsymbol{\lambda}_{\text{ref}}$, expends chemical regulatory energy; by extension, we show (see Appendix~\ref{section:homeostasisTheory}) via a chemical master equation of calcium-mediated homeostasis that the excess non-equilibrium regulatory power $\dot{W}_{\text{ex}}$ required to sustain a driven firing rate against homeostatic relaxation is lower-bounded by the relative entropy between the driven and baseline Poisson distributions:
	\begin{multline}\label{eq:thermodynamicBound}
		\dot{W}_{\text{ex}}(t) \ge \frac{k_{B} T_{\text{temp}} \xi}{\Delta t} \, D_{\text{KL}}\!\left(q_{\phi}(\mathbf{z}_{t} \mid \mathbf{x}_{\le t}, \mathbf{z}_{<t}) \,\|\, p(\mathbf{z}_{t})\right) \propto \beta_{\text{KL}} D_{\text{KL}}\!\left(q_{\phi}(\mathbf{z}_{t} \mid \mathbf{x}_{\le t}, \mathbf{z}_{<t}) \,\|\, p(\mathbf{z}_{t})\right),
	\end{multline}
	where $k_{B}$ is the Boltzmann constant, $T_{\text{temp}}$ is absolute temperature, and $\xi$ is the biophysical coupling constant linking calcium influx to transcriptional turnover. Under this formulation, eq.~\eqref{eq:evidenceLowerBound} maximizes sensory decoding fidelity subject to the homeostatic chemical work required to sustain it.
	
	\section{Model Architecture}
	
	By design, the SSN has been architected to incorporate core functional abstractions of empirical sensory neuron processing. As such, it can be deconstructed into the following components:
	\begin{enumerate}
		\item an integration backbone, which is implemented via a standard leaky-integrate dynamic variable;
		\item a set of dual excitatory and inhibitory filter banks, $\mathbf{g}_{\text{exc}}$ and $\mathbf{g}_{\text{inh}}$, respectively, along with a spike-history filter $\mathbf{h}$; and
		\item a probabilistic spike generation mechanism with $\mathbf{z}_{t} \sim \operatorname{Poisson}(\boldsymbol{\lambda}_{t})$.
	\end{enumerate}
	
	\paragraph{Integration backbone.} Temporal information integration was chosen to follow typical leaky-integrator models. In many ways, leaky integration can be considered as a canonical property of all neurons, mostly stemming from its link to the direct biophysical makeup of the neuron cell itself \cite{gerstnerNeuronalDynamicsSingle2014, izhikevichDynamicalSystemsNeuroscience2006} (originating from the capacitive RC-circuit dynamics of the lipid membrane). Indeed, leaky-integrate type models have shown huge success in reproducing actual experimental data \cite{jolivetPredictingSpikeTiming2006, gerstnerNeuroscienceHowGood2009, mensiParameterExtractionClassification2012}, and are also the primary model of choice for nearly all spiking neural network (SNN) applications on account of their computational efficiency and tractability \cite{eshraghianTrainingSpikingNeural2023, zenkeSuperspikeSupervisedLearning2018, tavanaeiDeepLearningSpiking2019, neftciSurrogateGradientLearning2019}. The continuous-time dynamics of the membrane potential $\mathbf{u}(t)$ are governed by the following ODE:
	\begin{equation}\label{eq:integrationBackboneODE}
		\frac{\mathrm{d}\mathbf{u}(t)}{\mathrm{d}t} = -\boldsymbol{\tau}_{\text{leak}}^{-1} \odot \mathbf{u}(t) + \mathbf{g}_{\text{exc}}(t) - \mathbf{g}_{\text{inh}}(t) + \mathbf{u}_{\text{rest}},
	\end{equation}
	which is integrated numerically across discrete time steps via forward Euler updates to obtain $\mathbf{u}_{t} = \sigma(\mathbf{g}_{\text{leak}}) \odot \mathbf{u}_{t - 1} + \mathbf{g}_{\text{exc}, t} - \mathbf{g}_{\text{inh}, t} + \mathbf{u}_{\text{rest}}$, where $\sigma(\mathbf{g}_{\text{leak}}) \in (0, 1)^{n}$ parametrizes the discrete membrane retention factor $1 - \Delta t \, \boldsymbol{\tau}_{\text{leak}}^{-1}$ and $\mathbf{u}_{\text{rest}} \in \mathbb{R}^{n}$ denotes the resting membrane bias.
	
	\paragraph{Filter banks.} Since we cannot know \emph{a priori} the appropriate responses and receptive fields suited to every type of stimulus or environment, the model consists of three freely trainable filter banks: two for excitatory and inhibitory influences, represented by $\mathbf{g}_{\text{exc}, t} = \operatorname{ReLU}\!\left((\mathbf{k}_{\text{exc}} * \mathbf{x})_{t} + \mathbf{b}_{\text{exc}}\right)$ and $\mathbf{g}_{\text{inh}, t} = \operatorname{ReLU}\!\left((\mathbf{k}_{\text{inh}} * \mathbf{x})_{t} + \mathbf{b}_{\text{inh}}\right)$, respectively, as well as a third responsible for introducing influence from the neuron's own historical spike activity up to a predetermined length $L_{\text{hist}}$; combined with the membrane dynamics $\mathbf{u}_{t}$, this gives:
	\begin{equation}\label{eq:modelDynamics}
		\hat{\mathbf{u}}_{t} = \mathbf{u}_{t} + \tilde{\mathbf{h}}_{t} = \mathbf{u}_{t} + \sum_{k = 1}^{L_{\text{hist}}} \mathbf{h}_{k} \odot \mathbf{z}_{t - k},
	\end{equation}
	where $\mathbf{z}_{t - k}$ denotes past emitted spikes and $\mathbf{h}$ parametrizes the refractory self-coupling kernel. Taken together, this architecture can be viewed as a flavour of the Spike-Response Model with escape noise \cite{gerstnerNeuronalDynamicsSingle2014}, or equivalently, a recurrent Poisson GLM equipped with non-linear feedforward receptive fields and continuous Poisson reparametrization. Intuitively, the stochastic elements sit on top of usual models as a scaffold to then apply variational theory.
	
	\paragraph{Dynamic approximate posterior.} The variational approximate posterior $q_{\phi}(\mathbf{z}_{t} \mid \mathbf{x}_{\le t}, \mathbf{z}_{<t})$ dynamically adjusts at each time step according to a time-varying firing rate $\boldsymbol{\lambda}_{t} \in \mathbb{R}^{n}_{+}$, conditioned on the recurrent state variables. Specifically, the effective membrane potential $\hat{\mathbf{u}}_{t} = \mathbf{u}_{t} + \tilde{\mathbf{h}}_{t}$ is mapped to the Poisson rate parameter $\boldsymbol{\lambda}_{t}$ via a softplus link function:
	\begin{equation}\label{eq:poissonRate}
		\boldsymbol{\lambda}_{t} = \operatorname{softplus}(\boldsymbol{\alpha}) \odot \operatorname{softplus}\!\left( \frac{\hat{\mathbf{u}}_{t} - \operatorname{softplus}(\boldsymbol{\vartheta})}{\operatorname{softplus}(\boldsymbol{\beta})} \right) + \epsilon,
	\end{equation}
	where $\boldsymbol{\alpha}, \boldsymbol{\beta}, \boldsymbol{\vartheta} \in \mathbb{R}^{n}$ represent trainable rate-scaling, inverse-gain (scale), and firing threshold parameters, respectively, and $\epsilon > 0$ provides numerical stability. The latent spike variable $\mathbf{z}_{t}$ is then distributed according to:
	\begin{equation}
		q_{\phi}(\mathbf{z}_{t} \mid \mathbf{x}_{\le t}, \mathbf{z}_{<t}) = \operatorname{Poisson}(\mathbf{z}_{t}; \boldsymbol{\lambda}_{t}).
	\end{equation}
	Paired with a stationary Poisson prior $p(\mathbf{z}_{t}) = \operatorname{Poisson}(\mathbf{z}_{t}; \boldsymbol{\lambda}_{\text{ref}})$ with baseline reference rate $\boldsymbol{\lambda}_{\text{ref}} \in \mathbb{R}^{n}_{+}$, the stepwise KL-divergence admits a closed-form analytic expression:
	\begin{equation}\label{eq:poissonKL}
		D_{\text{KL}}\!\left(q_{\phi}(\mathbf{z}_{t} \mid \cdot) \,\|\, p(\mathbf{z}_{t})\right) = \sum_{i = 1}^{n} \left[ \lambda_{t, i} \ln\left(\frac{\lambda_{t, i}}{\lambda_{\text{ref}, i}}\right) + \lambda_{\text{ref}, i} - \lambda_{t, i} \right],
	\end{equation}
	which is accumulated across sequence length $T$ to form the total regularization penalty in the recurrent ELBO objective.
	
	\section{Results}
	
	To evaluate the SSN, we compare against three other models across two complementary collections of tasks: one gauges the fidelity and representational efficiency of time series reconstruction, while the other evaluates how well the SSN captures empirical neuroscientific dynamics. The comparison (baseline) models represent the most established paradigms for trainable spiking architectures, being SNNs based on the leaky-integrate-and-fire (LIF) neuron, of which one was outfitted with adaptation (ASNN / ALIF). We also opted to include comparisons against the classical Poisson GLM for its similarity in design and purpose. As a means to isolate extraneous influences and to enforce fairness on each task, all of the models tested were outfitted identically with the same filter bank front-end. The only exception to this was the Poisson GLM, which was chosen to retain its single, canonical filter because it was observed to provide better performance.
	
	\paragraph{Datasets.} We make use of \textit{in vivo} electrophysiological recordings from 43 locust ORNs responding to naturalistic odor plumes, spanning continuous 40~s exposures and discrete pulse bursts \cite{kimOlfactoryReceptorNeurons2023}, for dual purposes: once for temporal signal reconstruction, where networks encode the turbulent stimulus measured at the antenna (Figure~\ref{fig:plume-reconstruction}) to evaluate non-stationary plume encoding; and again for when models are trained to reproduce the firing activity of the 43 ORNs themselves. Additionally, we experiment on a synthetic sinusoidal mixture to gauge absolute baseline reconstruction performance and noise robustness under controlled conditions. We give a more in-depth overview of both the data and training protocols in Appendix~\ref{section:trainingDetails}.
	
	\subsection{Reconstruction properties}
	
\begin{figure}[htbp]
		\centering
		\includegraphics[width=\textwidth]{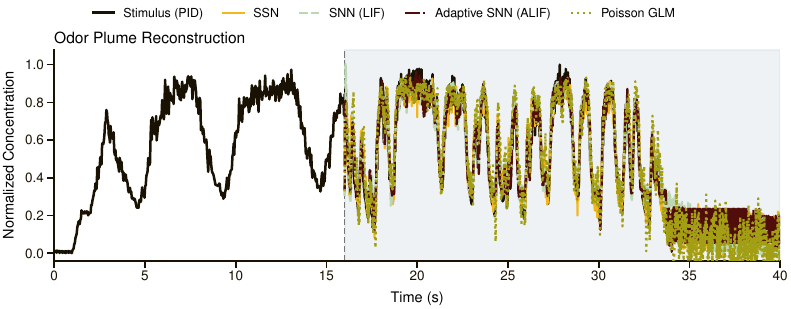}
		\caption{Naturalistic odor plume reconstruction results.}
		\label{fig:plume-reconstruction}
	\end{figure}

\begin{wrapfigure}[25]{r}{0.5\textwidth}
		\centering
		\vspace{-10pt}
		\includegraphics[width=0.5\textwidth]{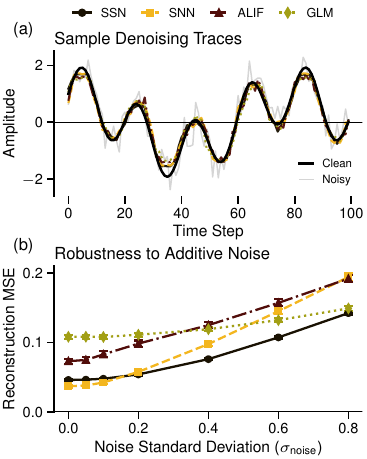}
		\caption{Noise robustness experiment on the synthetic wave task. \textbf{(a)} Example of the synthetic wave data. \textbf{(b)} Reconstruction performance when injected with increasing levels of noise.}
		\label{fig:noise-robustness}
	\end{wrapfigure}

	We considered two time series reconstruction tasks: the continuous encoding and decoding of a complex, natural odor plume recorded via a photoionization detector (PID) at the locust antenna \cite{kimOlfactoryReceptorNeurons2023}, and the reconstruction of a synthetic multi-frequency sinusoidal wave designed to benchmark baseline tracking fidelity and harmonic representation across disparate frequency bands (restricting the SSN to single-spike binary emissions per step for a direct comparison with binary LIF baselines). As summarized in Table~\ref{tab:statisticalComparison} and depicted in Figure~\ref{fig:plume-reconstruction} (with the synthetic wave illustrated in Figure~\ref{fig:noise-robustness}a), the SSN achieved high-fidelity signal reconstruction alongside superior population coding and statistics across both tasks.
	
	\begin{table}[t]
		\centering
		\caption{\textbf{Relative performance and representation metrics compared to the SSN baseline across sensory tasks.} Values indicate percentage change $\Delta = (\mu_{\text{model}} - \mu_{\text{SSN}}) / |\mu_{\text{SSN}}| \times 100\%$ (or absolute gain in dB for SNR: $\Delta\operatorname{SNR} = \mu_{\text{model}} - \mu_{\text{SSN}}$). Colored values indicate degraded performance or increased redundancy ($\textcolor{blackcherry}{\textbf{red}}$) and representational improvements ($\textcolor{oldgold}{\textbf{gold}}$) relative to the SSN. Paired two-tailed $t$-tests (or Wilcoxon signed-rank tests) with Benjamini--Hochberg FDR correction denote statistical significance ($^{*}p < 0.05$, $^{**}p < 0.01$, $^{***}p < 0.001$; \textit{n.s.}: not significant, $p \ge 0.05$).}
		\label{tab:statisticalComparison}
		\small
		\begin{tabular}{l rrr}
			\toprule
			\textbf{Metric} & \textbf{SNN (LIF)} & \textbf{Adaptive SNN} & \textbf{Poisson GLM} \\
			\midrule
			\multicolumn{4}{l}{\textit{Natural Odor Plume (PID)}} \\
			\midrule
			Reconstruction MSE                  & $\textcolor{blackcherry}{\textbf{+56.99\%}}^{*}$ & \textit{n.s.} & $\textcolor{blackcherry}{\textbf{+58.62\%}}^{*}$ \\
			Reconstruction SNR ($\mathrm{dB}$)  & $\textcolor{blackcherry}{\textbf{-1.75\text{ dB}}}^{*}$ & \textit{n.s.} & $\textcolor{blackcherry}{\textbf{-2.05\text{ dB}}}^{*}$ \\
			Stimulus MI                         & $\textcolor{blackcherry}{\textbf{-49.66\%}}^{***}$ & $\textcolor{blackcherry}{\textbf{-61.05\%}}^{***}$ & $\textcolor{oldgold}{\textbf{+22.63\%}}^{*}$ \\
			Shannon Entropy                     & $\textcolor{blackcherry}{\textbf{-70.06\%}}^{***}$ & $\textcolor{blackcherry}{\textbf{-68.87\%}}^{***}$ & $\textcolor{oldgold}{\textbf{+13.55\%}}^{*}$ \\
			$\mathrm{CV}_{\mathrm{ISI}}$        & \textit{n.s.} & $\textcolor{blackcherry}{\textbf{-22.98\%}}^{*}$ & $\textcolor{blackcherry}{\textbf{-60.98\%}}^{***}$ \\
			Participation Ratio                 & $\textcolor{blackcherry}{\textbf{-67.82\%}}^{***}$ & $\textcolor{blackcherry}{\textbf{-62.14\%}}^{***}$ & $\textcolor{blackcherry}{\textbf{-15.64\%}}^{*}$ \\
			Spatial Correlation                 & $\textcolor{blackcherry}{\textbf{+76.68\%}}^{**}$ & $\textcolor{blackcherry}{\textbf{+198.48\%}}^{***}$ & $\textcolor{blackcherry}{\textbf{+61.95\%}}^{**}$ \\
			Cross-Channel MI                    & $\textcolor{blackcherry}{\textbf{+90.88\%}}^{**}$ & \textit{n.s.} & $\textcolor{blackcherry}{\textbf{+106.93\%}}^{***}$ \\
			Coding Efficiency                   & $\textcolor{blackcherry}{\textbf{-4.31\%}}^{**}$ & $\textcolor{blackcherry}{\textbf{-0.99\%}}^{*}$ & $\textcolor{blackcherry}{\textbf{-1.56\%}}^{***}$ \\
			\midrule
			\multicolumn{4}{l}{\textit{Synthetic Wave}} \\
			\midrule
			Reconstruction MSE                  & \textit{n.s.} & $\textcolor{blackcherry}{\textbf{+108.99\%}}^{***}$ & $\textcolor{blackcherry}{\textbf{+85.52\%}}^{***}$ \\
			Reconstruction SNR ($\mathrm{dB}$)  & \textit{n.s.} & $\textcolor{blackcherry}{\textbf{-3.19\text{ dB}}}^{***}$ & $\textcolor{blackcherry}{\textbf{-2.69\text{ dB}}}^{***}$ \\
			Stimulus MI                         & $\textcolor{blackcherry}{\textbf{-15.13\%}}^{*}$ & $\textcolor{blackcherry}{\textbf{-20.65\%}}^{**}$ & $\textcolor{blackcherry}{\textbf{-17.69\%}}^{***}$ \\
			Shannon Entropy                     & $\textcolor{blackcherry}{\textbf{-22.64\%}}^{***}$ & $\textcolor{blackcherry}{\textbf{-20.77\%}}^{***}$ & $\textcolor{oldgold}{\textbf{+4.54\%}}^{*}$ \\
			$\mathrm{CV}_{\mathrm{ISI}}$        & $\textcolor{oldgold}{\textbf{+27.79\%}}^{**}$ & \textit{n.s.} & \textit{n.s.} \\
			Participation Ratio                 & $\textcolor{blackcherry}{\textbf{-9.69\%}}^{*}$ & $\textcolor{blackcherry}{\textbf{-13.83\%}}^{**}$ & $\textcolor{blackcherry}{\textbf{-8.41\%}}^{**}$ \\
			Spatial Correlation                 & \textit{n.s.} & $\textcolor{blackcherry}{\textbf{+27.66\%}}^{**}$ & $\textcolor{blackcherry}{\textbf{+20.12\%}}^{***}$ \\
			Cross-Channel MI                    & \textit{n.s.} & \textit{n.s.} & $\textcolor{blackcherry}{\textbf{+14.30\%}}^{**}$ \\
			Coding Efficiency                   & \textit{n.s.} & \textit{n.s.} & \textit{n.s.} \\
			\bottomrule
		\end{tabular}
	\end{table}
	
	\paragraph{SSNs learn high-dimensional, decorrelated population codes on naturalistic odor plumes.} When tasked with faithfully encoding a real-world natural odor plume, the SSN outperformed the SNN ($-1.75$~dB SNR, $p = 0.03$) while maintaining an average reconstruction performance statistically indistinguishable from that of the ASNN ($p = 0.22$). However, these models rely on low-entropy, redundant population codes; relative to the SSN, both the SNN and ASNN suffer severe reductions in spike representation entropy and participation ratio while exhibiting increases in spatial correlation and cross-channel MI (Table~\ref{tab:statisticalComparison}). Consequently, coding efficiency is significantly degraded in both models, accompanied by a steep collapse in per-channel stimulus MI. This pattern was consistent on the synthetic wave task, where the SSN outperformed all baselines across stimulus MI and participation ratio, with the ASNN ($-3.19$~dB SNR) and GLM ($-2.69$~dB SNR) also failing to match the SSN's reconstruction fidelity. Although the GLM showed slightly higher entropy and stimulus MI on the plume, its degraded SNR ($-2.05$~dB) and massive cross-channel redundancy indicate that its elevated information-theoretic metrics are skewed by unconstrained noise and input duplication across channels as opposed to the model having more expressive representations.
	
	\paragraph{SSNs demonstrate superior robustness to noise.} The natural environment navigated by any organism is inherently noisy. As such, real sensory neurons have been adapted to non-stationary, imperfect stimuli. We assessed the SSNs' robustness by evaluating the models trained on the synthetic wave task whilst noise of increasing intensity is progressively added to the original wave. It can be seen in Figure~\ref{fig:noise-robustness} that the SSN is more resilient to signal corruption at levels above $\sigma_{\text{noise}} = 0.2$ when compared against the other architectures tested.
	
	\subsection{Biological properties}
	
	A viable computational model of sensory neurons should be capable of fitting \textit{in vivo} electrophysiological recordings without refined tuning. Here, we shift from pure signal reconstruction into evaluating the SSN's merits as a neuroscientific model for peripheral olfactory transduction.
	
	\begin{figure}[htbp]
		\centering
		\includegraphics[width=\textwidth]{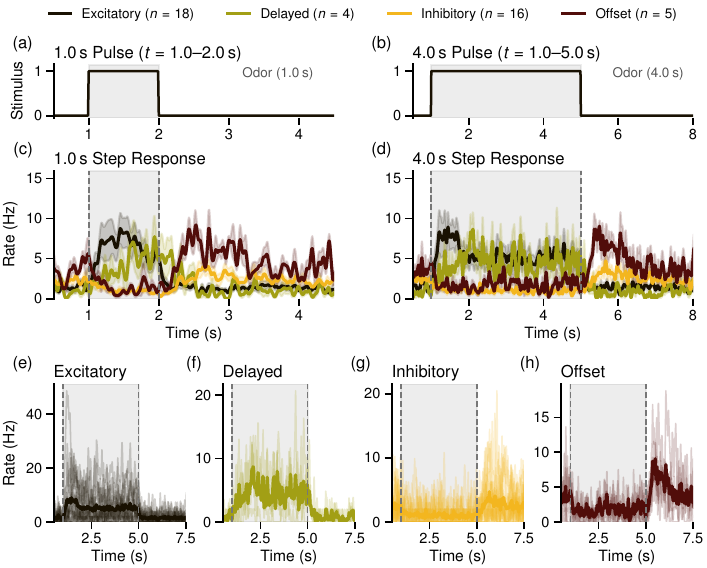}
		\caption{Average responses of trained SSNs after being given an artificial stimulus pulse. \textbf{(a)} Pulse of 1~s duration. \textbf{(b)} Pulse of 4~s duration. \textbf{(c)} Average response to the 1~s pulse. \textbf{(d)} Average response to the 4~s pulse. \textbf{(e)} Average response of the SSNs trained on an ORN classified to the excitatory motif. \textbf{(f)} Average response of the SSNs trained on an ORN classified to the delayed motif. \textbf{(g)} Average response of the SSNs trained on an ORN classified to the inhibitory motif. \textbf{(h)} Average response of the SSNs trained on an ORN classified to the offset motif. All models demonstrate very strong agreement with empirical, \textit{in vivo} results in addition to the models described by \citet{kimOlfactoryReceptorNeurons2023}.}
		\label{fig:ssn-pulse-response}
	\end{figure}
	
\begin{figure}[htbp]
		\centering
		\begin{minipage}[t]{0.48\textwidth}
			\centering
			\includegraphics[width=\linewidth]{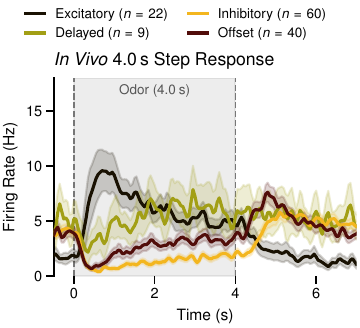}
			\caption{Empirical mean \textit{in vivo} electrophysiological response from locust ORNs during a 4~s odor pulse presentation, with all four motifs presented, serving as the biological ground truth for Figures~\ref{fig:ssn-pulse-response} and~\ref{fig:adaptation}. The data used is from \citet{kimOlfactoryReceptorNeurons2023}.}
			\label{fig:in-vivo-pulse-response}
		\end{minipage}\hfill
		\begin{minipage}[t]{0.48\textwidth}
			\centering
			\includegraphics[width=\linewidth]{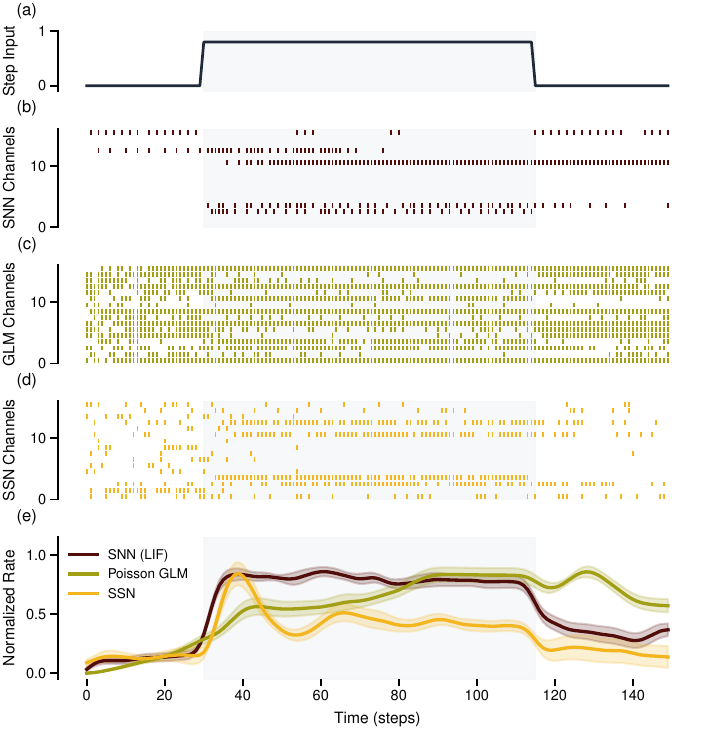}
			\caption{Spike-frequency adaptation emerges in the SSN. \textbf{(a)} Input step pulse. \textbf{(b--d)} Spike rasters of sample SNN, Poisson GLM, and SSN responses. \textbf{(e)} Averaged normalized rate response across models.}
			\label{fig:adaptation}
		\end{minipage}
	\end{figure}

	\paragraph{SSNs learn empirical ORN responses.} We considered the full 40~s-long PID and 43 ORN response dataset. It was shown by \citet{kimOlfactoryReceptorNeurons2023} that ORN responses adhered to four distinctly identified motifs: an excitatory motif (Figure~\ref{fig:ssn-pulse-response}e), wherein sensory neurons fire at a high rate upon stimulus detection; a delayed motif (Figure~\ref{fig:ssn-pulse-response}f), in which the response to the stimulus is slightly lagged and somewhat less intense; an inhibitory motif (Figure~\ref{fig:ssn-pulse-response}g), in which firing is suppressed for the duration of the stimulus before returning to baseline; and an offset motif (Figure~\ref{fig:ssn-pulse-response}h), characterized by sustained suppression during stimulation followed by a post-stimulus rebound burst. Alongside this, \citet{kimOlfactoryReceptorNeurons2023} produced a model that was successful in reproducing these empirically observed ORN responses. Yet, their model was not trained end-to-end; rather, it leveraged previously known biophysical archetypes and parameters. If our variational SSN is to be taken as a suitable model of peripheral ORNs, we expect that it should be capable of adhering to the proper motifs without class-specific parameter tuning; accordingly, we omitted any hand-tuned biophysical mechanisms in an attempt to confirm whether the SSN could discover the same behaviours from stimulus--response pairs. Following training, the models were stimulated with step pulse assays of $1$~s and $4$~s durations, mirroring examples from \citet{kimOlfactoryReceptorNeurons2023}. Visually, there is strong alignment between the SSN responses (Figure~\ref{fig:ssn-pulse-response}) and \textit{in vivo} results (also provided by \citet{kimOlfactoryReceptorNeurons2023}), shown in Figure~\ref{fig:in-vivo-pulse-response}. While the Poisson GLM also reproduced these response motifs under rate supervision (Appendix Figure~\ref{fig:glm-pulse-response}), this could not be said of either of the SNN-based models, which both failed to recreate the necessary distinct motifs (Appendix Figures~\ref{fig:snn-pulse-response} and \ref{fig:asnn-pulse-response}). Thus, the SSN was the only architecture tested that simultaneously achieved high reconstruction performance \emph{and} could be fit to empirical neurophysiological data.
	
	\paragraph{Spike-frequency adaptation emerges naturally in SSNs.} Given that the SSNs were successful in learning the specific motifs of real locust ORNs, we further sought to determine the extent to which naturalistic behaviour could arise even without being trained to deliberately match the responses found in real data. Thus, we returned to the models trained solely on odor plume reconstruction. After injecting a step pulse as stimulus (Figure~\ref{fig:adaptation}a), the responses of each model (across all seeds) were recorded, smoothed, normalized, and averaged; it was seen that telltale spike-frequency adaptation (SFA) emerged organically in the SSNs (Figure~\ref{fig:adaptation}d,e). Interestingly, the GLM did not acquire adaptation. Furthermore, we can exclude the input excitatory and inhibitory kernels as the primary driver since the SNNs equally failed to materialize SFA despite having equivalent filter banks (Figure~\ref{fig:adaptation}b,e). The dense spikes of the SSN are also far more visually reminiscent of true ORN activity \cite{kimOlfactoryReceptorNeurons2023, gorur-shandilyaOlfactoryReceptorNeurons2017, levakovaAdaptiveIntegrateandfireModel2019}.
	
	\section{Discussion}
	
	We have introduced a dynamic temporal extension to the Poisson spiking VAE framework that is closely related to, and equally inspired by, classical neuroscientific approaches used in modelling real-world sensory neurons. Empirically, SSNs match or beat the reconstruction fidelity of SNNs as well as traditional GLMs on complex time series stimuli while providing richer, higher-dimensional latent codes. Beyond signal processing, SSNs also serve as stand-ins for neuroscientific investigation; namely, they are capable of automatically discovering empirical ORN characteristics from naturalistic plume recordings without manual biophysical parametrization. Finally, we emphasize that SSNs are a \emph{class} of models whose front-end encoder can be tailored, as is done by nature, to specific stimulus modalities (e.g., a Mel filter bank for audio \cite{songSpikingLeafLearnableAuditory2024}).
	
	\paragraph{Limitations and future directions.} Although the SSN links statistical theory, peripheral neuroscience, and spike-based machine learning, the work herein possesses a few limitations that we leave for future exploration; namely, the SSN remains to be benchmarked on other datasets. Additionally, the SSN does not attempt to model higher-order aspects such as competitive lateral inhibition, feedback, or downstream secondary structures (e.g., the insect antennal lobe or mammalian olfactory bulb). Integrating SSNs into multi-layer, possibly recurrent arrangements should be investigated to evaluate their abilities within these contexts.
	
	\bibliographystyle{unsrtnat}
	\nocite{*}
	\bibliography{references}

\begin{thebibliography}{140}
\providecommand{\natexlab}[1]{#1}
\providecommand{\url}[1]{\texttt{#1}}
\expandafter\ifx\csname urlstyle\endcsname\relax
  \providecommand{\doi}[1]{doi: #1}\else
  \providecommand{\doi}{doi: \begingroup \urlstyle{rm}\Url}\fi

\bibitem[Arendt(2021)]{arendtElementaryNervousSystems2021}
Detlev Arendt.
\newblock Elementary nervous systems.
\newblock \emph{Philosophical Transactions of the Royal Society B},
  376\penalty0 (1821):\penalty0 20200347, 2021.
\newblock ISSN 0962-8436, 1471-2970.
\newblock \doi{10.1098/rstb.2020.0347}.
\newblock URL
  \url{https://royalsocietypublishing.org/rstb/article/376/1821/20200347/31504/Elementary-nervous-systemsThe-origin-of-the}.

\bibitem[Arendt et~al.(2016)Arendt, Musser, Baker, Bergman, Cepko, Erwin,
  Pavlicev, Schlosser, Widder, Laubichler, and
  Wagner]{arendtOriginEvolutionCell2016}
Detlev Arendt, Jacob~M. Musser, Clare V.~H. Baker, Aviv Bergman, Connie Cepko,
  Douglas~H. Erwin, Mihaela Pavlicev, Gerhard Schlosser, Stefanie Widder,
  Manfred~D. Laubichler, and G{\"u}nter~P. Wagner.
\newblock The origin and evolution of cell types.
\newblock \emph{Nature Reviews Genetics}, 17\penalty0 (12):\penalty0 744--757,
  2016.
\newblock ISSN 1471-0064.
\newblock \doi{10.1038/nrg.2016.127}.
\newblock URL \url{https://www.nature.com/articles/nrg.2016.127}.

\bibitem[Marmig{\`e}re and
  Ernfors(2007)]{marmigereSpecificationConnectivityNeuronal2007}
Fr{\'e}d{\'e}ric Marmig{\`e}re and Patrik Ernfors.
\newblock Specification and connectivity of neuronal subtypes in the sensory
  lineage.
\newblock \emph{Nature Reviews Neuroscience}, 8\penalty0 (2):\penalty0
  114--127, 2007.
\newblock ISSN 1471-0048.
\newblock \doi{10.1038/nrn2057}.
\newblock URL \url{https://www.nature.com/articles/nrn2057}.

\bibitem[Shepherd et~al.(2021)Shepherd, Rowe, and
  Greer]{shepherdEvolutionaryMicrocircuitApproach2021}
Gordon~M. Shepherd, Timothy~B. Rowe, and Charles~A. Greer.
\newblock An {{Evolutionary Microcircuit Approach}} to the {{Neural Basis}} of
  {{High Dimensional Sensory Processing}} in {{Olfaction}}.
\newblock \emph{Frontiers in Cellular Neuroscience}, 15:\penalty0 658480, 2021.
\newblock ISSN 1662-5102.
\newblock \doi{10.3389/fncel.2021.658480}.
\newblock URL
  \url{https://www.frontiersin.org/journals/cellular-neuroscience/articles/10.3389/fncel.2021.658480/full}.

\bibitem[Byrne(2019)]{byrneOxfordHandbookInvertebrate2019}
John~H. Byrne, editor.
\newblock \emph{The {{Oxford Handbook}} of {{Invertebrate Neurobiology}}}.
\newblock Oxford University Press, 2019.
\newblock ISBN 978-0-19-045675-7.
\newblock \doi{10.1093/oxfordhb/9780190456757.001.0001}.
\newblock URL \url{https://doi.org/10.1093/oxfordhb/9780190456757.001.0001}.

\bibitem[Gire et~al.(2013)Gire, Restrepo, Sejnowski, Greer, De~Carlos, and
  Lopez-Mascaraque]{gireTemporalProcessingOlfactory2013}
David~H. Gire, Diego Restrepo, Terrence~J. Sejnowski, Charles Greer, Juan~A.
  De~Carlos, and Laura Lopez-Mascaraque.
\newblock Temporal {{Processing}} in the {{Olfactory System}}: {{Can We See}} a
  {{Smell}}?
\newblock \emph{Neuron}, 78\penalty0 (3):\penalty0 416--432, 2013.
\newblock ISSN 0896-6273.
\newblock \doi{10.1016/j.neuron.2013.04.033}.
\newblock URL \url{https://www.cell.com/neuron/abstract/S0896-6273(13)00364-4}.

\bibitem[Kaupp(2010)]{kauppOlfactorySignallingVertebrates2010}
U.~Benjamin Kaupp.
\newblock Olfactory signalling in vertebrates and insects: Differences and
  commonalities.
\newblock \emph{Nature Reviews Neuroscience}, 11\penalty0 (3):\penalty0
  188--200, 2010.
\newblock ISSN 1471-0048.
\newblock \doi{10.1038/nrn2789}.
\newblock URL \url{https://www.nature.com/articles/nrn2789}.

\bibitem[Strausfeld and Hildebrand(1999)]{strausfeldOlfactorySystemsCommon1999}
Nicholas~J. Strausfeld and John~G. Hildebrand.
\newblock Olfactory systems: Common design, uncommon origins?
\newblock \emph{Current Opinion in Neurobiology}, 9\penalty0 (5):\penalty0
  634--639, 1999.
\newblock ISSN 0959-4388.
\newblock \doi{10.1016/S0959-4388(99)00019-7}.
\newblock URL
  \url{https://www.sciencedirect.com/science/article/pii/S0959438899000197}.

\bibitem[Hildebrand and
  Shepherd(1997)]{hildebrandMechanismsOlfactoryDiscrimination1997}
J.~G. Hildebrand and G.~M. Shepherd.
\newblock Mechanisms of olfactory discrimination: Converging evidence for
  common principles across phyla.
\newblock \emph{Annual Review of Neuroscience}, 20:\penalty0 595--631, 1997.
\newblock ISSN 0147-006X.
\newblock \doi{10.1146/annurev.neuro.20.1.595}.

\bibitem[Linsker(1988)]{linskerSelforganizationPerceptualNetwork1988}
R.~Linsker.
\newblock Self-organization in a perceptual network.
\newblock \emph{Computer}, 21\penalty0 (3):\penalty0 105--117, 1988.
\newblock ISSN 1558-0814.
\newblock \doi{10.1109/2.36}.
\newblock URL \url{https://ieeexplore.ieee.org/abstract/document/36}.

\bibitem[Laughlin et~al.(1998)Laughlin, {de Ruyter van Steveninck}, and
  Anderson]{laughlinMetabolicCostNeural1998}
Simon~B. Laughlin, Rob~R. {de Ruyter van Steveninck}, and John~C. Anderson.
\newblock The metabolic cost of neural information.
\newblock \emph{Nature Neuroscience}, 1\penalty0 (1):\penalty0 36--41, 1998.
\newblock ISSN 1546-1726.
\newblock \doi{10.1038/236}.
\newblock URL \url{https://www.nature.com/articles/nn0598_36}.

\bibitem[Stemmler and Koch(1998)]{stemmlerInformationMaximizationSingle1998}
Martin Stemmler and Christof Koch.
\newblock Information {{Maximization}} in {{Single Neurons}}.
\newblock In \emph{Advances in {{Neural Information Processing Systems}}},
  volume~11, pages 160--166. MIT Press, 1998.
\newblock URL
  \url{https://papers.nips.cc/paper_files/paper/1998/hash/452bf208bf901322968557227b8f6efe-Abstract.html}.

\bibitem[Hasenstaub et~al.(2010)Hasenstaub, Otte, Callaway, and
  Sejnowski]{hasenstaubMetabolicCostUnifying2010}
Andrea Hasenstaub, Stephani Otte, Edward Callaway, and Terrence~J. Sejnowski.
\newblock Metabolic cost as a unifying principle governing neuronal biophysics.
\newblock \emph{Proceedings of the National Academy of Sciences}, 107\penalty0
  (27):\penalty0 12329--12334, 2010.
\newblock \doi{10.1073/pnas.0914886107}.
\newblock URL \url{https://www.pnas.org/doi/full/10.1073/pnas.0914886107}.

\bibitem[Wang et~al.(2016)Wang, Wei, Stocker, and
  Lee]{wangEfficientNeuralCodes}
Zhuo Wang, Xue-Xin Wei, Alan~A. Stocker, and Daniel~D. Lee.
\newblock Efficient {{Neural Codes}} under {{Metabolic Constraints}}.
\newblock In \emph{Advances in {{Neural Information Processing Systems}}},
  volume~29, pages 4646--4654. Curran Associates, Inc., 2016.

\bibitem[Chalk et~al.(2018)Chalk, Marre, and
  Tka{\v{c}}ik]{chalkUnifiedTheoryEfficient2018}
Matthew Chalk, Olivier Marre, and Ga{\v{s}}per Tka{\v{c}}ik.
\newblock Toward a unified theory of efficient, predictive, and sparse coding.
\newblock \emph{Proceedings of the National Academy of Sciences}, 115\penalty0
  (1):\penalty0 186--191, 2018.
\newblock \doi{10.1073/pnas.1711114115}.
\newblock URL \url{https://www.pnas.org/doi/10.1073/pnas.1711114115}.

\bibitem[Helmholtz(1867)]{helmholtzHandbuchPhysiologischenOptik1867}
Hermann Helmholtz.
\newblock \emph{Handbuch der physiologischen Optik}.
\newblock Leopold Voss, Leipzig, 1867.
\newblock URL \url{http://archive.org/details/handbuchderphysi00helm}.

\bibitem[Fechner(1860)]{fechnerElementePsychophysik1860}
Gustav~Theodor Fechner.
\newblock \emph{Elemente der Psychophysik}.
\newblock Breitkopf und H{\"a}rtel, Leipzig, 1860.
\newblock URL \url{http://archive.org/details/elementederpsych001fech}.

\bibitem[Knill and Pouget(2004)]{knillBayesianBrainRole2004}
David~C. Knill and Alexandre Pouget.
\newblock The {{Bayesian}} brain: The role of uncertainty in neural coding and
  computation.
\newblock \emph{Trends in Neurosciences}, 27\penalty0 (12):\penalty0 712--719,
  2004.
\newblock ISSN 0166-2236.
\newblock \doi{10.1016/j.tins.2004.10.007}.
\newblock URL
  \url{https://www.sciencedirect.com/science/article/pii/S0166223604003352}.

\bibitem[Doya et~al.(2007)Doya, Ishii, Pouget, and
  Rao]{doyaBayesianBrainProbabilistic2011}
Kenji Doya, Shin Ishii, Alexandre Pouget, and Rajesh P.~N. Rao, editors.
\newblock \emph{Bayesian {{Brain}}: {{Probabilistic Approaches}} to {{Neural
  Coding}}}.
\newblock Computational {{Neuroscience Series}}. MIT Press, Cambridge, MA, USA,
  2007.
\newblock ISBN 978-0-262-51601-3.

\bibitem[Friston(2010)]{fristonFreeenergyPrincipleUnified2010}
Karl Friston.
\newblock The free-energy principle: A unified brain theory?
\newblock \emph{Nature Reviews Neuroscience}, 11\penalty0 (2):\penalty0
  127--138, 2010.
\newblock ISSN 1471-0048.
\newblock \doi{10.1038/nrn2787}.
\newblock URL \url{https://www.nature.com/articles/nrn2787}.

\bibitem[Aitchison et~al.(2021)Aitchison, Hennequin, and
  Lengyel]{aitchisonSamplingbasedProbabilisticInference2018}
Laurence Aitchison, Guillaume Hennequin, and M{\'a}t{\'e} Lengyel.
\newblock Sampling-based probabilistic inference emerges from learning in
  neural circuits with a cost on reliability.
\newblock \emph{Nature Neuroscience}, 24\penalty0 (4):\penalty0 565--573, 2021.
\newblock \doi{10.1038/s41593-021-00809-5}.
\newblock URL \url{http://arxiv.org/abs/1807.08952}.

\bibitem[Pillow et~al.(2008)Pillow, Shlens, Paninski, Sher, Litke,
  Chichilnisky, and Simoncelli]{pillowSpatiotemporalCorrelationsVisual2008}
Jonathan~W. Pillow, Jonathon Shlens, Liam Paninski, Alexander Sher, Alan~M.
  Litke, E.~J. Chichilnisky, and Eero~P. Simoncelli.
\newblock Spatio-temporal correlations and visual signalling in a complete
  neuronal population.
\newblock \emph{Nature}, 454\penalty0 (7207):\penalty0 995--999, 2008.
\newblock ISSN 1476-4687.
\newblock \doi{10.1038/nature07140}.
\newblock URL \url{https://www.nature.com/articles/nature07140}.

\bibitem[Capone et~al.(2018)Capone, Gigante, and
  Del~Giudice]{caponeSpontaneousActivityEmerging2018}
Cristiano Capone, Guido Gigante, and Paolo Del~Giudice.
\newblock Spontaneous activity emerging from an inferred network model captures
  complex spatio-temporal dynamics of spike data.
\newblock \emph{Scientific Reports}, 8\penalty0 (1):\penalty0 17056, 2018.
\newblock ISSN 2045-2322.
\newblock \doi{10.1038/s41598-018-35433-0}.
\newblock URL \url{https://www.nature.com/articles/s41598-018-35433-0}.

\bibitem[Vafaii et~al.(2024)Vafaii, Galor, and
  Yates]{vafaiiPoissonVariationalAutoencoder2024}
Hadi Vafaii, Dekel Galor, and Jacob~L. Yates.
\newblock Poisson {{Variational Autoencoder}}.
\newblock In \emph{Advances in {{Neural Information Processing Systems}}},
  volume~37, 2024.
\newblock \doi{10.48550/arXiv.2405.14473}.
\newblock URL \url{http://arxiv.org/abs/2405.14473}.

\bibitem[Vafaii et~al.(2025)Vafaii, Galor, and
  Yates]{vafaiiBrainlikeVariationalInference2025}
Hadi Vafaii, Dekel Galor, and Jacob~L. Yates.
\newblock Brain-like {{Variational Inference}}.
\newblock In \emph{International {{Conference}} on {{Learning Representations}}
  ({{ICLR}})}, 2025.
\newblock \doi{10.48550/arXiv.2410.19315}.
\newblock URL \url{http://arxiv.org/abs/2410.19315}.

\bibitem[Bell and Sejnowski(1997)]{bellIndependentComponentsNatural1997}
Anthony~J. Bell and Terrence~J. Sejnowski.
\newblock The ``independent components'' of natural scenes are edge filters.
\newblock \emph{Vision Research}, 37\penalty0 (23):\penalty0 3327--3338, 1997.
\newblock ISSN 0042-6989.
\newblock \doi{10.1016/S0042-6989(97)00121-1}.
\newblock URL
  \url{https://www.sciencedirect.com/science/article/pii/S0042698997001211}.

\bibitem[Fritz et~al.(2003)Fritz, Shamma, Elhilali, and
  Klein]{fritzRapidTaskrelatedPlasticity2003}
Jonathan Fritz, Shihab Shamma, Mounya Elhilali, and David Klein.
\newblock Rapid task-related plasticity of spectrotemporal receptive fields in
  primary auditory cortex.
\newblock \emph{Nature Neuroscience}, 6\penalty0 (11):\penalty0 1216--1223,
  2003.
\newblock ISSN 1546-1726.
\newblock \doi{10.1038/nn1141}.
\newblock URL \url{https://www.nature.com/articles/nn1141}.

\bibitem[Franco et~al.(2025)Franco, Garrig{\'o}s, Cap{\'o}, Serrano-Mar{\'i}n,
  Rivas-Santisteban, and Lillo]{francoOlfactoryReceptorsNeural2024}
Rafael Franco, Claudia Garrig{\'o}s, Toni Cap{\'o}, Joan Serrano-Mar{\'i}n,
  Rafael Rivas-Santisteban, and Jaume Lillo.
\newblock Olfactory receptors in neural regeneration in the central nervous
  system.
\newblock \emph{Neural Regeneration Research}, 20\penalty0 (9):\penalty0
  2480--2494, 2025.
\newblock ISSN 1673-5374.
\newblock \doi{10.4103/NRR.NRR-D-24-00495}.
\newblock URL \url{https://pmc.ncbi.nlm.nih.gov/articles/PMC11801295/}.

\bibitem[Hiratani and Latham(2020)]{hirataniRapidBayesianLearning2020}
Naoki Hiratani and Peter~E. Latham.
\newblock Rapid {{Bayesian}} learning in the mammalian olfactory system.
\newblock \emph{Nature Communications}, 11\penalty0 (1):\penalty0 3845, 2020.
\newblock ISSN 2041-1723.
\newblock \doi{10.1038/s41467-020-17490-0}.
\newblock URL \url{https://www.nature.com/articles/s41467-020-17490-0}.

\bibitem[Su et~al.(2009)Su, Menuz, and
  Carlson]{suOlfactoryPerceptionReceptors2009}
Chih-Ying Su, Karen Menuz, and John~R. Carlson.
\newblock Olfactory {{Perception}}: {{Receptors}}, {{Cells}}, and {{Circuits}}.
\newblock \emph{Cell}, 139\penalty0 (1):\penalty0 45--59, 2009.
\newblock ISSN 0092-8674.
\newblock \doi{10.1016/j.cell.2009.09.015}.
\newblock URL
  \url{https://linkinghub.elsevier.com/retrieve/pii/S0092867409011751}.

\bibitem[Vafaii and Yates(2026)]{vafaiiMetabolicCostInformation2026}
Hadi Vafaii and Jacob~L. Yates.
\newblock Metabolic cost of information processing in {{Poisson}} variational
  autoencoders.
\newblock In \emph{Proceedings of {{Cognitive Computational Neuroscience}}
  2026}, 2026.
\newblock \doi{10.32470/6ff31r0}.
\newblock URL \url{http://arxiv.org/abs/2602.13421}.

\bibitem[Balduzzi et~al.(2013)Balduzzi, Ortega, and
  Besserve]{balduzziMetabolicCostOrganizing2013}
David Balduzzi, Pedro~A. Ortega, and Michel Besserve.
\newblock Metabolic cost as an organizing principle for cooperative learning.
\newblock \emph{Advances in Complex Systems}, 16\penalty0 (02n03):\penalty0
  1350012, 2013.
\newblock \doi{10.1142/S0219525913500124}.
\newblock URL \url{http://arxiv.org/abs/1202.4482}.

\bibitem[Kingma and Welling(2014)]{kingmaAutoEncodingVariationalBayes2022}
Diederik~P. Kingma and Max Welling.
\newblock Auto-{{Encoding Variational Bayes}}.
\newblock In \emph{International {{Conference}} on {{Learning Representations}}
  ({{ICLR}})}, 2014.
\newblock URL \url{http://arxiv.org/abs/1312.6114}.

\bibitem[Rezende et~al.(2014)Rezende, Mohamed, and
  Wierstra]{rezendeStochasticBackpropagationApproximate}
Danilo~Jimenez Rezende, Shakir Mohamed, and Daan Wierstra.
\newblock Stochastic {{Backpropagation}} and {{Approximate Inference}} in
  {{Deep Generative Models}}.
\newblock In \emph{Proceedings of the 31st {{International Conference}} on
  {{Machine Learning}} ({{ICML}})}, volume~32, pages 1278--1286. PMLR, 2014.

\bibitem[Bowman et~al.(2016)Bowman, Vilnis, Vinyals, Dai, Jozefowicz, and
  Bengio]{bowmanGeneratingSentencesContinuous2016}
Samuel~R. Bowman, Luke Vilnis, Oriol Vinyals, Andrew Dai, Rafal Jozefowicz, and
  Samy Bengio.
\newblock Generating {{Sentences}} from a {{Continuous Space}}.
\newblock In Stefan Riezler and Yoav Goldberg, editors, \emph{Proceedings of
  the 20th {{SIGNLL Conference}} on {{Computational Natural Language
  Learning}}}, pages 10--21, Berlin, Germany, 2016. Association for
  Computational Linguistics.
\newblock \doi{10.18653/v1/K16-1002}.
\newblock URL \url{https://aclanthology.org/K16-1002/}.

\bibitem[S{\o}nderby et~al.(2016)S{\o}nderby, Raiko, Maal{\o}e, S{\o}nderby,
  and Winther]{sonderbyHowTrainDeep2016}
Casper~Kaae S{\o}nderby, Tapani Raiko, Lars Maal{\o}e, S{\o}ren~Kaae
  S{\o}nderby, and Ole Winther.
\newblock Ladder {{Variational Autoencoders}}.
\newblock In \emph{Advances in {{Neural Information Processing Systems}}},
  volume~29, pages 3738--3746. Curran Associates, Inc., 2016.
\newblock URL \url{http://arxiv.org/abs/1602.02282}.

\bibitem[Park et~al.(2019)Park, Kim, and
  Kim]{parkVariationalLaplaceAutoencoders2022}
Yookoon Park, Chris~Dongjoo Kim, and Gunhee Kim.
\newblock Variational {{Laplace Autoencoders}}.
\newblock In \emph{Proceedings of the 36th {{International Conference}} on
  {{Machine Learning}} ({{ICML}})}, pages 5032--5041. PMLR, 2019.
\newblock URL \url{http://arxiv.org/abs/2211.17267}.

\bibitem[Davidson et~al.(2018)Davidson, Falorsi, De~Cao, Kipf, and
  Tomczak]{davidsonHypersphericalVariationalAutoEncoders2022}
Tim~R. Davidson, Luca Falorsi, Nicola De~Cao, Thomas Kipf, and Jakub~M.
  Tomczak.
\newblock Hyperspherical {{Variational Auto-Encoders}}.
\newblock In \emph{Proceedings of the 34th {{Conference}} on {{Uncertainty}} in
  {{Artificial Intelligence}} ({{UAI}})}, pages 856--865, 2018.
\newblock URL \url{http://arxiv.org/abs/1804.00891}.

\bibitem[Kamata et~al.(2022)Kamata, Mukuta, and
  Harada]{kamataFullySpikingVariational2022}
Hiromichi Kamata, Yusuke Mukuta, and Tatsuya Harada.
\newblock Fully {{Spiking Variational Autoencoder}}.
\newblock \emph{Proceedings of the AAAI Conference on Artificial Intelligence},
  36\penalty0 (6):\penalty0 7059--7067, 2022.
\newblock ISSN 2374-3468.
\newblock \doi{10.1609/aaai.v36i6.20665}.
\newblock URL \url{https://ojs.aaai.org/index.php/AAAI/article/view/20665}.

\bibitem[Sablica and Hornik(2025)]{sablicaSphericalCauchyVariational2026}
Lukas Sablica and Kurt Hornik.
\newblock Spherical {{Cauchy Variational Autoencoders}}: {{Heavy Angular
  Tails}} and {{Exact KL Evaluation}}.
\newblock \emph{arXiv preprint arXiv:2506.21278}, 2025.
\newblock \doi{10.48550/arXiv.2506.21278}.
\newblock URL \url{http://arxiv.org/abs/2506.21278}.

\bibitem[Girin et~al.(2021)Girin, Leglaive, Bie, Diard, Hueber, and
  Alameda-Pineda]{girinDynamicalVariationalAutoencoders2022}
Laurent Girin, Simon Leglaive, Xiaoyu Bie, Julien Diard, Thomas Hueber, and
  Xavier Alameda-Pineda.
\newblock Dynamical {{Variational Autoencoders}}: {{A Comprehensive Review}}.
\newblock \emph{Foundations and Trends in Machine Learning}, 15\penalty0
  (1--2):\penalty0 1--175, 2021.
\newblock ISSN 1935-8237, 1935-8245.
\newblock \doi{10.1561/2200000089}.
\newblock URL \url{http://arxiv.org/abs/2008.12595}.

\bibitem[Chung et~al.(2015)Chung, Kastner, Dinh, Goel, Courville, and
  Bengio]{chungRecurrentLatentVariable2016}
Junyoung Chung, Kyle Kastner, Laurent Dinh, Kratarth Goel, Aaron Courville, and
  Yoshua Bengio.
\newblock A {{Recurrent Latent Variable Model}} for {{Sequential Data}}.
\newblock In \emph{Advances in {{Neural Information Processing Systems}}},
  volume~28, pages 2980--2988. Curran Associates, Inc., 2015.
\newblock URL \url{http://arxiv.org/abs/1506.02216}.

\bibitem[Fabius and {van
  Amersfoort}(2015)]{fabiusVariationalRecurrentAutoEncoders2015}
Otto Fabius and Joost~R. {van Amersfoort}.
\newblock Variational {{Recurrent Auto-Encoders}}.
\newblock In \emph{International {{Conference}} on {{Learning Representations}}
  ({{ICLR}}) {{Workshop}}}, 2015.
\newblock URL \url{http://arxiv.org/abs/1412.6581}.

\bibitem[Krishnan et~al.(2015)Krishnan, Shalit, and
  Sontag]{krishnanDeepKalmanFilters2015}
Rahul~G. Krishnan, Uri Shalit, and David Sontag.
\newblock Deep {{Kalman Filters}}.
\newblock \emph{arXiv preprint arXiv:1511.05121}, 2015.
\newblock \doi{10.48550/arXiv.1511.05121}.
\newblock URL \url{http://arxiv.org/abs/1511.05121}.

\bibitem[Cho et~al.(2014)Cho, {van Merri{\"e}nboer}, Bahdanau, and
  Bengio]{choPropertiesNeuralMachine2014}
Kyunghyun Cho, Bart {van Merri{\"e}nboer}, Dzmitry Bahdanau, and Yoshua Bengio.
\newblock On the {{Properties}} of {{Neural Machine Translation}}:
  {{Encoder-Decoder Approaches}}.
\newblock In \emph{Proceedings of {{SSST-8}}, {{Eighth Workshop}} on
  {{Syntax}}, {{Semantics}} and {{Structure}} in {{Statistical Translation}}},
  pages 103--111, 2014.
\newblock \doi{10.3115/v1/W14-4012}.
\newblock URL \url{http://arxiv.org/abs/1409.1259}.

\bibitem[Hochreiter and Schmidhuber(1997)]{hochreiterLongShortTermMemory1997}
Sepp Hochreiter and J{\"u}rgen Schmidhuber.
\newblock Long {{Short-Term Memory}}.
\newblock \emph{Neural Computation}, 9\penalty0 (8):\penalty0 1735--1780, 1997.
\newblock ISSN 0899-7667.
\newblock \doi{10.1162/neco.1997.9.8.1735}.
\newblock URL \url{https://doi.org/10.1162/neco.1997.9.8.1735}.

\bibitem[Barber and Agakov(2003)]{barberIMAlgorithmVariational}
David Barber and Felix Agakov.
\newblock The {{IM Algorithm}}: {{A}} variational approach to {{Information
  Maximization}}.
\newblock In \emph{Advances in {{Neural Information Processing Systems}}},
  volume~16, pages 201--208. MIT Press, 2003.

\bibitem[Brekelmans et~al.(2022)Brekelmans, Huang, Ghassemi, Ver~Steeg, Grosse,
  and Makhzani]{brekelmansImprovingMutualInformation2023}
Rob Brekelmans, Sicong Huang, Marzyeh Ghassemi, Greg Ver~Steeg, Roger Grosse,
  and Alireza Makhzani.
\newblock Improving {{Mutual Information Estimation}} with {{Annealed}} and
  {{Energy-Based Bounds}}.
\newblock In \emph{International {{Conference}} on {{Learning Representations}}
  ({{ICLR}})}, 2022.
\newblock URL \url{http://arxiv.org/abs/2303.06992}.

\bibitem[Prasad(2010)]{prasadCertainRelationsMutual2010}
Sudhakar Prasad.
\newblock Certain {{Relations}} between {{Mutual Information}} and {{Fidelity}}
  of {{Statistical Estimation}}.
\newblock \emph{arXiv preprint arXiv:1010.1508}, 2010.
\newblock \doi{10.48550/arXiv.1010.1508}.
\newblock URL \url{http://arxiv.org/abs/1010.1508}.

\bibitem[Kim et~al.(2023)Kim, Haney, Milan, Joshi, Aldworth, Rulkov, Kim,
  Bazhenov, and Stopfer]{kimOlfactoryReceptorNeurons2023}
Brian Kim, Seth Haney, Ana~P. Milan, Shruti Joshi, Zane Aldworth, Nikolai
  Rulkov, Alexander~T. Kim, Maxim Bazhenov, and Mark~A. Stopfer.
\newblock Olfactory receptor neurons generate multiple response motifs,
  increasing coding space dimensionality.
\newblock \emph{eLife}, 12:\penalty0 e79152, 2023.
\newblock ISSN 2050-084X.
\newblock \doi{10.7554/eLife.79152}.
\newblock URL \url{https://doi.org/10.7554/eLife.79152}.

\bibitem[Duchamp-Viret et~al.(1999)Duchamp-Viret, Chaput, and
  Duchamp]{duchamp-viretOdorResponseProperties1999}
P.~Duchamp-Viret, M.~A. Chaput, and A.~Duchamp.
\newblock Odor {{Response Properties}} of {{Rat Olfactory Receptor Neurons}}.
\newblock \emph{Science}, 284\penalty0 (5423):\penalty0 2171--2174, 1999.
\newblock \doi{10.1126/science.284.5423.2171}.
\newblock URL \url{https://www.science.org/doi/10.1126/science.284.5423.2171}.

\bibitem[Reisert(2010)]{reisertOriginBasalActivity2010}
Johannes Reisert.
\newblock Origin of basal activity in mammalian olfactory receptor neurons.
\newblock \emph{The Journal of General Physiology}, 136\penalty0 (5):\penalty0
  529--540, 2010.
\newblock ISSN 0022-1295.
\newblock \doi{10.1085/jgp.201010528}.
\newblock URL \url{https://pmc.ncbi.nlm.nih.gov/articles/PMC2964517/}.

\bibitem[Gerstner et~al.(2014)Gerstner, Kistler, Naud, and
  Paninski]{gerstnerNeuronalDynamicsSingle2014}
Wulfram Gerstner, Werner~M. Kistler, Richard Naud, and Liam Paninski.
\newblock \emph{Neuronal {{Dynamics}}: {{From Single Neurons}} to {{Networks}}
  and {{Models}} of {{Cognition}}}.
\newblock Cambridge University Press, Cambridge, 2014.
\newblock ISBN 978-1-107-06083-8.
\newblock \doi{10.1017/CBO9781107447615}.
\newblock URL
  \url{https://www.cambridge.org/product/75375090046733765596191E23B2959D}.

\bibitem[Izhikevich(2006)]{izhikevichDynamicalSystemsNeuroscience2006}
Eugene~M. Izhikevich.
\newblock \emph{Dynamical {{Systems}} in {{Neuroscience}}: {{The Geometry}} of
  {{Excitability}} and {{Bursting}}}.
\newblock The MIT Press, 2006.
\newblock ISBN 978-0-262-27607-8.
\newblock \doi{10.7551/mitpress/2526.001.0001}.
\newblock URL
  \url{https://direct.mit.edu/books/monograph/2589/Dynamical-Systems-in-NeuroscienceThe-Geometry-of}.

\bibitem[Jolivet et~al.(2006)Jolivet, Rauch, L{\"u}scher, and
  Gerstner]{jolivetPredictingSpikeTiming2006}
Renaud Jolivet, Alexander Rauch, Hans-Rudolf L{\"u}scher, and Wulfram Gerstner.
\newblock Predicting spike timing of neocortical pyramidal neurons by simple
  threshold models.
\newblock \emph{Journal of Computational Neuroscience}, 21\penalty0
  (1):\penalty0 35--49, 2006.
\newblock ISSN 0929-5313.
\newblock \doi{10.1007/s10827-006-7074-5}.

\bibitem[Gerstner and Naud(2009)]{gerstnerNeuroscienceHowGood2009}
Wulfram Gerstner and Richard Naud.
\newblock Neuroscience: {{How}} good are neuron models?
\newblock \emph{Science}, 326\penalty0 (5951):\penalty0 379--380, 2009.
\newblock ISSN 1095-9203.
\newblock \doi{10.1126/science.1181936}.

\bibitem[Mensi et~al.(2012)Mensi, Naud, Pozzorini, Avermann, Petersen, and
  Gerstner]{mensiParameterExtractionClassification2012}
Skander Mensi, Richard Naud, Christian Pozzorini, Michael Avermann, Carl C.~H.
  Petersen, and Wulfram Gerstner.
\newblock Parameter extraction and classification of three cortical neuron
  types reveals two distinct adaptation mechanisms.
\newblock \emph{Journal of Neurophysiology}, 107\penalty0 (6):\penalty0
  1756--1775, 2012.
\newblock ISSN 1522-1598.
\newblock \doi{10.1152/jn.00408.2011}.

\bibitem[Eshraghian et~al.(2023)Eshraghian, Ward, Neftci, Wang, Lenz, Dwivedi,
  Bennamoun, Jeong, and Lu]{eshraghianTrainingSpikingNeural2023}
Jason~K. Eshraghian, Max Ward, Emre~O. Neftci, Xinxin Wang, Gregor Lenz, Girish
  Dwivedi, Mohammed Bennamoun, Doo~Seok Jeong, and Wei~D. Lu.
\newblock Training {{Spiking Neural Networks Using Lessons From Deep
  Learning}}.
\newblock \emph{Proceedings of the IEEE}, 111\penalty0 (9):\penalty0
  1016--1054, 2023.
\newblock ISSN 1558-2256.
\newblock \doi{10.1109/JPROC.2023.3308088}.
\newblock URL \url{https://ieeexplore.ieee.org/document/10242251/}.

\bibitem[Zenke and Ganguli(2018)]{zenkeSuperspikeSupervisedLearning2018}
Friedemann Zenke and Surya Ganguli.
\newblock Super{{Spike}}: {{Supervised}} learning in multilayer spiking neural
  networks.
\newblock \emph{Neural Computation}, 30\penalty0 (6):\penalty0 1514--1541,
  2018.
\newblock ISSN 0899-7667.
\newblock \doi{10.1162/neco_a_01086}.
\newblock URL \url{https://doi.org/10.1162/neco_a_01086}.

\bibitem[Tavanaei et~al.(2019)Tavanaei, Ghodrati, Kheradpisheh, Masquelier, and
  Maida]{tavanaeiDeepLearningSpiking2019}
Amirhossein Tavanaei, Masoud Ghodrati, Saeed~Reza Kheradpisheh, Timoth{\'e}e
  Masquelier, and Anthony Maida.
\newblock Deep learning in spiking neural networks.
\newblock \emph{Neural Networks}, 111:\penalty0 47--63, 2019.
\newblock ISSN 0893-6080.
\newblock \doi{10.1016/j.neunet.2018.12.002}.
\newblock URL \url{https://doi.org/10.1016/j.neunet.2018.12.002}.

\bibitem[Neftci et~al.(2019)Neftci, Mostafa, and
  Zenke]{neftciSurrogateGradientLearning2019}
Emre~O. Neftci, Hesham Mostafa, and Friedemann Zenke.
\newblock Surrogate {{Gradient Learning}} in {{Spiking Neural Networks}}:
  {{Bringing}} the {{Power}} of {{Gradient-Based Optimization}} to {{Spiking
  Neural Networks}}.
\newblock \emph{IEEE Signal Processing Magazine}, 36\penalty0 (6):\penalty0
  51--63, 2019.
\newblock ISSN 1558-0792.
\newblock \doi{10.1109/MSP.2019.2931595}.
\newblock URL \url{https://ieeexplore.ieee.org/document/8891809/}.

\bibitem[Gorur-Shandilya et~al.(2017)Gorur-Shandilya, Demir, Long, Clark, and
  Emonet]{gorur-shandilyaOlfactoryReceptorNeurons2017}
Srinivas Gorur-Shandilya, Mahmut Demir, Junjiajia Long, Damon~A. Clark, and
  Thierry Emonet.
\newblock Olfactory receptor neurons use gain control and complementary
  kinetics to encode intermittent odorant stimuli.
\newblock \emph{eLife}, 6:\penalty0 e27670, 2017.
\newblock ISSN 2050-084X.
\newblock \doi{10.7554/eLife.27670}.
\newblock URL \url{https://doi.org/10.7554/eLife.27670}.

\bibitem[Levakova et~al.(2019)Levakova, Kostal, Monsemp{\`e}s, Lucas, and
  Kobayashi]{levakovaAdaptiveIntegrateandfireModel2019}
Marie Levakova, Lubomir Kostal, Christelle Monsemp{\`e}s, Philippe Lucas, and
  Ryota Kobayashi.
\newblock Adaptive integrate-and-fire model reproduces the dynamics of
  olfactory receptor neuron responses in a moth.
\newblock \emph{Journal of the Royal Society Interface}, 16\penalty0
  (157):\penalty0 20190246, 2019.
\newblock ISSN 1742-5689.
\newblock \doi{10.1098/rsif.2019.0246}.
\newblock URL \url{https://pmc.ncbi.nlm.nih.gov/articles/PMC6731495/}.

\bibitem[Song et~al.(2024)Song, Wu, Zhang, Shou, and
  Li]{songSpikingLeafLearnableAuditory2024}
Zeyang Song, Jibin Wu, Malu Zhang, Mike~Zheng Shou, and Haizhou Li.
\newblock Spiking-{{Leaf}}: {{A Learnable Auditory Front-End}} for {{Spiking
  Neural Networks}}.
\newblock In \emph{{{ICASSP}} 2024 - 2024 {{IEEE International Conference}} on
  {{Acoustics}}, {{Speech}} and {{Signal Processing}} ({{ICASSP}})}, pages
  226--230, 2024.
\newblock \doi{10.1109/ICASSP48485.2024.10446789}.
\newblock URL \url{https://ieeexplore.ieee.org/document/10446789}.

\bibitem[Alemi et~al.(2018)Alemi, Poole, Fischer, Dillon, Saurous, and
  Murphy]{alemiFixingBrokenELBO2018}
Alexander Alemi, Ben Poole, Ian Fischer, Joshua Dillon, Rif~A. Saurous, and
  Kevin Murphy.
\newblock Fixing a {{Broken ELBO}}.
\newblock In \emph{Proceedings of the 35th {{International Conference}} on
  {{Machine Learning}}}, pages 159--168. PMLR, 2018.
\newblock URL \url{https://proceedings.mlr.press/v80/alemi18a.html}.

\bibitem[Amir et~al.(2017)Amir, Taba, Berg, Melano, McKinstry, Di~Nolfo, Nayak,
  Andreopoulos, Garreau, Mendoza, Kusnitz, Debole, Esser, Delbruck, Flickner,
  and Modha]{amirLowPowerFully2017}
Arnon Amir, Brian Taba, David Berg, Timothy Melano, Jeffrey McKinstry, Carmelo
  Di~Nolfo, Tapan Nayak, Alexander Andreopoulos, Guillaume Garreau, Marcela
  Mendoza, Jeff Kusnitz, Michael Debole, Steve Esser, Tobi Delbruck, Myron
  Flickner, and Dharmendra Modha.
\newblock A {{Low Power}}, {{Fully Event-Based Gesture Recognition System}}.
\newblock In \emph{Proceedings of the {{IEEE Conference}} on {{Computer
  Vision}} and {{Pattern Recognition}} ({{CVPR}})}, pages 7243--7252, 2017.
\newblock URL
  \url{https://openaccess.thecvf.com/content_cvpr_2017/html/Amir_A_Low_Power_CVPR_2017_paper.html}.

\bibitem[Bair and Koch(1996)]{bairTemporalPrecisionSpike1996}
W.~Bair and C.~Koch.
\newblock Temporal precision of spike trains in extrastriate cortex of the
  behaving macaque monkey.
\newblock \emph{Neural Computation}, 8\penalty0 (6):\penalty0 1185--1202, 1996.
\newblock ISSN 0899-7667.
\newblock \doi{10.1162/neco.1996.8.6.1185}.

\bibitem[Barato and Chetrite(2015)]{baratoFormalViewLevel2015}
Andre~C. Barato and Raphael Chetrite.
\newblock A {{Formal View}} on {{Level}} 2.5 {{Large Deviations}} and
  {{Fluctuation Relations}}.
\newblock \emph{Journal of Statistical Physics}, 160\penalty0 (5):\penalty0
  1154--1172, 2015.
\newblock ISSN 1572-9613.
\newblock \doi{10.1007/s10955-015-1283-0}.
\newblock URL \url{https://doi.org/10.1007/s10955-015-1283-0}.

\bibitem[Barlow(1961)]{barlowPossiblePrinciplesUnderlying2012}
H.~B. Barlow.
\newblock Possible {{Principles Underlying}} the {{Transformations}} of
  {{Sensory Messages}}.
\newblock In Walter~A. Rosenblith, editor, \emph{Sensory {{Communication}}},
  pages 217--234. The MIT Press, 1961.
\newblock ISBN 978-0-262-51842-0.
\newblock \doi{10.7551/mitpress/9780262518420.003.0013}.
\newblock URL
  \url{https://academic.oup.com/mit-press-scholarship-online/book/20714/chapter/180090664}.

\bibitem[Benjamini and Hochberg(1995)]{benjaminiControllingFalseDiscovery1995}
Yoav Benjamini and Yosef Hochberg.
\newblock Controlling the {{False Discovery Rate}}: {{A Practical}} and
  {{Powerful Approach}} to {{Multiple Testing}}.
\newblock \emph{Journal of the Royal Statistical Society: Series B
  (Methodological)}, 57\penalty0 (1):\penalty0 289--300, 1995.
\newblock ISSN 0035-9246.
\newblock URL \url{https://www.jstor.org/stable/2346101}.

\bibitem[Bishop(1995)]{bishopTrainingNoiseEquivalent1995}
Chris~M. Bishop.
\newblock Training with {{Noise}} is {{Equivalent}} to {{Tikhonov
  Regularization}}.
\newblock \emph{Neural Computation}, 7\penalty0 (1):\penalty0 108--116, 1995.
\newblock ISSN 0899-7667.
\newblock \doi{10.1162/neco.1995.7.1.108}.
\newblock URL \url{https://ieeexplore.ieee.org/document/6796505}.

\bibitem[{de Bruyne} et~al.(1999){de Bruyne}, Clyne, and
  Carlson]{bruyneOdorCodingModel1999}
Marien {de Bruyne}, Peter~J. Clyne, and John~R. Carlson.
\newblock Odor {{Coding}} in a {{Model Olfactory Organ}}: {{The}}
  \emph{Drosophila} {{Maxillary Palp}}.
\newblock \emph{Journal of Neuroscience}, 19\penalty0 (11):\penalty0
  4520--4532, 1999.
\newblock ISSN 0270-6474, 1529-2401.
\newblock \doi{10.1523/JNEUROSCI.19-11-04520.1999}.
\newblock URL \url{https://www.jneurosci.org/content/19/11/4520}.

\bibitem[Cohen(1988)]{cohenStatisticalPowerAnalysis1988}
Jacob Cohen.
\newblock \emph{Statistical Power Analysis for the Behavioral Sciences}.
\newblock L. Erlbaum Associates, Hillsdale, N.J., 2nd edition, 1988.
\newblock ISBN 978-0-8058-0283-2.

\bibitem[Cramer et~al.(2022)Cramer, Stradmann, Schemmel, and
  Zenke]{cramerHeidelbergSpikingData2022}
Benjamin Cramer, Yannik Stradmann, Johannes Schemmel, and Friedemann Zenke.
\newblock The {{Heidelberg Spiking Data Sets}} for the {{Systematic
  Evaluation}} of {{Spiking Neural Networks}}.
\newblock \emph{IEEE Transactions on Neural Networks and Learning Systems},
  33\penalty0 (7):\penalty0 2744--2757, 2022.
\newblock ISSN 2162-2388.
\newblock \doi{10.1109/TNNLS.2020.3044364}.
\newblock URL \url{https://ieeexplore.ieee.org/document/9311226/}.

\bibitem[David et~al.(2009)David, Mesgarani, Fritz, and
  Shamma]{davidRapidSynapticDepression2009}
Stephen~V. David, Nima Mesgarani, Jonathan~B. Fritz, and Shihab~A. Shamma.
\newblock Rapid {{Synaptic Depression Explains Nonlinear Modulation}} of
  {{Spectro-Temporal Tuning}} in {{Primary Auditory Cortex}} by {{Natural
  Stimuli}}.
\newblock \emph{Journal of Neuroscience}, 29\penalty0 (11):\penalty0
  3374--3386, 2009.
\newblock ISSN 0270-6474, 1529-2401.
\newblock \doi{10.1523/JNEUROSCI.5249-08.2009}.
\newblock URL \url{https://www.jneurosci.org/content/29/11/3374}.

\bibitem[Elhilali et~al.(2007)Elhilali, Fritz, Chi, and
  Shamma]{elhilaliAuditoryCorticalReceptive2007}
Mounya Elhilali, Jonathan~B. Fritz, Tai-Shih Chi, and Shihab~A. Shamma.
\newblock Auditory {{Cortical Receptive Fields}}: {{Stable Entities}} with
  {{Plastic Abilities}}.
\newblock \emph{Journal of Neuroscience}, 27\penalty0 (39):\penalty0
  10372--10382, 2007.
\newblock ISSN 0270-6474, 1529-2401.
\newblock \doi{10.1523/JNEUROSCI.1462-07.2007}.
\newblock URL \url{https://www.jneurosci.org/content/27/39/10372}.

\bibitem[Esposito and Van~den Broeck(2011)]{espositoSecondLawLandauer2011}
M.~Esposito and C.~Van~den Broeck.
\newblock Second law and {{Landauer}} principle far from equilibrium.
\newblock \emph{Europhysics Letters}, 95\penalty0 (4):\penalty0 40004, 2011.
\newblock ISSN 0295-5075.
\newblock \doi{10.1209/0295-5075/95/40004}.
\newblock URL \url{https://doi.org/10.1209/0295-5075/95/40004}.

\bibitem[Faisal et~al.(2008)Faisal, Selen, and
  Wolpert]{faisalNoiseNervousSystem2008}
A.~Aldo Faisal, Luc P.~J. Selen, and Daniel~M. Wolpert.
\newblock Noise in the nervous system.
\newblock \emph{Nature Reviews Neuroscience}, 9\penalty0 (4):\penalty0
  292--303, 2008.
\newblock ISSN 1471-0048.
\newblock \doi{10.1038/nrn2258}.

\bibitem[Fox et~al.(2010)Fox, Fairhall, and
  Daniel]{foxEncodingPropertiesHaltere2010}
Jessica~L. Fox, Adrienne~L. Fairhall, and Thomas~L. Daniel.
\newblock Encoding properties of haltere neurons enable motion feature
  detection in a biological gyroscope.
\newblock \emph{Proceedings of the National Academy of Sciences}, 107\penalty0
  (8):\penalty0 3840--3845, 2010.
\newblock \doi{10.1073/pnas.0912548107}.
\newblock URL \url{https://www.pnas.org/doi/abs/10.1073/pnas.0912548107}.

\bibitem[Gao et~al.(2017)Gao, Trautmann, Yu, Santhanam, Ryu, Shenoy, and
  Ganguli]{gaoTheoryMultineuronalDimensionality2017}
Peiran Gao, Eric Trautmann, Byron Yu, Gopal Santhanam, Stephen Ryu, Krishna
  Shenoy, and Surya Ganguli.
\newblock A theory of multineuronal dimensionality, dynamics and measurement.
\newblock \emph{bioRxiv}, page 214262, 2017.
\newblock \doi{10.1101/214262}.
\newblock URL \url{https://www.biorxiv.org/content/10.1101/214262v2}.

\bibitem[Ge and Qian(2013)]{geChemicalMasterEquation2013}
Hao Ge and Hong Qian.
\newblock Chemical {{Master Equation}}.
\newblock In \emph{Encyclopedia of {{Systems Biology}}}, pages 396--399.
  Springer, New York, NY, 2013.
\newblock ISBN 978-1-4419-9863-7.
\newblock \doi{10.1007/978-1-4419-9863-7_278}.
\newblock URL
  \url{https://link.springer.com/rwe/10.1007/978-1-4419-9863-7_278}.

\bibitem[Ge and Qian(2016)]{geNonequilibriumThermodynamicFormalism2016}
Hao Ge and Hong Qian.
\newblock Nonequilibrium thermodynamic formalism of nonlinear chemical reaction
  systems with {{Waage}}--{{Guldberg}}'s law of mass action.
\newblock \emph{Chemical Physics}, 472:\penalty0 241--248, 2016.
\newblock ISSN 0301-0104.
\newblock \doi{10.1016/j.chemphys.2016.03.026}.
\newblock URL
  \url{https://www.sciencedirect.com/science/article/pii/S0301010416300866}.

\bibitem[Ghanbari et~al.(2019)Ghanbari, Lee, Read, and
  Stevenson]{ghanbariModelingStimulusdependentVariability2019}
Abed Ghanbari, Christopher~M. Lee, Heather~L. Read, and Ian~H. Stevenson.
\newblock Modeling stimulus-dependent variability improves decoding of
  population neural responses.
\newblock \emph{Journal of Neural Engineering}, 16\penalty0 (6):\penalty0
  066018, 2019.
\newblock ISSN 1741-2552.
\newblock \doi{10.1088/1741-2552/ab3a68}.

\bibitem[Hallem and Carlson(2006)]{hallemCodingOdorsReceptor2006}
Elissa~A. Hallem and John~R. Carlson.
\newblock Coding of {{Odors}} by a {{Receptor Repertoire}}.
\newblock \emph{Cell}, 125\penalty0 (1):\penalty0 143--160, 2006.
\newblock ISSN 0092-8674, 1097-4172.
\newblock \doi{10.1016/j.cell.2006.01.050}.
\newblock URL \url{https://www.cell.com/cell/abstract/S0092-8674(06)00363-1}.

\bibitem[Hayakawa et~al.(2014)Hayakawa, Kaneko, and
  Aoyagi]{hayakawaBiologicallyPlausibleLearning2014}
Takashi Hayakawa, Takeshi Kaneko, and Toshio Aoyagi.
\newblock A biologically plausible learning rule for the {{Infomax}} on
  recurrent neural networks.
\newblock \emph{Frontiers in Computational Neuroscience}, 8:\penalty0 143,
  2014.
\newblock ISSN 1662-5188.
\newblock \doi{10.3389/fncom.2014.00143}.
\newblock URL
  \url{https://www.frontiersin.org/journals/computational-neuroscience/articles/10.3389/fncom.2014.00143/full}.

\bibitem[Higgins et~al.(2017)Higgins, Matthey, Pal, Burgess, Glorot, Botvinick,
  Mohamed, and Lerchner]{higginsVVAELEARNINGBASIC2017}
Irina Higgins, Loic Matthey, Arka Pal, Christopher Burgess, Xavier Glorot,
  Matthew Botvinick, Shakir Mohamed, and Alexander Lerchner.
\newblock Beta-{{VAE}}: {{Learning Basic Visual Concepts}} with a {{Constrained
  Variational Framework}}.
\newblock In \emph{International {{Conference}} on {{Learning Representations}}
  ({{ICLR}})}, 2017.

\bibitem[Horn and Jackson(1972)]{hornGeneralMassAction1972}
F.~Horn and R.~Jackson.
\newblock General mass action kinetics.
\newblock \emph{Archive for Rational Mechanics and Analysis}, 47\penalty0
  (2):\penalty0 81--116, 1972.
\newblock ISSN 1432-0673.
\newblock \doi{10.1007/BF00251225}.
\newblock URL \url{https://doi.org/10.1007/BF00251225}.

\bibitem[Horowitz et~al.(2017)Horowitz, Zhou, and
  England]{horowitzMinimumEnergeticCost2017}
Jordan~M. Horowitz, Kevin Zhou, and Jeremy~L. England.
\newblock Minimum energetic cost to maintain a target nonequilibrium state.
\newblock \emph{Physical Review E}, 95\penalty0 (4):\penalty0 042102, 2017.
\newblock \doi{10.1103/PhysRevE.95.042102}.
\newblock URL \url{https://link.aps.org/doi/10.1103/PhysRevE.95.042102}.

\bibitem[Ibrahim et~al.(2026)Ibrahim, Zhao, Sennesh, Li, Wu, Yates, Li, and
  Vafaii]{ibrahimHitchhikersGuidePoisson2026}
Michael Ibrahim, Hanqi Zhao, Eli Sennesh, Zhi Li, Anqi Wu, Jacob~L. Yates,
  Chengrui Li, and Hadi Vafaii.
\newblock A {{Hitchhiker}}'s {{Guide}} to {{Poisson Gradient Estimation}}.
\newblock \emph{arXiv preprint arXiv:2602.03896}, 2026.
\newblock \doi{10.48550/arXiv.2602.03896}.
\newblock URL \url{http://arxiv.org/abs/2602.03896}.

\bibitem[Mercado~III(2014)]{iiiRelatingCorticalWave2014}
Eduardo Mercado~III.
\newblock Relating {{Cortical Wave Dynamics}} to {{Learning}} and
  {{Remembering}}.
\newblock \emph{AIMS Neuroscience}, 1\penalty0 (3):\penalty0 185--209, 2014.
\newblock ISSN 2373-7972.
\newblock \doi{10.3934/Neuroscience.2014.3.185}.
\newblock URL
  \url{http://www.aimspress.com/article/doi/10.3934/Neuroscience.2014.3.185}.

\bibitem[Jang et~al.(2019)Jang, Simeone, Gardner, and
  Gruning]{jangIntroductionProbabilisticSpiking2019}
Hyeryung Jang, Osvaldo Simeone, Brian Gardner, and Andre Gruning.
\newblock An {{Introduction}} to {{Probabilistic Spiking Neural Networks}}:
  {{Probabilistic Models}}, {{Learning Rules}}, and {{Applications}}.
\newblock \emph{IEEE Signal Processing Magazine}, 36\penalty0 (6):\penalty0
  64--77, 2019.
\newblock ISSN 1558-0792.
\newblock \doi{10.1109/MSP.2019.2935234}.
\newblock URL \url{https://ieeexplore.ieee.org/document/8891810}.

\bibitem[Jarzynski(1997)]{jarzynskiNonequilibriumEqualityFree1997}
C.~Jarzynski.
\newblock Nonequilibrium {{Equality}} for {{Free Energy Differences}}.
\newblock \emph{Physical Review Letters}, 78\penalty0 (14):\penalty0
  2690--2693, 1997.
\newblock \doi{10.1103/PhysRevLett.78.2690}.
\newblock URL \url{https://link.aps.org/doi/10.1103/PhysRevLett.78.2690}.

\bibitem[Kim et~al.(2025)Kim, Peng, Chen, Liu, and
  Shlizerman]{kimGenerationBiophysicalNeuron2025}
Jimin Kim, Minxian Peng, Shuqi Chen, Qiang Liu, and Eli Shlizerman.
\newblock Generation of biophysical neuron model parameters from recorded
  electrophysiological responses.
\newblock \emph{eLife}, 13:\penalty0 RP95607, 2025.
\newblock ISSN 2050-084X.
\newblock \doi{10.7554/eLife.95607}.
\newblock URL \url{https://doi.org/10.7554/eLife.95607}.

\bibitem[Kraskov et~al.(2004)Kraskov, St{\"o}gbauer, and
  Grassberger]{kraskovEstimatingMutualInformation2004}
Alexander Kraskov, Harald St{\"o}gbauer, and Peter Grassberger.
\newblock Estimating mutual information.
\newblock \emph{Physical Review E}, 69\penalty0 (6):\penalty0 066138, 2004.
\newblock \doi{10.1103/PhysRevE.69.066138}.
\newblock URL \url{https://link.aps.org/doi/10.1103/PhysRevE.69.066138}.

\bibitem[Lan et~al.(2012)Lan, Sartori, Neumann, Sourjik, and
  Tu]{lanEnergySpeedAccuracy2012}
Ganhui Lan, Pablo Sartori, Silke Neumann, Victor Sourjik, and Yuhai Tu.
\newblock The energy--speed--accuracy trade-off in sensory adaptation.
\newblock \emph{Nature Physics}, 8\penalty0 (5):\penalty0 422--428, 2012.
\newblock ISSN 1745-2481.
\newblock \doi{10.1038/nphys2276}.
\newblock URL \url{https://www.nature.com/articles/nphys2276}.

\bibitem[Laurent(2002)]{laurentOlfactoryNetworkDynamics2002}
Gilles Laurent.
\newblock Olfactory network dynamics and the coding of multidimensional
  signals.
\newblock \emph{Nature Reviews Neuroscience}, 3\penalty0 (11):\penalty0
  884--895, 2002.
\newblock ISSN 1471-0048.
\newblock \doi{10.1038/nrn964}.
\newblock URL \url{https://www.nature.com/articles/nrn964}.

\bibitem[Lee(1998)]{leeICAFeatureExtraction1998}
Te-Won Lee.
\newblock {{ICA}} for {{Feature Extraction}}.
\newblock In Te-Won Lee, editor, \emph{Independent {{Component Analysis}}:
  {{Theory}} and {{Applications}}}, pages 167--175. Springer US, Boston, MA,
  1998.
\newblock ISBN 978-1-4757-2851-4.
\newblock \doi{10.1007/978-1-4757-2851-4_8}.
\newblock URL \url{https://doi.org/10.1007/978-1-4757-2851-4_8}.

\bibitem[Litwin-Kumar et~al.(2017)Litwin-Kumar, Harris, Axel, Sompolinsky, and
  Abbott]{litwin-kumarOptimalDegreesSynaptic2017}
Ashok Litwin-Kumar, Kameron~Decker Harris, Richard Axel, Haim Sompolinsky, and
  L.~F. Abbott.
\newblock Optimal {{Degrees}} of {{Synaptic Connectivity}}.
\newblock \emph{Neuron}, 93\penalty0 (5):\penalty0 1153--1164.e7, 2017.
\newblock ISSN 0896-6273.
\newblock \doi{10.1016/j.neuron.2017.01.030}.
\newblock URL \url{https://www.cell.com/neuron/abstract/S0896-6273(17)30054-5}.

\bibitem[Ma et~al.(2023)Ma, Yan, and Tang]{maExploitingNoiseResource2023}
Gehua Ma, Rui Yan, and Huajin Tang.
\newblock Exploiting noise as a resource for computation and learning in
  spiking neural networks.
\newblock \emph{Patterns}, 4\penalty0 (10):\penalty0 100831, 2023.
\newblock ISSN 2666-3899.
\newblock \doi{10.1016/j.patter.2023.100831}.
\newblock URL
  \url{https://www.sciencedirect.com/science/article/pii/S2666389923002003}.

\bibitem[Makkeh et~al.(2025)Makkeh, Graetz, Schneider, Ehrlich, Priesemann, and
  Wibral]{makkehGeneralFrameworkInterpretable2025}
Abdullah Makkeh, Marcel Graetz, Andreas~C. Schneider, David~A. Ehrlich, Viola
  Priesemann, and Michael Wibral.
\newblock A general framework for interpretable neural learning based on local
  information-theoretic goal functions.
\newblock \emph{Proceedings of the National Academy of Sciences}, 122\penalty0
  (10):\penalty0 e2408125122, 2025.
\newblock \doi{10.1073/pnas.2408125122}.
\newblock URL \url{https://www.pnas.org/doi/10.1073/pnas.2408125122}.

\bibitem[Marder and
  Goaillard(2006)]{marderVariabilityCompensationHomeostasis2006}
Eve Marder and Jean-Marc Goaillard.
\newblock Variability, compensation and homeostasis in neuron and network
  function.
\newblock \emph{Nature Reviews Neuroscience}, 7\penalty0 (7):\penalty0
  563--574, 2006.
\newblock ISSN 1471-003X.
\newblock \doi{10.1038/nrn1949}.

\bibitem[Meister and Berry(1999)]{meisterNeuralCodeRetina1999}
Markus Meister and Michael~J. Berry.
\newblock The {{Neural Code}} of the {{Retina}}.
\newblock \emph{Neuron}, 22\penalty0 (3):\penalty0 435--450, 1999.
\newblock ISSN 0896-6273.
\newblock \doi{10.1016/S0896-6273(00)80700-X}.
\newblock URL \url{https://www.cell.com/neuron/abstract/S0896-6273(00)80700-X}.

\bibitem[Meyer et~al.(2014)Meyer, Diepenbrock, Happel, Ohl, and
  Anem{\"u}ller]{meyerDiscriminativeLearningReceptive2014}
Arne~F. Meyer, Jan-Philipp Diepenbrock, Max F.~K. Happel, Frank~W. Ohl, and
  J{\"o}rn Anem{\"u}ller.
\newblock Discriminative {{Learning}} of {{Receptive Fields}} from
  {{Responses}} to {{Non-Gaussian Stimulus Ensembles}}.
\newblock \emph{PLOS ONE}, 9\penalty0 (4):\penalty0 e93062, 2014.
\newblock ISSN 1932-6203.
\newblock \doi{10.1371/journal.pone.0093062}.
\newblock URL
  \url{https://journals.plos.org/plosone/article?id=10.1371/journal.pone.0093062}.

\bibitem[Miller(1955)]{miller1955}
George Miller.
\newblock Note on the bias of information estimates.
\newblock \emph{Information theory in psychology: Problems and methods}, 1955.

\bibitem[Mozolewski et~al.(2021)Mozolewski, Jeziorek, Schuster, Bading, Frost,
  and Dobrowolski]{mozolewskiRoleNuclearCa22021}
Pawel Mozolewski, Maciej Jeziorek, Christoph~M. Schuster, Hilmar Bading, Bess
  Frost, and Radek Dobrowolski.
\newblock The role of nuclear {{Ca}}$^{2+}$ in maintaining neuronal homeostasis
  and brain health.
\newblock \emph{Journal of Cell Science}, 134\penalty0 (8):\penalty0 jcs254904,
  2021.
\newblock ISSN 0021-9533.
\newblock \doi{10.1242/jcs.254904}.
\newblock URL \url{https://doi.org/10.1242/jcs.254904}.

\bibitem[Ohga(2022)]{ohgaInformationgeometricStructureChemical2022}
Naruo Ohga.
\newblock Information-geometric structure for chemical thermodynamics: {{An}}
  explicit construction of dual affine coordinates.
\newblock \emph{Physical Review E}, 106\penalty0 (4):\penalty0 044131, 2022.
\newblock \doi{10.1103/PhysRevE.106.044131}.

\bibitem[O'Leary et~al.(2014)O'Leary, Williams, Franci, and
  Marder]{olearyCellTypesNetwork2014}
Timothy O'Leary, Alex~H. Williams, Alessio Franci, and Eve Marder.
\newblock Cell {{Types}}, {{Network Homeostasis}}, and {{Pathological
  Compensation}} from a {{Biologically Plausible Ion Channel Expression
  Model}}.
\newblock \emph{Neuron}, 82\penalty0 (4):\penalty0 809--821, 2014.
\newblock ISSN 0896-6273.
\newblock \doi{10.1016/j.neuron.2014.04.002}.
\newblock URL \url{https://www.cell.com/neuron/abstract/S0896-6273(14)00292-X}.

\bibitem[Olshausen and Field(1996)]{olshausenEmergenceSimplecellReceptive1996}
Bruno~A. Olshausen and David~J. Field.
\newblock Emergence of simple-cell receptive field properties by learning a
  sparse code for natural images.
\newblock \emph{Nature}, 381\penalty0 (6583):\penalty0 607--609, 1996.
\newblock ISSN 1476-4687.
\newblock \doi{10.1038/381607a0}.
\newblock URL \url{https://www.nature.com/articles/381607a0}.

\bibitem[Olshausen and Field(2004)]{olshausenSparseCodingSensory2004}
Bruno~A. Olshausen and David~J. Field.
\newblock Sparse coding of sensory inputs.
\newblock \emph{Current Opinion in Neurobiology}, 14\penalty0 (4):\penalty0
  481--487, 2004.
\newblock ISSN 0959-4388.
\newblock \doi{10.1016/j.conb.2004.07.007}.

\bibitem[Pakdaman et~al.(2010)Pakdaman, Thieullen, and
  Wainrib]{pakdamanFluidLimitTheorems2010}
K.~Pakdaman, M.~Thieullen, and G.~Wainrib.
\newblock Fluid limit theorems for stochastic hybrid systems with application
  to neuron models.
\newblock \emph{Advances in Applied Probability}, 42\penalty0 (3):\penalty0
  761--794, 2010.
\newblock ISSN 0001-8678, 1475-6064.
\newblock \doi{10.1239/aap/1282924062}.
\newblock URL
  \url{https://www.cambridge.org/core/product/identifier/S0001867800050436/type/journal_article}.

\bibitem[Pandarinath et~al.(2018)Pandarinath, O'Shea, Collins, Jozefowicz,
  Stavisky, Kao, Trautmann, Kaufman, Ryu, Hochberg, Henderson, Shenoy, Abbott,
  and Sussillo]{pandarinathInferringSingletrialNeural2018}
Chethan Pandarinath, Daniel~J. O'Shea, Jasmine Collins, Rafal Jozefowicz,
  Sergey~D. Stavisky, Jonathan~C. Kao, Eric~M. Trautmann, Matthew~T. Kaufman,
  Stephen~I. Ryu, Leigh~R. Hochberg, Jaimie~M. Henderson, Krishna~V. Shenoy,
  L.~F. Abbott, and David Sussillo.
\newblock Inferring single-trial neural population dynamics using sequential
  auto-encoders.
\newblock \emph{Nature Methods}, 15\penalty0 (10):\penalty0 805--815, 2018.
\newblock ISSN 1548-7105.
\newblock \doi{10.1038/s41592-018-0109-9}.
\newblock URL \url{https://www.nature.com/articles/s41592-018-0109-9}.

\bibitem[Pandey et~al.(2022)Pandey, Pachitariu, Brunton, and
  Harris]{pandeyStructuredRandomReceptive2022}
Biraj Pandey, Marius Pachitariu, Bingni~W. Brunton, and Kameron~Decker Harris.
\newblock Structured random receptive fields enable informative sensory
  encodings.
\newblock \emph{PLOS Computational Biology}, 18\penalty0 (10):\penalty0
  e1010484, 2022.
\newblock ISSN 1553-734X.
\newblock \doi{10.1371/journal.pcbi.1010484}.
\newblock URL \url{https://pmc.ncbi.nlm.nih.gov/articles/PMC9584455/}.

\bibitem[Paninski(2004)]{paninskiMaximumLikelihoodEstimation2004}
Liam Paninski.
\newblock Maximum likelihood estimation of cascade point-process neural
  encoding models.
\newblock \emph{Network: Computation in Neural Systems}, 15\penalty0
  (4):\penalty0 243--262, 2004.
\newblock ISSN 0954-898X.

\bibitem[Parrondo et~al.(2015)Parrondo, Horowitz, and
  Sagawa]{parrondoThermodynamicsInformation2015}
Juan M.~R. Parrondo, Jordan~M. Horowitz, and Takahiro Sagawa.
\newblock Thermodynamics of information.
\newblock \emph{Nature Physics}, 11\penalty0 (2):\penalty0 131--139, 2015.
\newblock ISSN 1745-2481.
\newblock \doi{10.1038/nphys3230}.
\newblock URL \url{https://www.nature.com/articles/nphys3230}.

\bibitem[Pedregosa et~al.(2011)Pedregosa, Varoquaux, Gramfort, Michel, Thirion,
  Grisel, Blondel, Prettenhofer, Weiss, Dubourg, Vanderplas, Passos,
  Cournapeau, Brucher, Perrot, and
  Duchesnay]{pedregosaScikitlearnMachineLearning}
Fabian Pedregosa, Ga{\"e}l Varoquaux, Alexandre Gramfort, Vincent Michel,
  Bertrand Thirion, Olivier Grisel, Mathieu Blondel, Peter Prettenhofer, Ron
  Weiss, Vincent Dubourg, Jake Vanderplas, Alexandre Passos, David Cournapeau,
  Matthieu Brucher, Matthieu Perrot, and {\'E}douard Duchesnay.
\newblock Scikit-learn: {{Machine Learning}} in {{Python}}.
\newblock \emph{Journal of Machine Learning Research}, 12:\penalty0 2825--2830,
  2011.
\newblock URL \url{https://jmlr.org/papers/v12/pedregosa11a.html}.

\bibitem[Perez-Orive et~al.(2002)Perez-Orive, Mazor, Turner, Cassenaer, Wilson,
  and Laurent]{perez-oriveOscillationsSparseningOdor2002}
Javier Perez-Orive, Ofer Mazor, Glenn~C. Turner, Stijn Cassenaer, Rachel~I.
  Wilson, and Gilles Laurent.
\newblock Oscillations and sparsening of odor representations in the mushroom
  body.
\newblock \emph{Science}, 297\penalty0 (5580):\penalty0 359--365, 2002.
\newblock ISSN 1095-9203.
\newblock \doi{10.1126/science.1070502}.

\bibitem[Pratt et~al.(2017)Pratt, Deora, Mohren, and
  Daniel]{prattNeuralEvidenceSupports2017}
Brandon Pratt, Tanvi Deora, Thomas Mohren, and Thomas Daniel.
\newblock Neural evidence supports a dual sensory-motor role for insect wings.
\newblock \emph{Proceedings of the Royal Society B: Biological Sciences},
  284\penalty0 (1862):\penalty0 20170969, 2017.
\newblock ISSN 0962-8452.
\newblock \doi{10.1098/rspb.2017.0969}.
\newblock URL \url{https://doi.org/10.1098/rspb.2017.0969}.

\bibitem[Pratt(1959)]{prattRemarksZerosTies1959}
John~W. Pratt.
\newblock Remarks on {{Zeros}} and {{Ties}} in the {{Wilcoxon Signed Rank
  Procedures}}.
\newblock \emph{Journal of the American Statistical Association}, 54\penalty0
  (287):\penalty0 655--667, 1959.
\newblock ISSN 0162-1459.
\newblock \doi{10.2307/2282543}.
\newblock URL \url{https://www.jstor.org/stable/2282543}.

\bibitem[Rao and Esposito(2016)]{raoNonequilibriumThermodynamicsChemical2016}
Riccardo Rao and Massimiliano Esposito.
\newblock Nonequilibrium {{Thermodynamics}} of {{Chemical Reaction Networks}}:
  {{Wisdom}} from {{Stochastic Thermodynamics}}.
\newblock \emph{Physical Review X}, 6\penalty0 (4):\penalty0 041064, 2016.
\newblock \doi{10.1103/PhysRevX.6.041064}.

\bibitem[Ross(2014)]{rossMutualInformationDiscrete2014}
Brian~C. Ross.
\newblock Mutual {{Information}} between {{Discrete}} and {{Continuous Data
  Sets}}.
\newblock \emph{PLOS ONE}, 9\penalty0 (2):\penalty0 e87357, 2014.
\newblock ISSN 1932-6203.
\newblock \doi{10.1371/journal.pone.0087357}.
\newblock URL
  \url{https://journals.plos.org/plosone/article?id=10.1371/journal.pone.0087357}.

\bibitem[Schmiedl and
  Seifert(2007)]{schmiedlStochasticThermodynamicsChemical2007}
Tim Schmiedl and Udo Seifert.
\newblock Stochastic thermodynamics of chemical reaction networks.
\newblock \emph{The Journal of Chemical Physics}, 126\penalty0 (4):\penalty0
  044101, 2007.
\newblock ISSN 0021-9606, 1089-7690.
\newblock \doi{10.1063/1.2428297}.
\newblock URL \url{http://arxiv.org/abs/cond-mat/0605080}.

\bibitem[Seifert(2005)]{seifertEntropyProductionStochastic2005}
Udo Seifert.
\newblock Entropy {{Production}} along a {{Stochastic Trajectory}} and an
  {{Integral Fluctuation Theorem}}.
\newblock \emph{Physical Review Letters}, 95\penalty0 (4):\penalty0 040602,
  2005.
\newblock \doi{10.1103/PhysRevLett.95.040602}.
\newblock URL \url{https://link.aps.org/doi/10.1103/PhysRevLett.95.040602}.

\bibitem[Seifert(2012)]{seifertStochasticThermodynamicsFluctuation2012}
Udo Seifert.
\newblock Stochastic thermodynamics, fluctuation theorems, and molecular
  machines.
\newblock \emph{Reports on Progress in Physics}, 75\penalty0 (12):\penalty0
  126001, 2012.
\newblock ISSN 0034-4885, 1361-6633.
\newblock \doi{10.1088/0034-4885/75/12/126001}.
\newblock URL \url{http://arxiv.org/abs/1205.4176}.

\bibitem[Shadlen and Newsome(1998)]{shadlenVariableDischargeCortical1998}
Michael~N. Shadlen and William~T. Newsome.
\newblock The {{Variable Discharge}} of {{Cortical Neurons}}: {{Implications}}
  for {{Connectivity}}, {{Computation}}, and {{Information Coding}}.
\newblock \emph{Journal of Neuroscience}, 18\penalty0 (10):\penalty0
  3870--3896, 1998.
\newblock ISSN 0270-6474, 1529-2401.
\newblock \doi{10.1523/JNEUROSCI.18-10-03870.1998}.
\newblock URL \url{https://www.jneurosci.org/content/18/10/3870}.

\bibitem[Shamma(2013)]{shammaSpectroTemporalReceptiveFields2013}
Shihab Shamma.
\newblock Spectro-{{Temporal Receptive Fields}}.
\newblock In \emph{Encyclopedia of {{Computational Neuroscience}}}, pages 1--6.
  Springer, New York, NY, 2013.
\newblock ISBN 978-1-4614-7320-6.
\newblock \doi{10.1007/978-1-4614-7320-6_437-1}.
\newblock URL
  \url{https://link.springer.com/rwe/10.1007/978-1-4614-7320-6_437-1}.

\bibitem[Shannon(1948)]{shannonMathematicalTheoryCommunication1948}
C.~E. Shannon.
\newblock A mathematical theory of communication.
\newblock \emph{The Bell System Technical Journal}, 27\penalty0 (3):\penalty0
  379--423, 1948.
\newblock ISSN 0005-8580.
\newblock \doi{10.1002/j.1538-7305.1948.tb01338.x}.
\newblock URL \url{https://ieeexplore.ieee.org/document/6773024}.

\bibitem[Shapiro and Wilk(1965)]{shapiroAnalysisVarianceTest1965}
S.~S. Shapiro and M.~B. Wilk.
\newblock An {{Analysis}} of {{Variance Test}} for {{Normality}} ({{Complete
  Samples}}).
\newblock \emph{Biometrika}, 52\penalty0 (3/4):\penalty0 591--611, 1965.
\newblock ISSN 0006-3444.
\newblock \doi{10.2307/2333709}.
\newblock URL \url{https://www.jstor.org/stable/2333709}.

\bibitem[Shiraishi(2023)]{shiraishiIntroductionStochasticThermodynamics2023}
Naoto Shiraishi.
\newblock \emph{An {{Introduction}} to {{Stochastic Thermodynamics}}: {{From
  Basic}} to {{Advanced}}}, volume 212 of \emph{Fundamental {{Theories}} of
  {{Physics}}}.
\newblock Springer Nature, Singapore, 2023.
\newblock ISBN 978-981-19-8185-2.
\newblock \doi{10.1007/978-981-19-8186-9}.
\newblock URL \url{https://link.springer.com/10.1007/978-981-19-8186-9}.

\bibitem[Softky and Koch(1993)]{softkyHighlyIrregularFiring1993}
W.~R. Softky and C.~Koch.
\newblock The highly irregular firing of cortical cells is inconsistent with
  temporal integration of random {{EPSPs}}.
\newblock \emph{Journal of Neuroscience}, 13\penalty0 (1):\penalty0 334--350,
  1993.
\newblock ISSN 0270-6474.
\newblock \doi{10.1523/JNEUROSCI.13-01-00334.1993}.

\bibitem[Stroud et~al.(2025)Stroud, Wojcik, Jensen, Kusunoki, Kadohisa,
  Buckley, Duncan, Stokes, and Lengyel]{stroudEffectsNoiseMetabolic2025}
Jake~Patrick Stroud, Michal Wojcik, Kristopher~Torp Jensen, Makoto Kusunoki,
  Mikiko Kadohisa, Mark~J. Buckley, John Duncan, Mark~G. Stokes, and
  M{\'a}t{\'e} Lengyel.
\newblock Effects of noise and metabolic cost on cortical task representations.
\newblock \emph{eLife}, 13:\penalty0 RP94961, 2025.
\newblock ISSN 2050-084X.
\newblock \doi{10.7554/eLife.94961}.
\newblock URL \url{https://doi.org/10.7554/eLife.94961}.

\bibitem[Tanaka and Nakamura(2013)]{tanakaInformationMaximizationPrinciple2013}
Takuma Tanaka and Kiyohiko Nakamura.
\newblock Information maximization principle explains the emergence of complex
  cell-like neurons.
\newblock \emph{Frontiers in Computational Neuroscience}, 7:\penalty0 165,
  2013.
\newblock ISSN 1662-5188.
\newblock \doi{10.3389/fncom.2013.00165}.
\newblock URL
  \url{https://www.frontiersin.org/journals/computational-neuroscience/articles/10.3389/fncom.2013.00165/full}.

\bibitem[Theunissen et~al.(2000)Theunissen, Sen, and
  Doupe]{theunissenSpectralTemporalReceptiveFields2000}
Fr{\'e}d{\'e}ric~E. Theunissen, Kamal Sen, and Allison~J. Doupe.
\newblock Spectral-{{Temporal Receptive Fields}} of {{Nonlinear Auditory
  Neurons Obtained Using Natural Sounds}}.
\newblock \emph{Journal of Neuroscience}, 20\penalty0 (6):\penalty0 2315--2331,
  2000.
\newblock ISSN 0270-6474, 1529-2401.
\newblock \doi{10.1523/JNEUROSCI.20-06-02315.2000}.
\newblock URL \url{https://www.jneurosci.org/content/20/6/2315}.

\bibitem[Todorov(2009)]{todorovEfficientComputationOptimal2009}
Emanuel Todorov.
\newblock Efficient computation of optimal actions.
\newblock \emph{Proceedings of the National Academy of Sciences}, 106\penalty0
  (28):\penalty0 11478--11483, 2009.
\newblock \doi{10.1073/pnas.0710743106}.
\newblock URL \url{https://www.pnas.org/doi/10.1073/pnas.0710743106}.

\bibitem[Toups et~al.(2011)Toups, Fellous, Thomas, Sejnowski, and
  Tiesinga]{toupsFindingEventStructure2011}
J.~Vincent Toups, Jean-Marc Fellous, Peter~J. Thomas, Terrence~J. Sejnowski,
  and Paul~H. Tiesinga.
\newblock Finding the {{Event Structure}} of {{Neuronal Spike Trains}}.
\newblock \emph{Neural Computation}, 23\penalty0 (9):\penalty0 2169--2208,
  2011.
\newblock ISSN 0899-7667.
\newblock \doi{10.1162/NECO_a_00173}.
\newblock URL \url{https://pmc.ncbi.nlm.nih.gov/articles/PMC3220920/}.

\bibitem[Toyoizumi et~al.(2005)Toyoizumi, Pfister, Aihara, and
  Gerstner]{toyoizumiGeneralizedBienenstockCooper2005}
Taro Toyoizumi, Jean-Pascal Pfister, Kazuyuki Aihara, and Wulfram Gerstner.
\newblock Generalized {{Bienenstock}}--{{Cooper}}--{{Munro}} rule for spiking
  neurons that maximizes information transmission.
\newblock \emph{Proceedings of the National Academy of Sciences}, 102\penalty0
  (14):\penalty0 5239--5244, 2005.
\newblock ISSN 0027-8424, 1091-6490.
\newblock \doi{10.1073/pnas.0500495102}.
\newblock URL \url{https://pnas.org/doi/full/10.1073/pnas.0500495102}.

\bibitem[Turrigiano(2012)]{turrigianoHomeostaticSynapticPlasticity2012}
Gina Turrigiano.
\newblock Homeostatic {{Synaptic Plasticity}}: {{Local}} and {{Global
  Mechanisms}} for {{Stabilizing Neuronal Function}}.
\newblock \emph{Cold Spring Harbor Perspectives in Biology}, 4\penalty0
  (1):\penalty0 a005736, 2012.
\newblock ISSN 1943-0264.
\newblock \doi{10.1101/cshperspect.a005736}.
\newblock URL \url{https://pmc.ncbi.nlm.nih.gov/articles/PMC3249629/}.

\bibitem[Wilcoxon(1945)]{wilcoxonIndividualComparisonsRanking1945}
Frank Wilcoxon.
\newblock Individual {{Comparisons}} by {{Ranking Methods}}.
\newblock \emph{Biometrics Bulletin}, 1\penalty0 (6):\penalty0 80--83, 1945.
\newblock ISSN 0099-4987.
\newblock \doi{10.2307/3001968}.
\newblock URL \url{https://www.jstor.org/stable/3001968}.

\bibitem[Zhang et~al.(2025)Zhang, Sheng, Zhang, Kong, and
  Zhou]{zhangNegativeBinomialVariational}
Yixuan Zhang, Jinhao Sheng, Wenxin Zhang, Quyu Kong, and Feng Zhou.
\newblock Negative {{Binomial Variational Autoencoders}} for {{Overdispersed
  Latent Modeling}}.
\newblock \emph{arXiv preprint arXiv:2505.19518}, 2025.
\newblock URL \url{https://github.com/co234/NegBio-VAE}.

\bibitem[Zhou and Wei(2020)]{zhouLearningIdentifiableInterpretable2020}
Ding Zhou and Xue-Xin Wei.
\newblock Learning identifiable and interpretable latent models of
  high-dimensional neural activity using pi-{{VAE}}.
\newblock In \emph{Advances in {{Neural Information Processing Systems}}},
  volume~33, pages 7234--7247. Curran Associates, Inc., 2020.
\newblock URL
  \url{https://proceedings.neurips.cc/paper_files/paper/2020/hash/510f2318f324cf07fce24c3a4b89c771-Abstract.html}.

\bibitem[Ziener et~al.(2015)Ziener, Maritan, and
  Hinrichsen]{zienerEntropyProductionNonequilibrium2015}
Robert Ziener, Amos Maritan, and Haye Hinrichsen.
\newblock On entropy production in nonequilibrium systems.
\newblock \emph{Journal of Statistical Mechanics: Theory and Experiment},
  2015\penalty0 (8):\penalty0 P08014, 2015.
\newblock ISSN 1742-5468.
\newblock \doi{10.1088/1742-5468/2015/08/P08014}.
\newblock URL \url{https://doi.org/10.1088/1742-5468/2015/08/P08014}.

\end{thebibliography}
	
	
	\clearpage
	
	\appendix
	
	\addtocontents{toc}{\protect\setcounter{tocdepth}{2}}
	
	{
		\renewcommand{\contentsname}{Appendix}
		\tableofcontents
	}
	\vspace{1.5em}
	
	\section{Model Architectures and Algorithmic Implementations}\label{section:trainingDetails}
	
	\subsection{Continuous differentiable Poisson reparametrization}
	
	We used the continuous Poisson reparametrization algorithm given in \citet{vafaiiPoissonVariationalAutoencoder2024} coupled with a straight-through estimator to enable end-to-end training of the SSN. For the sake of completeness, the method has been reproduced here in Algorithm~\ref{alg:poissonSamplingAlgorithm}.
	
	\begin{algorithm}[htbp]
		\caption{Continuous Differentiable Poisson Reparametrization \citep{vafaiiPoissonVariationalAutoencoder2024}}
		\label{alg:poissonSamplingAlgorithm}
		\begin{algorithmic}[1]
			\REQUIRE Firing rates $\boldsymbol{\lambda}_{t} \in \mathbb{R}_{>0}^{B \times n}$, number of inter-arrival samples $n_{\text{exp}}$, relaxation temperature $\tau > 0$
			\ENSURE Differentiable spike count $\mathbf{z}_t \in \mathbb{R}_{+}^{B \times n}$
			\STATE Instantiate point-process inter-arrival distribution: $\operatorname{Exp} \leftarrow \operatorname{Exponential}(\boldsymbol{\lambda}_{t})$
			\STATE Sample reparametrized waiting times: $\Delta \mathbf{t} \leftarrow \operatorname{Exp.rsample}((n_{\text{exp}}, )) \quad$ \COMMENT{$\Delta \mathbf{t} \in \mathbb{R}_{+}^{n_{\text{exp}} \times B \times n}$}
			\STATE Compute cumulative event arrival times: $\mathbf{t}_{\text{arr}} \leftarrow \operatorname{cumsum}(\Delta \mathbf{t}, \, \text{dim} = 0)$
			\STATE Compute exact discrete spike count within unit window: $\mathbf{z}_{\text{hard}} \leftarrow \sum(\mathbb{I}(\mathbf{t}_{\text{arr}} \le 1), \, \text{dim} = 0)$
			\STATE Evaluate continuous sigmoidal arrival relaxation within unit window:
			\STATE \quad $\displaystyle \mathbf{I}_{\text{soft}} \leftarrow \sigma\left(\frac{1 - \mathbf{t}_{\text{arr}}}{\tau}\right) = \frac{1}{1 + \exp\left(-\frac{1 - \mathbf{t}_{\text{arr}}}{\tau}\right)}$
			\STATE Compute continuous relaxed spike counts: $\tilde{\mathbf{z}}_{t} \leftarrow \sum(\mathbf{I}_{\text{soft}}, \, \text{dim}=0)$
			\STATE Apply straight-through estimator: $\mathbf{z}_{t} \leftarrow \operatorname{stop\_grad}(\mathbf{z}_{\text{hard}} - \tilde{\mathbf{z}}_{t}) + \tilde{\mathbf{z}}_{t}$
			\STATE \textbf{return} $\mathbf{z}_{t}$
		\end{algorithmic}
	\end{algorithm}
	
	In an ideal Poisson process with intensity $\boldsymbol{\lambda}_{t}$, the inter-event intervals $\Delta \mathbf{t}$ follow an exponential distribution. The cumulative arrival time of the $k$-th spike is given by:
	\begin{equation}
		\mathbf{t}_{k} = \sum_{j = 1}^{k} \Delta \mathbf{t}_{j}.
	\end{equation}
	Yet, because the exact number of spikes occurring in a normalized unit interval $[0, 1]$ corresponds to the discrete step count $\mathbf{z}_{\text{hard}} = \sum_{k} \mathbb{I}(\mathbf{t}_{k} \le 1)$, \citet{vafaiiPoissonVariationalAutoencoder2024} replace the non-differentiable step indicator $\mathbb{I}(1 - \mathbf{t}_{k} \ge 0)$ with a smooth logistic sigmoid relaxed by temperature $\tau$:
	\begin{equation}
		\tilde{\mathbf{z}}_{t} = \sum_{k = 1}^{n_{\text{exp}}} \sigma\left( \frac{1 - \mathbf{t}_{k}}{\tau} \right) = \sum_{k = 1}^{n_{\text{exp}}} \frac{1}{1 + \exp\left(-\frac{1 - \mathbf{t}_{k}}{\tau}\right)}.
	\end{equation}
	As $\tau \to 0^{+}$, the sigmoid converges to the true discrete Poisson count. In our implementation, we combine $\mathbf{z}_{\text{hard}}$ and $\tilde{\mathbf{z}}_{t}$ via a straight-through estimator, $\mathbf{z}_{t} = \operatorname{stop\_grad}(\mathbf{z}_{\text{hard}} - \tilde{\mathbf{z}}_{t}) + \tilde{\mathbf{z}}_{t}$, so that exact discrete counts are propagated during the forward pass while path-wise gradients $\frac{\partial \tilde{\mathbf{z}}_{t}}{\partial \boldsymbol{\lambda}_{t}}$ flow through the unrolled recurrent membrane dynamics during backpropagation.
	
	Another gradient estimator has been proposed by \citet{ibrahimHitchhikersGuidePoisson2026}.
	
	\subsection{Baseline model architectures}\label{section:baselineModelArchitectures}
	
	All baseline models were equipped with identical receptive field filter lengths ($L_{\text{ker}}$), channel dimensions ($n$), and training parameters when applicable.
	
	\paragraph{Spiking Neural Network (SNN).} The standard leaky integrate-and-fire model shares the dual excitatory and inhibitory receptive field convolutions of the SSN:
	\begin{align}
		\mathbf{i}_{t} &= \mathbf{g}_{\text{exc}, t} - \mathbf{g}_{\text{inh}, t}, \\
		\mathbf{u}_{t} &= \sigma(\mathbf{g}_{\text{leak}}) \odot \mathbf{u}_{t - 1} \odot \left(1 - \mathbf{z}_{t - 1}^{\text{detached}}\right) + \mathbf{i}_{t} + \mathbf{u}_{\text{rest}}, \\
		\hat{\mathbf{u}}_{t} &= \mathbf{u}_{t} + \sum_{k = 1}^{L_{\text{hist}}} \mathbf{h}_{k} \odot \mathbf{z}_{t - k}, \\
		\mathbf{z}_{t} &= \Theta \left( \hat{\mathbf{u}}_{t} - \operatorname{softplus}(\boldsymbol{\vartheta}) + \boldsymbol{\xi}_{t} \right),
	\end{align}
	where $\mathbf{g}_{\text{leak}} \in \mathbb{R}^{n}$ is a learnable leak logit parameter, $\mathbf{u}_{\text{rest}} \in \mathbb{R}^{n}$ is the resting membrane bias, $\boldsymbol{\vartheta} \in \mathbb{R}^{n}$ is the unconstrained threshold parameter initialized to $0.5413$ such that $\operatorname{softplus}(\boldsymbol{\vartheta}) \approx 1.0$, and $\Theta(\cdot)$ denotes the Heaviside step function. Upon the moment of spiking, an instantaneous membrane reset is triggered via the hard reset term $(1 - \mathbf{z}_{t - 1}^{\text{detached}})$; this action is detached from backpropagation with the surrogate gradient in eq.~\eqref{eq:surrogateGradient} used instead. We also inject a small amount of threshold noise $\boldsymbol{\xi}_{t} \sim \mathcal{N}(\mathbf{0}, \sigma_{\text{noise}}^{2}\mathbf{I})$, where $\sigma_{\text{noise}} = 0.08$; this was chosen to maintain consistency with the stochastic models, improve training \cite{maExploitingNoiseResource2023, neftciSurrogateGradientLearning2019, bishopTrainingNoiseEquivalent1995}, receive benefit from using multiple Monte Carlo samples, and uphold better biological realism \cite{faisalNoiseNervousSystem2008, gerstnerNeuronalDynamicsSingle2014}.
	
	\paragraph{Adaptive Spiking Neural Network (ASNN).} The adaptive integrate-and-fire model augments the LIF formulation with an activity-dependent firing threshold $\boldsymbol{\vartheta}_{t} \in \mathbb{R}^{n}$ determined through an additional set of dynamics \cite{levakovaAdaptiveIntegrateandfireModel2019}:
	\begin{align}
		\boldsymbol{\vartheta}_{t} &= \operatorname{softplus}(\boldsymbol{\vartheta}_{0}) + \operatorname{softplus}(\boldsymbol{\gamma}_{a}) \odot \mathbf{a}_{t - 1}, \\
		\mathbf{z}_{t} &= \Theta \left( \hat{\mathbf{u}}_{t} - \boldsymbol{\vartheta}_{t} + \boldsymbol{\xi}_{t} \right), \\
		\mathbf{a}_{t} &= \sigma(\boldsymbol{\beta}_{a}) \odot \mathbf{a}_{t - 1} + \mathbf{z}_{t}.
	\end{align}
	The base threshold parameter $\boldsymbol{\vartheta}_{0} \in \mathbb{R}^{n}$ is initialized to $0.5413$ ($\operatorname{softplus}(\boldsymbol{\vartheta}_{0}) \approx 1.0$), the unconstrained threshold decay parameter $\boldsymbol{\beta}_{a} \in \mathbb{R}^{n}$ is initialized to $0.9$ (yielding decay factor $\sigma(\boldsymbol{\beta}_a) \approx 0.71$), and $\boldsymbol{\gamma}_{a} \in \mathbb{R}^{n}$ is a trainable adaptation coupling initialized to $0.5$. On the temporal reconstruction tasks, the spike history filter $\mathbf{h}$ is disabled for the ASNN so as to isolate its intrinsic threshold adaptation mechanism $\mathbf{a}_{t}$.
	
	\paragraph{Poisson Generalized Linear Model (GLM).} The Poisson GLM uses a single feedforward causal filter bank $\mathbf{k} \in \mathbb{R}^{n \times 1 \times L_{\text{ker}}}$ with bias $\mathbf{b} \in \mathbb{R}^{n}$ without dual inhibition:
	\begin{align}
		\mathbf{u}_{t} &= \left(\mathbf{k} *_{\text{causal}} \mathbf{x}\right)_{t} + \mathbf{b}, \\
		\hat{\mathbf{u}}_{t} &= \mathbf{u}_{t} + \sum_{k = 1}^{L_{\text{hist}}} \mathbf{h}_{k} \odot \mathbf{z}_{t - k}, \\
		\boldsymbol{\lambda}_{t} &= \exp\left( \operatorname{clamp}(\hat{\mathbf{u}}_{t}, \max = 15.0) \right), \\
		\mathbf{z}_{t} &= \Theta\left(\hat{\mathbf{u}}_{t} + \mathbf{G}_{t}\right),
	\end{align}
	where $\mathbf{G}_{t} = -\ln\left(-\ln(\mathbf{U}_{t} + \epsilon_{\text{G}}) + \epsilon_{\text{G}}\right)$ with $\mathbf{U}_{t} \sim \operatorname{Uniform}(0, 1)^{n}$ and $\epsilon_{\text{G}} = 10^{-20}$, providing Gumbel noise perturbations corresponding to stochastic binary Poisson threshold crossing with firing probability $\Pr(\mathbf{z}_{t} = 1 \mid \hat{\mathbf{u}}_{t}) = 1 - \exp(-e^{\hat{\mathbf{u}}_{t}}) = 1 - e^{-\boldsymbol{\lambda}_{t}}$. To be comparable with the SSN, it too was equipped with a history filter.
	
	\subsection{Surrogate gradient descent}
	
	Only the SSN was trained using the variational ELBO objective and continuous Poisson relaxation; all other comparison models, i.e., the SNN, ASNN, and GLM, were trained via surrogate gradient descent \cite{zenkeSuperspikeSupervisedLearning2018, neftciSurrogateGradientLearning2019}. Specifically, the \emph{fast sigmoid} surrogate function was used during the backward pass:
	\begin{equation}\label{eq:surrogateGradient}
		\frac{\partial \mathbf{z}_{t}}{\partial \hat{\mathbf{u}}_{t}} \approx \frac{1}{\left( 1 + k \left| \hat{\mathbf{u}}_{t} - \tilde{\boldsymbol{\vartheta}}_{t} + \tilde{\boldsymbol{\xi}}_{t} \right| \right)^{2}},
	\end{equation}
	where $\tilde{\boldsymbol{\vartheta}}_{t}$ denotes the effective threshold (e.g., $\operatorname{softplus}(\boldsymbol{\vartheta})$ for the SNN, $\boldsymbol{\vartheta}_{t}$ for the ASNN, and $\mathbf{0}$ for the GLM), $\tilde{\boldsymbol{\xi}}_{t}$ denotes the corresponding stochastic perturbation ($\boldsymbol{\xi}_{t}$ for the SNN and ASNN, and $\mathbf{G}_{t}$ for the GLM), and the surrogate slope scale parameter is set to $k = 25.0$ as is standard in prior literature \cite{zenkeSuperspikeSupervisedLearning2018}.
	
	\subsection{Temporal Reconstruction Tasks}\label{section:temporalReconstructionTask}
	
	The temporal reconstruction benchmarks were the central method of assessment for all models.
	
	\paragraph{Synthetic wave generation.} The multi-frequency sinusoidal waveform was synthesized over $t \in [0, 2\pi]$ discretized into $T = 100$ steps according to:
	\begin{equation}
		x(t) = \sin\left(f t\right) + \cos\left(0.25 f t\right) + \epsilon(t), \quad \epsilon(t) \sim \mathcal{N}\left(0, \sigma_{\text{noise}}^{2}\right),
	\end{equation}
	with base frequency $f = 5.0$ and additive Gaussian noise $\epsilon(t)$. Training batches were drawn dynamically with batch size $B = 4$, sequence length $T = 100$ steps across $n = 8$ channels, and training noise standard deviation $\sigma_{\text{noise}} = 0.1$. For evaluation, held-out test splits of $B = 8$ (or $B = 16$ for statistical and noise robustness comparisons) sequences ($T = 100$, $n = 8$, $\sigma_{\text{noise}} = 0.1$) were generated with deterministic seed offset $s + 10\,000$.
	
	\paragraph{In vivo turbulent odor plume (PID).} For experimenting with real-world sensory processing, we used a photoionization detector (PID) recording of an odor plume sampled at $f_{s} = 100.0$~Hz collected by \citet{kimOlfactoryReceptorNeurons2023}. The continuous 40-second recording ($4\,000$ time points) was split chronologically:
	\begin{enumerate}
		\item The training split was taken to be the first $0\text{--}16$~s ($1\,600$ time points), mirroring \citet{kimOlfactoryReceptorNeurons2023}, and trained as an unbroken causal sequence of length $T = 1\,600$ steps ($B = 1$).
		\item The held-out testing split was the remaining $16\text{--}40$~s ($2\,400$ time points) evaluated as an unbroken, continuous $24$~s sequence.
	\end{enumerate}
	The single-channel PID concentration was broadcast across $n = 16$ channels representing an olfactory receptor neuron (ORN) population to mimic multiple ORNs being exposed to the same plume.
	
	\subsubsection{Temporal reconstruction training and implementation details}
	
	A total of $10$ random seeds were used to instantiate each model architecture:
	\begin{equation}\label{eq:experimentalSeeds}
		\mathcal{S} = \{42, 121, 741, 555, 9000, 2161, 31, 32, 302, 2154\},
	\end{equation}
	with SSN and baseline models sharing identical front-end settings (where applicable) per task. Additionally, we chose to clamp the SSN's number of exponential inter-arrival samples per time step to $n_{\text{exp}} = 1$, i.e., forming a binary point process with spike emission probability $1 - e^{-\lambda_{t, i}}$ (for which the Poisson KL-divergence in eq.~\eqref{eq:poissonKL} forms a strict mathematical upper bound on the binary Bernoulli KL-divergence via the data processing inequality, hence still preserving a valid ELBO lower bound), to maintain fairness with the SNN and ASNN baseline models. All networks were trained for $E = 1000$ epochs using the AdamW optimizer with a learning rate of $\eta = 2 \times 10^{-3}$ on the synthetic wave task and $\eta = 3 \times 10^{-3}$ on the PID plume task, alongside gradient norm clipping with a maximum norm of $1.0$. Weight decay was set to $1 \times 10^{-2}$ for synthetic benchmarks and $1 \times 10^{-4}$ for the odor plume, though the leak, bias, threshold, adaptation, link function, and decoder parameters for all models were excluded. A generalized loss function was considered for every model:
	\begin{equation}\label{eq:generalizedLossFunction}
		\mathcal{L}(\theta, \phi) = \operatorname{MSE}\left(\mathbf{x}, \hat{\mathbf{x}}\right) + \frac{\beta_{\text{KL}}}{n T} \mathcal{L}_{\text{aux}},
	\end{equation}
	where $\beta_{\text{KL}} = 1 \times 10^{-4}$. For the SSN, $\mathcal{L}_{\text{aux}} = \sum_{t = 1}^{T} D_{\text{KL}}\!\left(q_\phi(\mathbf{z}_t \mid \cdot) \,\|\, p(\mathbf{z}_{t})\right) = \sum_{t = 1}^{T} \sum_{i = 1}^{n} D_{\text{KL}}\!\left(q_\phi(z_{t, i} \mid \cdot) \,\|\, p(z_{t, i})\right)$, whereas for baseline models (SNN, ASNN, GLM), $\mathcal{L}_{\text{aux}} = \sum_{t = 1}^{T} \sum_{i = 1}^{n} z_{t, i}$ represents an auxiliary metabolic penalty on overall spike emissions (averaged over the batch dimension $B$). In this sense, $\beta_{\text{KL}}$ does not necessarily refer to an actual KL-divergence for the SNN, ASNN, or GLM baselines, but instead takes the role of an energetic cost meant to parallel that of the SSN. Both $\beta_{\text{KL}} = 1 \times 10^{-4}$ and the relaxation temperature $\tau = 0.25$ were held constant throughout training. The full breakdown of the experimental hyperparameters is summarized in Table~\ref{table:temporalReconstructionHyperparameters}.
	
	\begin{table}[htbp]
		\centering
		\caption{\textbf{Experimental parameters and hyperparameters for the temporal reconstruction tasks.}}
		\label{table:temporalReconstructionHyperparameters}
		\small
		\begin{tabular}{lcc}
			\toprule
			\textbf{Hyperparameter} & \textbf{Synthetic Wave} & \textbf{In Vivo PID Plume} \\
			\midrule
			Sampling Rate ($f_s$)                  & Arbitrary (unit step)   & $100.0$~Hz \\
			Total Sequence Length                  & 100 time steps          & $40.0$~s ($4\,000$ time steps) \\
			Train / Test Split                     & Held-out seed batch     & $0\text{--}16$~s ($1\,600$) / $16\text{--}40$~s ($2\,400$) \\
			Window Length ($W$) / Stride ($S$)     & Full ($100$) / N/A      & Full ($1\,600$) / N/A \\
			Number of Channels ($n$)               & 8                       & 16 ORNs \\
			Receptive Field Length ($L_{\text{ker}}$) & 10 time steps        & 60 time steps ($0.6\text{ s}$) \\
			Refractory Kernel Length ($L_{\text{hist}}$) & 10 time steps     & 30 time steps ($0.3\text{ s}$) \\
			Decoder Kernel Size ($K_{\text{dec}}$) & 1                       & 15 time steps ($0.15\text{ s}$) \\
			Prior Reference Rate ($\lambda_{\text{ref}}$) & 0.15             & 0.03 \\
			Optimizer                              & AdamW                   & AdamW \\
			Learning Rate ($\eta$)                 & $2 \times 10^{-3}$      & $3 \times 10^{-3}$ \\
			Weight Decay                           & $1 \times 10^{-2}$      & $1 \times 10^{-4}$ \\
			Batch Size ($B$)                       & 4                       & 1 \\
			Epochs ($E$)                           & 1000                    & 1000 \\
			KL Regularizer ($\beta_{\text{KL}}$)   & $1 \times 10^{-4}$      & $1 \times 10^{-4}$ \\
			Temperature ($\tau$)                   & $0.25$ (constant)       & $0.25$ (constant) \\
			\bottomrule
		\end{tabular}
	\end{table}
	
	\paragraph{Decoders and population-level readout.} In order to reconstruct a target signal from a given model's spiking latents, it was necessary to pass $\mathbf{z}_{t}$ through a readout decoder. To obtain the predicted reconstruction value, we chose to use a strictly causal one-dimensional convolutional filter:
	\begin{equation}\label{eq:causalDecoder}
		\hat{\mathbf{x}}_{t} = \sum_{k = 0}^{K_{\text{dec}} - 1} \mathbf{W}_{\text{dec}, k} \, \mathbf{z}_{t - k} + \mathbf{b}_{\text{dec}},
	\end{equation}
	where $\mathbf{z}_{t - k} \in \mathbb{R}^{n}$ is the population spike count vector emitted at time $t - k$, $\mathbf{W}_{\text{dec}, k} \in \mathbb{R}^{d \times n}$ represents the linear readout weight tensor associated with lag $k$, and $\mathbf{b}_{\text{dec}} \in \mathbb{R}^{d}$ is a trainable bias.
	
	\paragraph{Causal padding and weight initialization.} An explicit left-padding of $P = K_{\text{dec}} - 1$ was applied to the latent spike tensor, with no padding on the right, to ensure that the reconstructions were causal and dependent on current and previous spikes $\mathbf{z}_{\le t}$. The decoder weights and biases were initialized using standard Kaiming uniform initialization across all models and benchmark tasks. The history filter weights were initialized from a zero-mean Gaussian distribution $\mathbf{h} \sim \mathcal{N}(\mathbf{0}, L_{\text{hist}}^{-1}\mathbf{I})$. We also set the thresholds to begin from $0.5413$, which corresponds directly to $\operatorname{softplus}(0.5413) \approx 1$. The resting potential bias $\mathbf{u}_{\text{rest}}$ began at $\mathbf{0}$ for all models. For the ASNN, we initialized $\boldsymbol{\beta}_{a}$ and $\boldsymbol{\gamma}_{a}$ to $0.9$ and $0.5$, respectively.
	
	\paragraph{Spatio-temporal population pooling.} In the case of the turbulent PID odor plume, the physical stimulus is a singular one-dimensional concentration profile ($d = 1$) broadcast to an array of $n = 16$ olfactory receptor neurons. Because a downstream biological decoder (e.g., projection neurons in the insect antennal lobe) must pool convergent sensory input from multiple upstream ORNs over finite synaptic integration windows, the decoder was configured with $K_{\text{dec}} = 15$ causal filter taps ($150$~ms temporal integration span at $f_{s} = 100$~Hz):
	\begin{equation}
		\hat{x}_{t} = \sum_{i = 1}^{16} \sum_{k = 0}^{14} W_{i, k} \, z_{i, t - k} + b_{\text{dec}}, \quad \mathbf{W}_{\text{dec}} \in \mathbb{R}^{1 \times 16 \times 15}.
	\end{equation}
	This architecture simultaneously performs spatial pooling across the $16$ spiking ORN channels and causal temporal smoothing over the preceding $150$~ms of firing history to synthesize the single-channel continuous plume trace $\hat{x}_{t} \in \mathbb{R}^{1}$.
	
	\paragraph{Channel isolation in the encoder and convergence in the decoder.} It should be emphasized that, unlike the decoder, the encoder itself maintains absolutely zero lateral crosstalk; indeed, this is done deliberately so as to match the biology of primary ORNs, which do not communicate via direct synaptic connections. Only after sensory transduction at the antenna does the output of individual neurons converge and get read out at the population level by downstream antennal lobe circuits \cite{shepherdEvolutionaryMicrocircuitApproach2021, byrneOxfordHandbookInvertebrate2019}.
	
	\subsection{Additive noise robustness evaluation}\label{section:additiveNoiseRobustnessTask}
	
	To quantify noise robustness on the synthetic wave task, trained checkpoints were loaded in discrete inference mode and tested against synthetic wave inputs corrupted by increasing levels of additive white Gaussian noise:
	\begin{equation}
		\tilde{\mathbf{x}}_{t} = \mathbf{x}_{\text{clean}}(t) + \sigma_{\text{noise}} \boldsymbol{\xi}_t, \quad \boldsymbol{\xi}_{t} \sim \mathcal{N}(\mathbf{0}, \mathbf{I}), \quad \sigma_{\text{noise}} \in \{0.0, 0.05, 0.1, 0.2, 0.4, 0.6, 0.8\}.
	\end{equation}
	The reconstruction MSE relative to the uncorrupted target $\mathbf{x}_{\text{clean}}$ was measured across all 10 trained seeds per noise level with batch size $B = 16$. We recorded both the mean and standard error of the mean (SEM) for each architecture. The sample traces in Figure~\ref{fig:noise-robustness} were rendered at noise level $\sigma_{\text{noise}} = 0.4$ by ensembling reconstructed waveforms across seeds.
	
	\subsection{Locust olfactory motif reproduction}\label{section:olfactoryMotifTask}
	
	In order to demonstrate whether the SSN was capable of also bridging towards neuroscientific applications and serving as a proxy model of primary sensory transduction, we benchmarked the architecture on learning the physiological response motifs of biological olfactory receptor neurons (ORNs) directly from continuous stimulus--response recordings.
	
	\paragraph{Electrophysiological dataset.} We again returned to the work of \citet{kimOlfactoryReceptorNeurons2023}; this time, we utilized their continuous \textit{in vivo} electrophysiological neural activity recordings from 43 individual ORNs in the desert locust (\textit{Schistocerca americana}). The measurements were taken in response to the same PID plume concentration as in the temporal reconstruction experiments and were sampled at $f_{s} = 100\,\mathrm{Hz}$ ($\Delta t = 10\,\mathrm{ms}$) across a $T = 40.0\,\mathrm{s}$ duration. Then, the recording was chronologically partitioned into a training split spanning the first $16.0\,\mathrm{s}$ ($1\,600$ frames), extracted using sliding causal windows of length $W = 300$ steps ($3.0\,\mathrm{s}$) with a stride of $S = 5$ steps ($50\,\mathrm{ms}$), yielding 261 training windows sampled in shuffled mini-batches of size $B = 16$ with drop-last enabled.
	
	\subsubsection{ORN motif matching training and implementation details}
	
	Since an SSN with one channel is meant to stand in for a single ORN, we instantiated a 43-channel SSN with no cross-talk allowed; the same was done for each of the baseline models. Afterwards, the models were fed the odor PID stimulus, replicated 43 times across the channel dimension, and tasked with reconstructing the equally 43-dimensional, continuous multivariate firing rate trajectory in biological units of $\mathrm{Hz}$; in this manner, each per-channel filter front-end learns an individual ORN. To remain as close as possible to the original study, we used the same filter lengths as in the linear non-linear model from \citet{kimOlfactoryReceptorNeurons2023}, that is, $L_{\text{ker}} = 200$ steps ($2.0\,\mathrm{s}$) for both excitatory and inhibitory causal temporal filter banks in the SSN, SNN, and ASNN; \citet{kimOlfactoryReceptorNeurons2023} did not, however, include a refractory filter, so we opted for a biologically realistic length of $L_{\text{hist}} = 50$ steps ($0.5\,\mathrm{s}$) initialized to zero. For the SSN, rate link parameters were initialized homogeneously across channels to $\alpha_{i} = -3.5$, $\beta_{i} = 0.0$, and $\vartheta_{i} = 0.0$, with a resting prior reference rate of $\lambda_{\text{ref}} = 0.03$; the Poisson GLM similarly had its single feedforward filter ($L_{\text{ker}} = 200$) initialized with a constant bias of $-3.5$; for the SNN and ASNN baselines, the resting bias and threshold parameters were initialized to $u_{\text{rest}, i} = 0.5$ and $\vartheta_{i} = 1.0$; the history filters for all models were set to have $L_{\text{hist}} = 50$, initialized to zero.
	
	\paragraph{Monte Carlo sampling.} Due to this task involving rate predictions, a number of experimental adjustments from the temporal reconstruction experiments were made during training: specifically, we used $M = 5$ Monte Carlo samples per training step and increased the SSN inter-arrival exponential spike capacity from one to five. Averaging across the parallel stochastic forward passes provides lower gradient variance and a superior pathwise estimate of the underlying rate $\boldsymbol{\lambda}_{t}$ when backpropagating through predictions.
	
	\paragraph{Learning rate annealing and optimization.} Optimization was conducted using AdamW with an initial learning rate $\eta = 2 \times 10^{-3}$, scheduled via cosine annealing down to $\eta_{\mathrm{min}} = 1 \times 10^{-5}$. Weight decay of $1 \times 10^{-4}$ was applied exclusively to the convolutional and history filter kernels, while link function parameters, thresholds, membrane leak rates, biases, and decoder weights received no decay. The network was trained for $E = 50$ epochs as it was deemed experimentally sufficient; training for $E = 100$, $E = 250$, and $E = 1000$ did not provide a visible difference in the quality of the learned step-pulse responses for any of the models. We minimized the generalized loss function in eq.~\eqref{eq:generalizedLossFunction} against empirical firing rates ($\hat{\mathbf{y}}_{t} \in \mathbb{R}_{+}^{43}$ in $\mathrm{Hz}$), with $\beta_{\text{KL}} = 1 \times 10^{-4}$ and relaxation temperature $\tau = 0.25$. Lastly, we employed gradient norm clipping with a maximum norm of $5.0$.
	
	\paragraph{Rate decoding and continuous supervision.} Because supervision is conducted against continuous firing rates in $\mathrm{Hz}$, network spike emissions are mapped to instantaneous rates via a channel-wise Gaussian rate decoder ($\operatorname{groups} = 43$). The spike outputs during training are convolved with a fixed, normalized Gaussian smoothing filter ($\sigma_{\mathrm{ms}} = 20.0\,\mathrm{ms}$, corresponding to $\sigma_{\mathrm{steps}} = 2.0$ bins, truncated at radius $r = 6$ bins, yielding kernel size $K_{\mathrm{dec}} = 13$ frames) with replicate boundary padding:
	\begin{equation}\label{eq:gaussianDecoder}
		\hat{y}_{i, t} = \sum_{k = -r}^{r} w_{k} \, z_{i, t - k}, \quad \text{where} \quad w_{k} = f_{s} \cdot \frac{\exp\left(-\frac{k^{2}}{2\sigma_{\mathrm{steps}}^{2}}\right)}{\sum_{j = -r}^{r}\exp\left(-\frac{j^{2}}{2\sigma_{\mathrm{steps}}^{2}}\right)}.
	\end{equation}
	The decoder operates independently across channels, yielding continuous predicted rate estimates $\hat{\mathbf{y}}_{t} \in \mathbb{R}_{+}^{43}$ in biological $\mathrm{Hz}$ units. Given that the decoder is fixed, the burden to learn faithful ORN dynamics is placed entirely onto the encoder.
	
\begin{figure}[htbp]
		\centering
		\includegraphics[width=\textwidth]{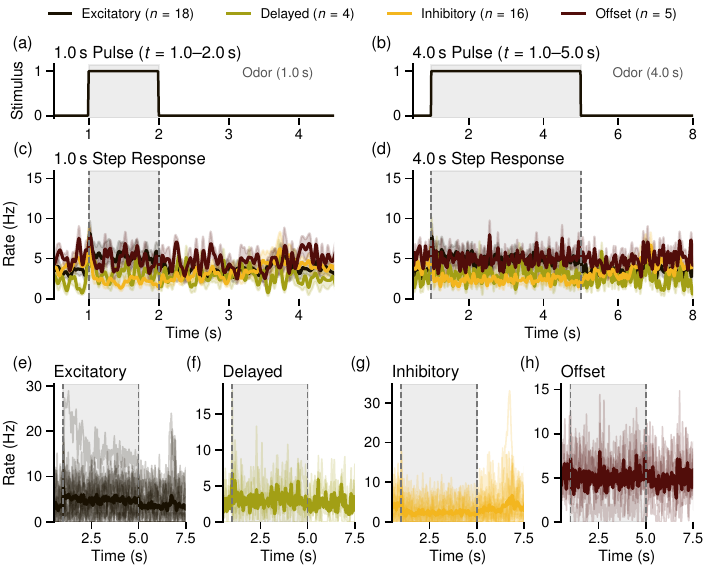}
		\caption{Average responses of trained SNNs after being given an artificial stimulus pulse. \textbf{(a)} Pulse of 1~s duration. \textbf{(b)} Pulse of 4~s duration. \textbf{(c)} Average response to the 1~s pulse. \textbf{(d)} Average response to the 4~s pulse. \textbf{(e)} Average response of the SNNs trained on an ORN classified to the excitatory motif. \textbf{(f)} Average response of the SNNs trained on an ORN classified to the delayed motif. \textbf{(g)} Average response of the SNNs trained on an ORN classified to the inhibitory motif. \textbf{(h)} Average response of the SNNs trained on an ORN classified to the offset motif.}
		\label{fig:snn-pulse-response}
	\end{figure}

\begin{figure}[htbp]
		\centering
		\includegraphics[width=\textwidth]{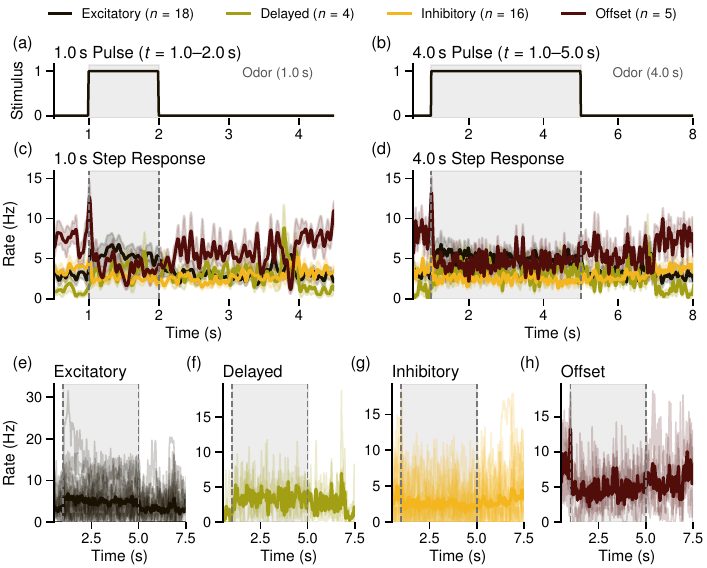}
		\caption{Average responses of trained ASNNs after being given an artificial stimulus pulse. \textbf{(a)} Pulse of 1~s duration. \textbf{(b)} Pulse of 4~s duration. \textbf{(c)} Average response to the 1~s pulse. \textbf{(d)} Average response to the 4~s pulse. \textbf{(e)} Average response of the ASNNs trained on an ORN classified to the excitatory motif. \textbf{(f)} Average response of the ASNNs trained on an ORN classified to the delayed motif. \textbf{(g)} Average response of the ASNNs trained on an ORN classified to the inhibitory motif. \textbf{(h)} Average response of the ASNNs trained on an ORN classified to the offset motif.}
		\label{fig:asnn-pulse-response}
	\end{figure}

\begin{figure}[htbp]
		\centering
		\includegraphics[width=\textwidth]{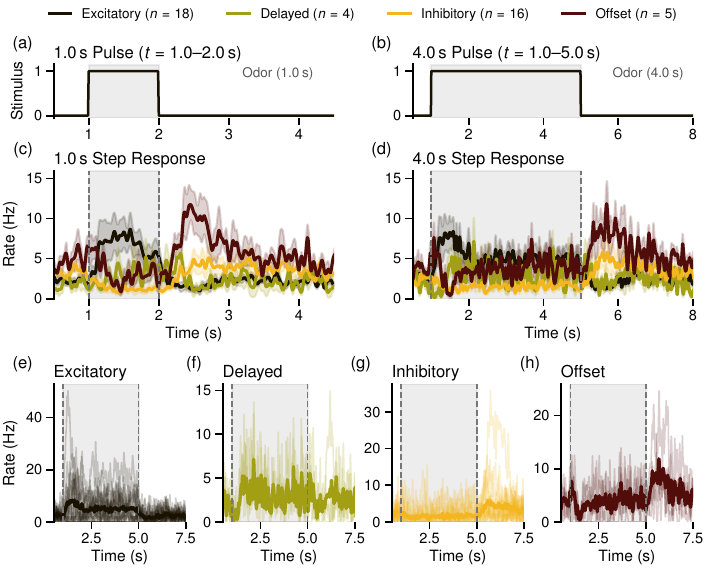}
		\caption{Average responses of trained GLMs after being given an artificial stimulus pulse. \textbf{(a)} Pulse of 1~s duration. \textbf{(b)} Pulse of 4~s duration. \textbf{(c)} Average response to the 1~s pulse. \textbf{(d)} Average response to the 4~s pulse. \textbf{(e)} Average response of the GLMs trained on an ORN classified to the excitatory motif. \textbf{(f)} Average response of the GLMs trained on an ORN classified to the delayed motif. \textbf{(g)} Average response of the GLMs trained on an ORN classified to the inhibitory motif. \textbf{(h)} Average response of the GLMs trained on an ORN classified to the offset motif. Much like the SSN, the GLM was capable of learning the necessary response motifs.}
		\label{fig:glm-pulse-response}
	\end{figure}

	\paragraph{Sustained step pulse assay.} The pre-trained models were subjected to square step pulse protocols without further parameter modification (Figure~\ref{fig:ssn-pulse-response}, Figure~\ref{fig:snn-pulse-response}, Figure~\ref{fig:asnn-pulse-response}, and Figure~\ref{fig:glm-pulse-response}). Once again matching the experiments done by \citet{kimOlfactoryReceptorNeurons2023}, two stimulus regimes were examined: a $1.0\,\mathrm{s}$ pulse ($5.0\,\mathrm{s}$ total sequence, $500$ frames at $100\,\mathrm{Hz}$, with unit odor concentration $x = 1.0$ during $t \in [1.0, 2.0)\,\mathrm{s}$) and a sustained $4.0\,\mathrm{s}$ pulse ($8.5\,\mathrm{s}$ total sequence, $850$ frames at $100\,\mathrm{Hz}$, with $x = 1.0$ during $t \in [1.0, 5.0)\,\mathrm{s}$). Model responses were generated across 10 evaluation trials and decoded into firing rates via the Gaussian rate filter.
	
	\subsection{Emergent spike-frequency adaptation evaluation}\label{section:adaptationTask}
	
	To evaluate whether spike-frequency adaptation (SFA) could emerge naturally from unsupervised stimulus reconstruction without explicit adaptation kinetics, we examined the models pre-trained on the turbulent PID odor plume task ($n = 16$, $L_{\text{ker}} = 60$, $L_{\text{hist}} = 30$, $\lambda_{\text{ref}} = 0.03$, $\tau = 0.25$). The ASNN was excluded from this comparison as it incorporates a hard-coded threshold adaptation state $\mathbf{a}_{t}$ by construction.
	
	\paragraph{Step stimulus protocol and PSTH computation.} Pre-trained SSN, SNN, and Poisson GLM checkpoints were loaded in evaluation mode and stimulated with a synthetic square step pulse of total duration $T = 150$ steps ($1.5\,\mathrm{s}$ at $f_{s} = 100\,\mathrm{Hz}$), with step onset at $t_{\text{on}} = 30$, step offset at $t_{\text{off}} = 115$, and amplitude $x_{t} = 0.80$ for $t \in [30, 115)$ (and $0.0$ otherwise), broadcast across all $n = 16$ channels and repeated for $M = 25$ Monte Carlo realizations per seed. Representative population spike rasters (Figure~\ref{fig:adaptation}b--d) were extracted from the first realization of the first seed ($s = 42$) by thresholding emitted spikes at $0.5$. To form the peri-stimulus time histograms (PSTHs, Figure~\ref{fig:adaptation}e) for each seed, spike realizations across the $M = 25$ trials were sampled from the inferred rate trajectories $\operatorname{Poisson}(\boldsymbol{\lambda}_{t})$ for the rate-parameterized models (SSN and GLM) and taken directly from the discrete binary spike emissions for the SNN. For each seed, the resulting $M \times n \times T$ spike tensor was averaged across trials and channels to obtain a mean population response $\bar{r}_{t} = \frac{1}{M n} \sum_{m = 1}^{M} \sum_{i = 1}^{n} z_{m, i, t}$, smoothed along the temporal dimension with a one-dimensional Gaussian filter to produce $\tilde{r}_{t}$, and min-max normalized to $[0, 1]$:
	\begin{equation}
		\hat{r}_{t} = \frac{\tilde{r}_{t} - \min_{\tau} \tilde{r}_{\tau}}{\max_{\tau} \tilde{r}_{\tau} - \min_{\tau} \tilde{r}_{\tau} + \epsilon},
	\end{equation}
	with $\epsilon = 10^{-8}$. The normalized trajectories $\hat{r}_{t}$ were then aggregated across the $10$ seeds to obtain the across-seed mean and SEM trajectories plotted in Figure~\ref{fig:adaptation}e.
	
	\subsection{Evaluation metrics and statistical testing}\label{section:evaluationSuite}
	
	\paragraph{Reconstruction fidelity.} The target sequences $\mathbf{x}$ and decoded reconstructions $\hat{\mathbf{x}}$ were evaluated using signal-to-noise ratio, measured in decibels ($\operatorname{SNR}_{\text{dB}}$), alongside the underlying Mean Squared Error (MSE), which served as the main distortion metric:
	\begin{align}
		\operatorname{MSE}(\mathbf{x}, \hat{\mathbf{x}}) &= \frac{1}{B d T} \sum_{b = 1}^{B} \sum_{i = 1}^{d} \sum_{t = 1}^{T} \left(x_{b, i, t} - \hat{x}_{b, i, t}\right)^{2}, \\
		\operatorname{SNR}_{\text{dB}}(\mathbf{x}, \hat{\mathbf{x}}) &= 10 \log_{10}\left( \frac{\frac{1}{B d T} \sum_{b, i, t} x_{b, i, t}^{2}}{\max\left( \operatorname{MSE}(\mathbf{x}, \hat{\mathbf{x}}), \, 10^{-12} \right)} \right).
	\end{align}
	
	\paragraph{Stimulus mutual information and channel entropy.} The mutual information (MI) between the continuous stimulus trajectory $X$ and emitted discrete spike counts $Z \in \{0, 1\}$ was estimated via a non-parametric $k$-nearest neighbours formulation of Kraskov, St\"ogbauer, and Grassberger (KSG) \cite{kraskovEstimatingMutualInformation2004} equipped with a continuous-to-discrete extension \cite{rossMutualInformationDiscrete2014}:
	\begin{equation}
		I(X; Z) = \psi(N_{\text{eval}}) - \frac{1}{N_{\text{eval}}}\sum_{z \in \{0, 1\}} N_{z} \psi(N_{z}) + \frac{1}{N_{\text{eval}}}\sum_{i = 1}^{N_{\text{eval}}} \left[ \psi(k) - \psi(k_{i}) \right],
	\end{equation}
	where $\psi(\cdot)$ is the digamma function, $N_{\text{eval}} = M \cdot B \cdot (T - \ell_{\max})$ denotes the total number of evaluation time points pooled across $M = 5$ Monte Carlo realizations and $B$ batch trials, $N_{z}$ is the count of samples emitting discrete state $z \in \{0, 1\}$, $k = 3$ is the number of nearest neighbours queried within the same discrete class $z_{i}$ to define the local stimulus radius $d_{i}$, and $k_{i}$ is the total number of points across the full evaluation set falling within radius $d_{i}$. The KSG estimator was provided by the Scikit-Learn library \cite{pedregosaScikitlearnMachineLearning}. Because the resulting MI from the estimator is computed in natural units (nats), it was converted to bits via:
	\begin{equation}
		I_{\text{bits}}(X; Z) = \frac{I_{\text{nats}}(X; Z)}{\ln 2} = \log_{2}(e) \cdot I_{\text{nats}}(X; Z).
	\end{equation}
	Yet, primary sensory receptor neurons integrate input over finite biophysical timescales; to resolve these conduction latencies and transmission delays without inductive bias, the continuous stimulus segment $\mathbf{x}_{:, c, :T - \ell_{\max}} \in \mathbb{R}^{N_{\text{eval}} \times 1}$ was systematically evaluated against causal spike offsets $\mathbf{z}_{:, c, \ell : T - \ell_{\max} + \ell} \in \{0, 1\}^{N_{\text{eval}}}$ across temporal integration lags $\ell \in \{0, 1, \dots, \ell_{\max} - 1\}$; here, the maximal lag bound was set to $\ell_{\max} = \min(L_{\text{lag}} + 1, \lfloor T/2 \rfloor)$, where $L_{\text{lag}} = \min(L_{\text{ker}}, \lfloor T/2 \rfloor)$ ($L_{\text{lag}} = 10$ for synthetic waves and $L_{\text{lag}} = 60$ for turbulent odor plumes). The channel-wise stimulus mutual information was then identified as the supremum over all queried lags:
	\begin{equation}
		I_{\text{stim}, c} = \max_{\ell \in [0, \ell_{\max} - 1]} I\left(\mathbf{x}_{:, c, :T - \ell_{\max}}; \, \mathbf{z}_{:, c, \ell : T - \ell_{\max} + \ell}\right),
	\end{equation}
	where channels exhibiting degenerate stimulus variance ($\sigma_{x, c} < 10^{-10}$) or insufficient active or inactive class counts ($\sum z < 5$ or $\sum (1 - z) < 5$) were skipped or set identically to zero; the overall population stimulus MI was subsequently taken as the unweighted mean across all channels:
	\begin{equation}
		I_{\text{stim}} = \frac{1}{n} \sum_{c = 1}^{n} I_{\text{stim}, c}.
	\end{equation}
	As a complement to stimulus transmission, we also considered the classical Shannon channel entropy, which quantified the baseline representational capacity of the neural population from the marginal spike event probabilities pooled across $N_{\text{tot}} = M \cdot B \cdot T$ evaluation samples; that is:
	\begin{equation}
		p_{1, i} = \operatorname{clip}\left(\frac{1}{N_{\text{tot}}} \sum_{m = 1}^{M} \sum_{b = 1}^{B} \sum_{t = 1}^{T} \mathbb{I}\left(z_{m, b, i, t} \ge 0.5\right), \, 10^{-12}, \, 1 - 10^{-12}\right), \quad p_{0, i} = 1 - p_{1, i}.
	\end{equation}
	The average channel entropy across the population is then given by:
	\begin{equation}
		H(Z) = -\frac{1}{n}\sum_{i = 1}^{n} \left( p_{0, i} \log_{2} p_{0, i} + p_{1, i} \log_{2} p_{1, i} \right).
	\end{equation}
	
	\paragraph{Population dynamics and representation diversity.} We employed two methods of quantifying the representational diversity of the latent spike trains: the first is via the Coefficient of Variation of inter-spike intervals, denoted $\mathrm{CV}_{\mathrm{ISI}}$, which is a standard metric within computational neuroscience used to assess the \emph{irregularity} of neural firing \cite{softkyHighlyIrregularFiring1993, gerstnerNeuronalDynamicsSingle2014}; the second method, introduced by \citet{gaoTheoryMultineuronalDimensionality2017} and referred to as the Participation Ratio ($\operatorname{PR}$), is a measure of representational dimensionality.
	
	The sample coefficient of variation with Bessel's correction ($\operatorname{ddof} = 1$) was computed according to:
	\begin{equation}\label{eq:coefficientOfVariation}
		\mathrm{CV}_{\mathrm{ISI}, i} = \frac{s_{\mathrm{ISI}, i}}{\bar{\Delta t}_{i}} = \frac{\sqrt{\frac{1}{M_{\text{isi}} - 1}\sum_{k = 1}^{M_{\text{isi}}} \left(\Delta t_{k} - \bar{\Delta t}_{i}\right)^{2}}}{\bar{\Delta t}_{i}},
	\end{equation}
	where $\Delta t_{k} = t_{k + 1} - t_{k}$ represents the $i$-th channel's inter-spike interval pooled from each of the trials; eq.~\eqref{eq:coefficientOfVariation} was subsequently averaged across all channels to obtain the population-level $\mathrm{CV}_{\mathrm{ISI}}$, though we imposed a restriction regarding low-activity so that only channels possessing at least $M_{\text{isi}} \ge 5$ of such pooled intervals were included in the average. We chose this criterion deliberately in an effort to prevent finite-sample bias.
	
	The PR is especially insightful as it allows for moving beyond single-channel statistics into the collective geometry of the neural code; specifically, it quantifies the effective linear dimensionality of the spiking population and is derived from the eigen-spectrum of the sample covariance matrix $\operatorname{Cov}[\mathbf{z}] \in \mathbb{R}^{n \times n}$:
	\begin{equation}
		\operatorname{PR} = \frac{\operatorname{Tr}\left( \operatorname{Cov}[\mathbf{z}] \right)^{2}}{\operatorname{Tr}\left( \operatorname{Cov}[\mathbf{z}]^{2} \right)} = \frac{\left( \sum_{i = 1}^{n} \lambda_{i} \right)^{2}}{\sum_{i = 1}^{n} \lambda_{i}^{2}},
	\end{equation}
	where $\{\lambda_{i}\}_{i = 1}^{n} \ge 0$ are the eigenvalues of $\operatorname{Cov}[\mathbf{z}]$. In this circumstance, we take $\mathbf{z} \in \{0, 1\}^{n \times N_{\text{tot}}}$ to be the population spike raster flattened across both trials and time.
	
	\paragraph{Spatial correlation.} To assess linear rate co-fluctuations across the population, we computed the so-called spatial correlation by first convolving each binary spike train with a one-dimensional Gaussian smoothing filter ($\sigma = 2.0$ time steps) along the temporal axis to obtain continuous rate traces $\tilde{\mathbf{r}} \in \mathbb{R}^{n \times N_{\text{tot}}}$, restricted to active channels with non-degenerate variance ($\sigma_{\tilde{r}, i} > 10^{-10}$, yielding $n_{\text{val}}$ valid channels). Afterwards, we evaluated the pairwise Pearson correlation matrix $\mathbf{R} \in [-1, 1]^{n_{\text{val}} \times n_{\text{val}}}$ and averaged the magnitude of its strictly upper-triangular off-diagonal elements:
	\begin{equation}
		\operatorname{Corr}_{\text{spatial}} = \frac{2}{n_{\text{val}}(n_{\text{val}} - 1)} \sum_{1 \le i < j \le n_{\text{val}}} \left| R_{i, j} \right|.
	\end{equation}
	
	\paragraph{Cross-channel mutual information.} A component of the information-maximization theory involves minimal redundancy \cite{barlowPossiblePrinciplesUnderlying2012, leeICAFeatureExtraction1998, meisterNeuralCodeRetina1999, bellIndependentComponentsNatural1997, fristonFreeenergyPrincipleUnified2010, chalkUnifiedTheoryEfficient2018}; hence, we assessed the level of mutual information carried between channels, evaluated across all distinct channel pairs $1 \le i < j \le n$ emitting at least two spikes each ($\sum_{t} z_{i, t} \ge 2$ and $\sum_{t} z_{j, t} \ge 2$). From the empirical joint probability distribution $\hat{p}(z_{i}, z_{j})$ over $N_{\text{tot}} = M \cdot B \cdot T$ pooled time steps, the plug-in discrete mutual information (in bits) was corrected for finite-sample bias via the Miller--Madow correction \cite{miller1955}:
	\begin{equation}\label{eq:crossChannelMI}
		I_{\text{cross}}(Z_{i}; Z_{j}) = \max\left( 0, \, \sum_{z_{i} \in \{0, 1\}} \sum_{z_{j} \in \{0, 1\}} \hat{p}(z_{i}, z_{j}) \log_{2}\!\left( \frac{\hat{p}(z_{i}, z_{j})}{\hat{p}(z_{i})\hat{p}(z_{j})} \right) - \frac{1}{2 N_{\text{tot}} \ln 2} \right),
	\end{equation}
	which was subsequently averaged across all valid channel pairs $(i, j)$.
	
	\paragraph{Coding efficiency.} The coding efficiency was defined to quantify the fraction of each channel pair's representational entropy that remains non-redundant. For all channel pairs $1 \le i < j \le n$ satisfying $\min\left(H(Z_{i}), H(Z_{j})\right) > 0.02$~bits, the pairwise efficiency was computed using the bias-corrected cross-channel mutual information from eq.~\eqref{eq:crossChannelMI} and averaged across the population:
	\begin{equation}
		\eta_{\text{code}} = \frac{1}{|\mathcal{P}_{\text{val}}|} \sum_{(i, j) \in \mathcal{P}_{\text{val}}} \operatorname{clip}\!\left( 1 - \frac{I_{\text{cross}}(Z_{i}; Z_{j})}{\min\left( H(Z_{i}), H(Z_{j}) \right)}, \, 0.0, \, 1.0 \right),
	\end{equation}
	where $\mathcal{P}_{\text{val}} = \left\{ (i, j) : 1 \le i < j \le n, \, \min\left(H(Z_{i}), H(Z_{j})\right) > 0.02 \right\}$.
	
	\paragraph{Statistical significance testing.} To ensure statistical rigour, all comparative benchmarks against the SSN baseline (encompassing SNN, ASNN, and Poisson GLM) were evaluated across the $n_{\text{seeds}} = 10$ seeds from eq.~\eqref{eq:experimentalSeeds}, where, for each seed $s \in \mathcal{S}$, evaluation was executed on identical pseudo-random test trajectories synthesized via deterministic seed offset $s + 10\,000$; paired differences were established as $d_{s} = M_{\text{model}}(s) - M_{\text{SSN}}(s)$. Effect sizes were formally quantified as absolute decibel shifts for Reconstruction SNR ($\Delta\operatorname{SNR} = \mu_{\text{model}} - \mu_{\text{SSN}}$) and relative percentage deviations for all remaining representation metrics ($\Delta\% = \frac{\mu_{\text{model}} - \mu_{\text{SSN}}}{|\mu_{\text{SSN}}|} \times 100\%$); in tandem, standardized paired effect magnitudes were indexed via Cohen's $d_{z}$ \cite{cohenStatisticalPowerAnalysis1988}:
	\begin{equation}
		d_{z} = \frac{\bar{d}}{s_d}, \quad \text{where} \quad \bar{d} = \frac{1}{n_{\text{seeds}}}\sum_{s = 1}^{n_{\text{seeds}}} d_{s}, \quad s_{d} = \sqrt{\frac{1}{n_{\text{seeds}} - 1}\sum_{s = 1}^{n_{\text{seeds}}} (d_{s} - \bar{d})^{2}},
	\end{equation}
	where $\bar{d}$ is the sample mean difference and $s_{d}$ is the sample standard deviation with Bessel's correction. Given that certain metrics, particularly those concerning information-theoretic measurements, do not typically adhere to assumptions of normality, the empirical difference distribution $\{d_{s}\}_{s = 1}^{n_{\text{seeds}}}$ was screened via the Shapiro--Wilk test \cite{shapiroAnalysisVarianceTest1965}; whenever the normality hypothesis was rejected ($p_{\text{Shapiro}} < 0.05$ for sample size $n_{\text{seeds}} \ge 8$), a non-parametric two-tailed Wilcoxon signed-rank test \cite{wilcoxonIndividualComparisonsRanking1945} was administered using Pratt's treatment for zero-difference ties \cite{prattRemarksZerosTies1959}. Conversely, when the differences conformed to normality ($p_{\text{Shapiro}} \ge 0.05$), we executed a standard paired two-tailed Student's $t$-test:
	\begin{equation}
		t = \frac{\bar{d}}{s_{d} / \sqrt{n_{\text{seeds}}}}.
	\end{equation}
	As a means to alleviate type~I error rates across the simultaneous metric and model evaluations for each task, raw $p$-values were adjusted using the Benjamini--Hochberg (BH) step-up procedure \cite{benjaminiControllingFalseDiscovery1995} to bound the False Discovery Rate (FDR) strictly at $\alpha = 0.05$:
	\begin{equation}
		p_{(k)}^{\text{adj}} = \min_{j \ge k} \left\{ \min\left( 1, \, \frac{m}{j} p_{(j)} \right) \right\},
	\end{equation}
	where $p_{(1)} \le p_{(2)} \le \dots \le p_{(m)}$ designate the ascending ranked raw $p$-values across all $m$ evaluated hypotheses for a given task. The pairwise differences were adjudicated as statistically significant if and only if $p^{\text{adj}} < 0.05$ ($^*p_{\text{adj}} < 0.05$, $^{**}p_{\text{adj}} < 0.01$, $^{***}p_{\text{adj}} < 0.001$; comparisons with $p_{\text{adj}} \ge 0.05$ are annotated as not significant, \textit{n.s.}). In Table~\ref{tab:statisticalComparison}, the reported $p$-values are the $p_{\text{adj}}$-values.
	
	\subsection{Hardware and training duration}
	
	Every experiment performed throughout was done on an Apple M4 Pro processor with 64~GB of RAM. Although individual model training is fast given the small number of parameters, running every experiment, for all model types, across $10$ seeds and $1000$ epochs, takes about 12 hours.
	
	\section{Ablation and Hyperparameter Experiments}\label{section:ablations}
	
	To explore the effects of hyperparameters on the performance of the model, we ran a varied suite of ablation experiments. We consider the noisy wave ($\sigma_{\text{noise}} = 0.1$) reconstruction benchmark under a set of standardized conditions ($n = 8$, $T = 100$, $L_{\text{ker}} = 10$, $L_{\text{hist}} = 10$, $\lambda_{\text{ref}} = 0.15$, $\beta_{\text{KL}} = 1 \times 10^{-4}$, $\tau = 0.25$, $\eta = 2 \times 10^{-3}$), specifically chosen because the task is simple and the model is known to perform well; consequently, any change or degradation in performance can be confidently isolated to having been the result of a parameter modification. All trials were evaluated across the same $10$ random seeds as in previous sections.
	
\begin{figure}[htbp]
		\centering
		\includegraphics[width=\textwidth]{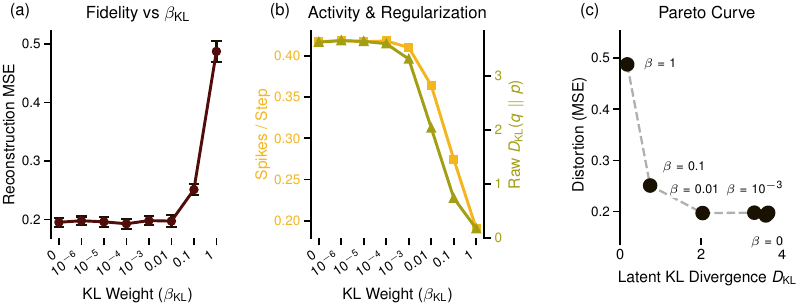}
		\caption{KL-divergence regularization weight sweep ($\beta_{\text{KL}}$). \textbf{(a)} The reconstruction MSE remains largely flat from $\beta_{\text{KL}} = 0$ to $0.01$ ($\operatorname{MSE} \approx 0.193\text{--}0.198$), before climbing significantly beginning from $\beta_{\text{KL}} = 0.1$ ($\operatorname{MSE} = 0.251$) to $\beta_{\text{KL}} = 1.0$ ($\operatorname{MSE} = 0.488$). \textbf{(b)} The population firing rate (yellow squares, left axis) and raw latent divergence $D_{\text{KL}}(q \,\|\, p)$ (olive triangles, right axis) remain elevated until $\beta_{\text{KL}} = 10^{-3}$, dropping steeply thereafter due to the significant penalty imposed by the prior reference rate. \textbf{(c)} The empirical Pareto frontier tracing signal distortion against metabolic cost.}
		\label{fig:ablationKLWeight}
	\end{figure}

	\paragraph{KL-divergence weight and rate-distortion trade-offs.} The strength of the KL-divergence regularizer $\beta_{\text{KL}}$ governs the penalty imposed when deviating from the prescribed homeostatic baseline. As such, it balances stimulus reconstruction fidelity and distance from the prior. We swept $\beta_{\text{KL}} \in \{0.0, 10^{-6}, 10^{-5}, 10^{-4}, 10^{-3}, 0.01, 0.1, 1.0\}$ to track the impact of various regularization strengths on reconstruction performance; the results are shown in Figure~\ref{fig:ablationKLWeight}, with all models being trained for $250$ epochs.
	
\begin{figure}[htbp]
		\centering
		\includegraphics[width=0.6\textwidth]{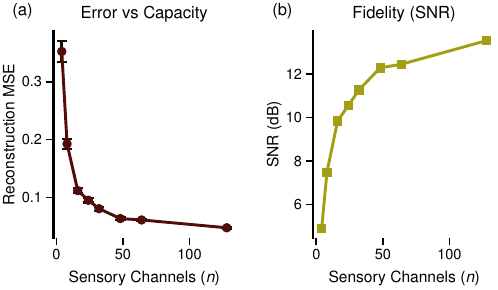}
		\caption{Reconstruction fidelity against sensory channel scaling. \textbf{(a)} The average MSE declines monotonically from $0.352$ ($n = 4$) down to $0.061$ ($n = 64$) and $0.048$ ($n = 128$). \textbf{(b)} The average reconstruction SNR rises steeply from $4.90$~dB ($n = 4$) to $11.27$~dB ($n = 32$), reaching $12.45$~dB ($n = 64$) and peaking at $13.54$~dB ($n = 128$).}
		\label{fig:ablationChannel}
	\end{figure}

	\paragraph{Sensory channel scaling and representational capacity.} We evaluated the effects of population size by sweeping channel counts $n \in \{4, 8, 16, 24, 32, 48, 64, 128\}$ trained for $250$ epochs. As expected, higher channel counts yielded better reconstruction fidelity while small populations lack the linear capacity to reconstruct multi-frequency waveforms ($n = 4$ yields $\operatorname{MSE} = 0.352$ and $\operatorname{SNR} = 4.90$~dB; $n = 8$ yields $\operatorname{MSE} = 0.193$ and $\operatorname{SNR} = 7.49$~dB). Expanding the population produces sharp performance gains through $n = 16$ ($\operatorname{MSE} = 0.112$, $\operatorname{SNR} = 9.85$~dB) and $n = 32$ ($\operatorname{MSE} = 0.081$, $\operatorname{SNR} = 11.27$~dB), continuing to improve monotonically across $n = 48$ ($\operatorname{MSE} = 0.064$, $\operatorname{SNR} = 12.28$~dB) and $n = 64$ ($\operatorname{MSE} = 0.061$, $\operatorname{SNR} = 12.45$~dB) to reach an empirical optimum at $n = 128$ ($\operatorname{MSE} = 0.048$, $\operatorname{SNR} = 13.54$~dB). The full visualization of the results is given in Figure~\ref{fig:ablationChannel}.
	
\begin{figure}[htbp]
		\centering
		\includegraphics[width=0.5\textwidth]{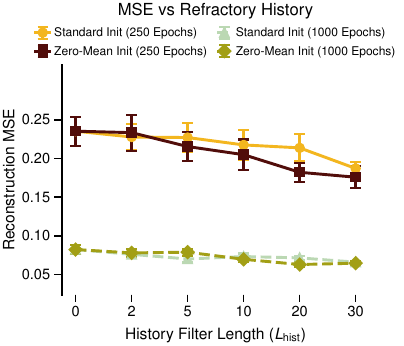}
		\caption{Spike history kernel length sweep ($L_{\text{hist}}$) taken with different history kernel parameter initializations. The average reconstruction MSE across trials decreases with increasing filter lengths, achieving an error of $0.176$ at $L_{\text{hist}} = 30$ ($0.182$ at $L_{\text{hist}} = 20$) for $E = 250$, and $0.063$ at $L_{\text{hist}} = 20$ ($0.064$ at $L_{\text{hist}} = 30$) for $E = 1000$ under zero-mean initialization (with standard initialization reaching $0.187$ at $E = 250$ and $0.066$ at $E = 1000$ for $L_{\text{hist}} = 30$). In contrast, omitting the history entirely ($L_{\text{hist}} = 0$) degrades mean error to $0.235$ at $E = 250$ (and $0.082$ at $E = 1000$).}
		\label{fig:ablationHistory}
	\end{figure}

	\paragraph{Influence of history kernel length.} To assess whether the activity history channel $\mathbf{h}$ provided any functional contribution to the results, we swept its history kernel length across $L_{\text{hist}} \in \{0, 2, 5, 10, 20, 30\}$, where $L_{\text{hist}} = 0$ corresponds to completely disabling self-spike feedback ($\mathbf{h} \equiv \mathbf{0}$). We trained the model for both $250$ and $1000$ epochs to evaluate convergence across short and extended schedules. As illustrated in Figure~\ref{fig:ablationHistory}, both the zero-mean (Gaussian) and standard (uniform) parameter initializations yield consistent performance improvements with increasing history length, showing no major difference between the two initialization schemes across epochs. Depending on the number of epochs used, an intermediate-to-long history length, i.e., $L_{\text{hist}} \ge 10$, provides an optimal trade-off, converging rapidly while maintaining low parameter counts and high fidelity. In general, the SSN is robust to the choice of $L_{\text{hist}}$ provided it spans sufficient refractory duration; as shown in Figure~\ref{fig:ablationHistory}, history kernel lengths of at least $5$ deliver strong performance.
	
\begin{figure}[htbp]
		\centering
		\includegraphics[width=0.55\textwidth]{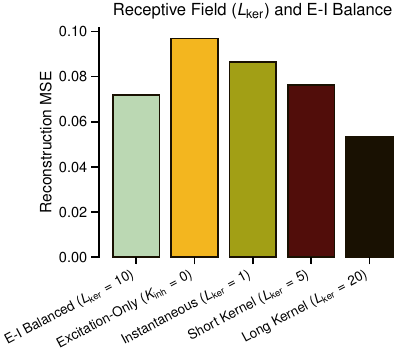}
		\caption{\textbf{Receptive field extent ($L_{\text{ker}}$) and E-I balance ablation.} Reconstruction MSE across architectural conditions. Disabling feedforward inhibition (Excitation-Only, yellow bar) severely impairs tracking ($\operatorname{MSE} = 0.097$ vs.\ $0.072$ for E-I balanced). Receptive field expansion monotonically improves performance, with $L_{\text{ker}} = 20$ achieving the lowest error ($\operatorname{MSE} = 0.053$, $\operatorname{SNR} = 13.04$~dB).}
		\label{fig:ablationKernelEI}
	\end{figure}

	\paragraph{Temporal receptive fields and excitatory-inhibitory balance.} We investigated the necessity of feedforward inhibition and temporal kernel length ($L_{\text{ker}}$) by evaluating five filter bank configurations: balanced excitation-inhibition ($L_{\text{ker}} = 10$), excitation-only ($\mathbf{k}_{\text{inh}} \equiv \mathbf{0}$), instantaneous ($L_{\text{ker}} = 1$), short kernel ($L_{\text{ker}} = 5$), and long kernel ($L_{\text{ker}} = 20$) at $E = 1000$ epochs. Disabling the inhibitory filter bank ($\mathbf{k}_{\text{inh}} \equiv \mathbf{0}$) triggers severe performance degradation: reconstruction MSE rises from $0.072$ to $0.097$, and SNR drops from $11.78$~dB to $10.51$~dB (Figure~\ref{fig:ablationKernelEI}). Without feedforward inhibition, positive stimulus drive accumulates continuously in the leaky membrane, driving the double-softplus rate link function into saturation. The dual E-I bank acts as an adaptive temporal differentiator, enabling the receptor neuron to encode stimulus transients and contrast without exhausting dynamic range. Expanding the temporal kernel monotonically improves tracking from $L_{\text{ker}} = 1$ ($\operatorname{MSE} = 0.087$, $\operatorname{SNR} = 10.95$~dB) and $L_{\text{ker}} = 5$ ($\operatorname{MSE} = 0.076$, $\operatorname{SNR} = 11.49$~dB) to $L_{\text{ker}} = 20$, reaching an MSE of $0.053$ ($\operatorname{SNR} = 13.04$~dB).
	
\begin{figure}[htbp]
		\centering
		\includegraphics[width=\textwidth]{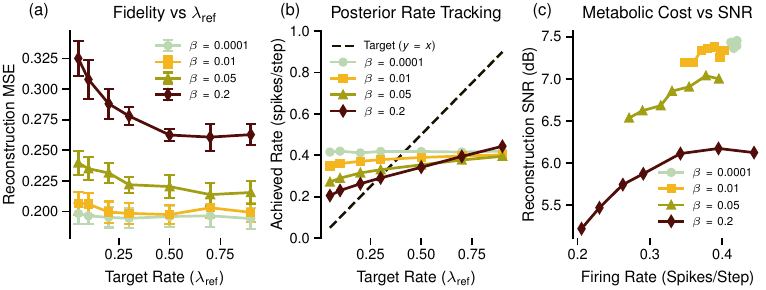}
		\caption{Prior reference rate sweep ($\lambda_{\text{ref}}$) across regularization weights ($\beta_{\text{KL}}$). \textbf{(a)} Reconstruction MSE remains flat across target rates for weak regularization ($\beta_{\text{KL}} = 10^{-4}$, $\operatorname{MSE} \approx 0.194\text{--}0.198$; $\beta_{\text{KL}} = 0.01$, $\operatorname{MSE} \approx 0.198\text{--}0.206$). At $\beta_{\text{KL}} = 0.2$, low reference rates penalize firing, increasing MSE from $0.263$ ($\lambda_{\text{ref}} = 0.9$) to $0.325$ ($\lambda_{\text{ref}} = 0.05$). \textbf{(b)} Posterior rate tracking. At low $\beta_{\text{KL}}$ ($10^{-4}$ and $0.01$), achieved firing rates remain largely decoupled from the prior ($\approx 0.35\text{--}0.42$ spikes/step). As $\beta_{\text{KL}}$ increases to $0.2$, achieved rates rotate toward the identity diagonal ($y = x$, dashed line). \textbf{(c)} Metabolic cost vs.\ SNR curves, illustrating the trade-off between mean spike count and reconstruction fidelity.}
		\label{fig:ablationPriorRate}
	\end{figure}

	\paragraph{Prior reference rate and posterior rate steering.} The influence of the homeostatic prior $p(\mathbf{z}_{t}) = \operatorname{Poisson}(\mathbf{z}_{t}; \boldsymbol{\lambda}_{\text{ref}})$ on overall population firing was assessed by sweeping target rates $\lambda_{\text{ref}} \in \{0.05, 0.1, 0.2, 0.3, 0.5, 0.7, 0.9\}$ across four regularization tiers: $\beta_{\text{KL}} \in \{10^{-4}, 0.01, 0.05, 0.2\}$. As shown in Figure~\ref{fig:ablationPriorRate}b, under weak regularization ($\beta_{\text{KL}} = 10^{-4}$), achieved firing rates remain nearly constant ($\approx 0.412\text{--}0.420$ spikes/step), showing independence from $\lambda_{\text{ref}}$. In this regime, the network prioritizes signal tracking over prior alignment, achieving optimal SNRs ($\approx 7.36\text{--}7.45\text{ dB}$, $\operatorname{MSE} \approx 0.194\text{--}0.198$, Figure~\ref{fig:ablationPriorRate}a,c). Under intermediate regularization ($\beta_{\text{KL}} = 0.01$), firing rates adjust moderately ($0.348\text{--}0.402$ spikes/step), with SNR remaining high ($7.19\text{--}7.38\text{ dB}$). As $\beta_{\text{KL}}$ increases to $0.05$ and $0.2$, the achieved firing rate progressively rotates toward the identity diagonal ($y = x$). At $\beta_{\text{KL}} = 0.2$, setting a low prior ($\lambda_{\text{ref}} = 0.05$) constrains the achieved rate to $0.205$ spikes/step, reducing SNR to $5.22\text{ dB}$ and increasing MSE to $0.325$ (Figure~\ref{fig:ablationPriorRate}a,c). This confirms that the Poisson KL term functions as an effective, tunable metabolic constraint.
	
\begin{figure}[htbp]
		\centering
		\includegraphics[width=0.85\textwidth]{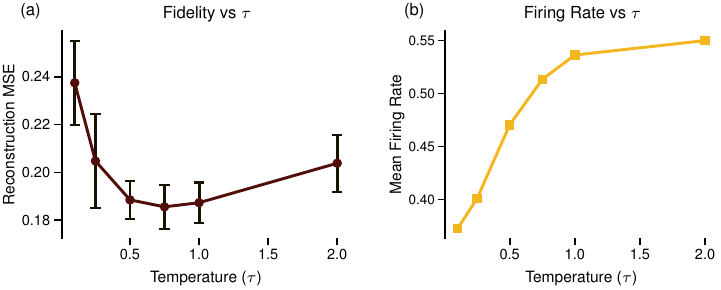}
		\caption{\textbf{Continuous Poisson relaxation temperature sweep ($\tau$).} \textbf{(a)} Reconstruction MSE across $\tau \in [0.1, 2.0]$ at $\beta_{\text{KL}} = 10^{-4}$. Optimal performance occurs around $\tau \in [0.5, 1.0]$. \textbf{(b)} Mean population firing rate as a function of temperature.}
		\label{fig:ablationTemperature}
	\end{figure}

	\paragraph{Continuous Poisson relaxation temperature and pathwise gradient estimation.} The continuous relaxation temperature $\tau > 0$ governs the sharpness of the sigmoidal event-arrival approximation in the straight-through Poisson reparametrization (Algorithm~\ref{alg:poissonSamplingAlgorithm}). As $\tau \to 0^{+}$, the relaxed soft arrival indicator $\sigma\left(\frac{1 - \mathbf{t}_{\text{arr}}}{\tau}\right)$ converges to the discrete Heaviside step function, while larger values of $\tau$ expand the non-zero gradient window $\frac{\partial \tilde{\mathbf{z}}_{t}}{\partial \boldsymbol{\lambda}_{t}}$. To characterize the sensitivity of the unrolled membrane dynamics to this smoothing parameter, we evaluated $\tau \in \{0.1, 0.25, 0.5, 0.75, 1.0, 2.0\}$. In the cold-temperature regime ($\tau = 0.1$), the weak gradient impedes backpropagation through event times, which subsequently manifests as an elevation in reconstruction distortion ($\operatorname{MSE} = 0.237 \pm 0.018$, $\operatorname{SNR} = 6.65 \pm 0.30$~dB). Moderately warming the temperature smooths the optimization landscape, driving a monotonic reduction in error through $\tau = 0.25$ ($\operatorname{MSE} = 0.205 \pm 0.020$, $\operatorname{SNR} = 7.34 \pm 0.35$~dB) to an optimal plateau across $\tau \in [0.5, 1.0]$. Excessively high temperatures ($\tau = 2.0$) lead to the relaxation becoming overly diffuse, mildly degrading performance ($\operatorname{MSE} = 0.204 \pm 0.012$, $\operatorname{SNR} = 7.28 \pm 0.25$~dB). Concurrently, the mean population firing rate scales monotonically with temperature (Figure~\ref{fig:ablationTemperature}b), rising from $0.373 \pm 0.015$ spikes/step at $\tau = 0.1$ to $0.513 \pm 0.020$ spikes/step at $\tau = 0.75$ and plateauing at $0.550 \pm 0.017$ spikes/step at $\tau = 2.0$. Therefore, our choice of default baseline setting of $\tau = 0.25$ sits appropriately within the ideal range.
	
\begin{figure}[htbp]
		\centering
		\begin{minipage}[t]{0.32\textwidth}
			\centering
			\includegraphics[width=\linewidth]{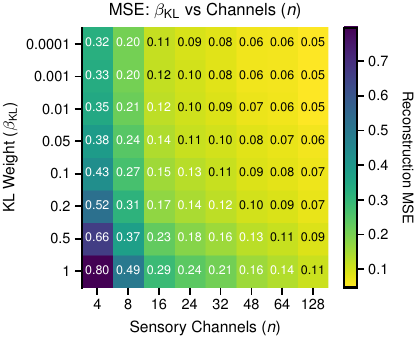}
		\end{minipage}\hfill
		\begin{minipage}[t]{0.32\textwidth}
			\centering
			\includegraphics[width=\linewidth]{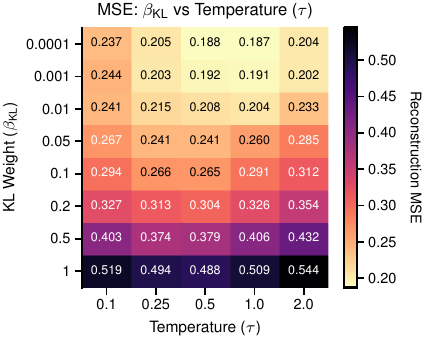}
		\end{minipage}\hfill
		\begin{minipage}[t]{0.32\textwidth}
			\centering
			\includegraphics[width=\linewidth]{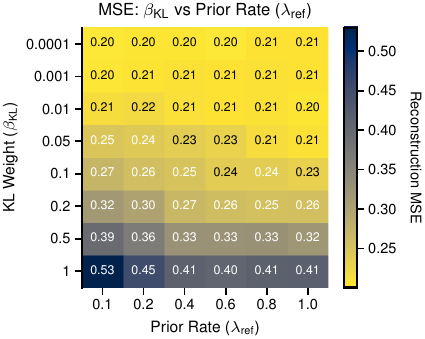}
		\end{minipage}
		\caption{Two-dimensional hyperparameter interaction heatmaps on reconstruction MSE. \textbf{(left)} $\beta_{\text{KL}}$ vs.\ Sensory Channels ($n$). Monotonic capacity improvements appear across all regularization strengths. \textbf{(middle)} $\beta_{\text{KL}}$ vs.\ Temperature ($\tau$). Evaluation MSE decreases with warmer relaxation temperatures up to $\tau = 1.0$. \textbf{(right)} $\beta_{\text{KL}}$ vs.\ Prior Rate ($\lambda_{\text{ref}}$). Reconstruction error remains stable ($\operatorname{MSE} \approx 0.20\text{--}0.22$) for all $\beta_{\text{KL}} \le 0.01$ across all target rates.}
		\label{fig:ablationHeatmap}
	\end{figure}

	\paragraph{$\beta_{\text{KL}}$ vs.\ Channels ($n$).} Across all channel dimensions ($n \in [4, 128]$), increasing population size monotonically reduces reconstruction error (Figure~\ref{fig:ablationHeatmap}, left). At baseline $\beta_{\text{KL}} = 10^{-4}$, MSE declines smoothly from $0.323$ ($n = 4$) to $0.205$ ($n = 8$), $0.114$ ($n = 16$), $0.081$ ($n = 32$), $0.060$ ($n = 64$), and $0.048$ ($n = 128$). Even under severe metabolic penalties ($\beta_{\text{KL}} = 0.5$), expanding channels reduces MSE from $0.660$ to $0.088$, confirming that population expansion compensates effectively for heavy individual rate penalties.
	
	\paragraph{$\beta_{\text{KL}}$ vs.\ Temperature ($\tau$).} The evaluation error decreases as $\tau$ increases up to $\tau \in [0.5, 1.0]$ (at $\beta_{\text{KL}} = 10^{-4}$, MSE declines from $0.237$ at $\tau = 0.1$ to $0.205$ at $\tau = 0.25$, $0.188$ at $\tau = 0.5$, and $0.187$ at $\tau = 1.0$, rising slightly to $0.204$ at $\tau = 2.0$). Intermediate-to-warm temperatures produce smoother surrogate arrival gradients that facilitate optimization. The landscape remains stable along the vertical axis for $\beta_{\text{KL}} \le 0.01$ (Figure~\ref{fig:ablationHeatmap}, middle).
	
	\paragraph{Prior Rate ($\lambda_{\text{ref}}$) vs.\ $\beta_{\text{KL}}$.} As shown in Figure~\ref{fig:ablationHeatmap} (right), for all $\beta_{\text{KL}} \le 0.01$, reconstruction error is virtually invariant to the reference rate ($\operatorname{MSE} \approx 0.201\text{--}0.217$ across all $\lambda_{\text{ref}} \in [0.1, 1.0]$). Only when regularization becomes strict ($\beta_{\text{KL}} = 0.2$) does lower baseline activity increase error (from $0.255$ at $\lambda_{\text{ref}} = 1.0$ and $0.254$ at $\lambda_{\text{ref}} = 0.8$ up to $0.315$ at $\lambda_{\text{ref}} = 0.1$). This establishes that the SSN functions reliably across diverse physiological setpoints without hyperparameter sensitivity cliffs.
	
	\section{Supplementary Discussion}
	
	\begin{wrapfigure}[17]{r}{0.5\textwidth}
		\centering
		\vspace{-10pt}
		\includegraphics[width=0.5\textwidth]{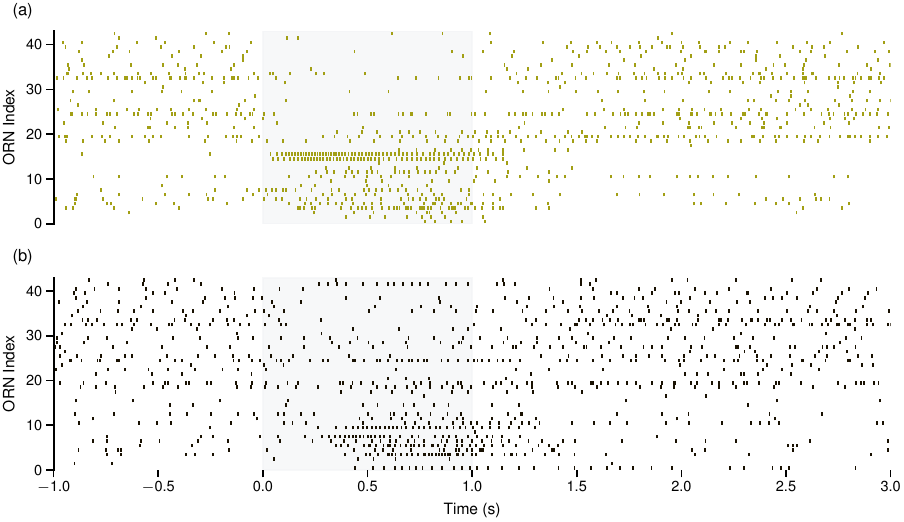}
		\caption{Empirical ORN spike raster data from \citet{kimOlfactoryReceptorNeurons2023}. \textbf{(a)} ORN response to an exposure of Hexanol. \textbf{(b)} ORN response to an exposure of Cyclohexanol.}
		\label{fig:orn-spike-raster}
	\end{wrapfigure}
	
	\paragraph{Considerations for sparsity.} It can be observed that the SSN produces denser spikes than the usual leaky-integrate-and-fire SNN (Figure~\ref{fig:adaptation}). Certainly, this does, in part, depend on both the prior reference rate $\lambda_{\text{ref}}$ as well as the strength of its enforcement (i.e., via the KL-divergence $\beta_{\text{KL}}$ parameter). However, for a model of first-stage sensory neurons, higher spike density is correct. In fact, most sensory neurons are \emph{not} sparse; instead, they typically fire a salvo of spikes in response to finely-tuned receptive fields \cite{kimOlfactoryReceptorNeurons2023, gorur-shandilyaOlfactoryReceptorNeurons2017}. Sparsity, in contrast, is a property that occurs later in downstream higher-order parts of the brain \cite{perez-oriveOscillationsSparseningOdor2002, laurentOlfactoryNetworkDynamics2002}. We provide visualizations for some of the true spike rasters from the exact ORNs from \citet{kimOlfactoryReceptorNeurons2023} in Figure~\ref{fig:orn-spike-raster}.
	
	\paragraph{Choice of Poisson distribution.} The choice of using a Poisson distribution for both the prior $p(\mathbf{z}_{t})$ and approximate posterior $q_{\phi}(\mathbf{z}_{t} \mid \mathbf{x}_{\le t}, \mathbf{z}_{<t})$ comes from a long-established history of neuroscience \cite{softkyHighlyIrregularFiring1993, shadlenVariableDischargeCortical1998, paninskiMaximumLikelihoodEstimation2004, pillowSpatiotemporalCorrelationsVisual2008, gerstnerNeuronalDynamicsSingle2014}. Indeed, it has been well-known that first-stage sensory neurons can be sufficiently modelled by Poissonian-type models; further still, many cortical models can exhibit such firing statistics. Recent work has also demonstrated success in the reverse task of inferring latents \emph{from} already-recorded biological spiking activity, with these models directly imposing Poisson inductive biases on the population activity \cite{pandarinathInferringSingletrialNeural2018, zhouLearningIdentifiableInterpretable2020}. In any case, our model carries a dynamically changing rate $\boldsymbol{\lambda}_{t}$ per time step $t$, so all that is necessarily imposed is that within each discretized time step bucket of duration $\Delta t$, the neural firing within that bucket carries a Poisson distribution with rate $\boldsymbol{\lambda}_{t}$; although most neurons, sensory neurons included, are known to be fairly precise in their spike-times, there is still an allowable window accounting for stochastic jitter usually in the order of a few milliseconds \cite{bairTemporalPrecisionSpike1996, jolivetPredictingSpikeTiming2006, toupsFindingEventStructure2011}. Thus, we consider a model having sufficiently granular discretization as appropriate. A good exposition on the Poisson matter is provided in \cite{vafaiiPoissonVariationalAutoencoder2024, vafaiiBrainlikeVariationalInference2025}. 
	
	Nevertheless, the Poisson distribution for neural spike modelling has some drawbacks; most markedly, a Poisson distribution, by definition, imposes equal variance and mean---in practice, this is often not satisfied in real neurons; instead, empirical spiking activity undergoes periods of overdispersion or underdispersion, wherein a given neuron's Fano factor (the ratio between the variance and the mean, which is unity for a true Poisson distribution) changes dynamically. Interestingly, the Negative Binomial distribution has been proposed as an alternative, possibly more suitably apt firing distribution candidate precisely because it is capable of capturing this dispersion \cite{zhangNegativeBinomialVariational, ghanbariModelingStimulusdependentVariability2019}. Future work is encouraged to explore flavours of the SSN outfitted with a Negative Binomial distribution in lieu of the Poisson distribution used here. A reparametrization trick for the Negative Binomial distribution is given by \citet{zhangNegativeBinomialVariational}.
	
	\paragraph{Relation to previous work.} The SSN sits as a mixture between variational Bayesian theory, traditional non-linear GLM models, and canonical leaky-integrate neuron dynamics. The model carries elements of similarity with previous work by \citet{kamataFullySpikingVariational2022}, who also introduced a recurrent VAE formulation, relying on a leaky-integrate integration backbone, and using a Bernoulli distribution per time step; indeed, when the Poisson distribution has its support conditioned on at most one spike per time step, i.e., $z \in \{0, 1\}$, it becomes equivalent to a Bernoulli distribution with probability of success $p = \frac{\lambda}{1 + \lambda}$ (or $p = 1 - e^{-\lambda}$ when spike counts $z \ge 1$ are clamped to $1$). Therefore, in this context, the recurrent spiking architecture of \citet{kamataFullySpikingVariational2022} can be viewed as conceptually corresponding to the binary limiting case of our SSN formulation, wherein single-spike Bernoulli events replace count intensities. Additionally, \citet{kamataFullySpikingVariational2022}, much like \citet{vafaiiPoissonVariationalAutoencoder2024}, only evaluated their model on static image reconstruction tasks; it was not tested on time-varying stimuli or discussed within a broader biological context. The concerns of the SSN are skewed towards the latter end, as it attempts to fill this gap.
	
	\section{Theoretical links between variational inference and biology}
	
	\subsection{Theoretical considerations for stimulus reconstruction}\label{section:stimulusReconstructionTheory}
	
	Here we detail the step-by-step derivation connecting the continuous trajectory mutual information $I(\mathbf{x}_{1:T}; \mathbf{z}_{1:T})$ to the Mean Squared Error ($\mathrm{MSE}$) decoding objective minimized by the recurrent VAE framework, as given in eq.~\eqref{eq:variationalMSEBound}.
	
	\paragraph{Variational lower bound on trajectory mutual information.} Let $\mathbf{x} \equiv \mathbf{x}_{1:T} \in \mathbb{R}^{d \times T}$ denote the continuous sensory stimulus sequence of length $T$ across $d$ input channels, and let $\mathbf{z} \equiv \mathbf{z}_{1:T} \in \mathbb{N}_{0}^{n \times T}$ denote the elicited latent spike sequence generated by the sensory encoder $q_{\phi}(\mathbf{z} \mid \mathbf{x})$. By definition, the trajectory mutual information is:
	\begin{align}
		I(\mathbf{x}; \mathbf{z}) &= h(\mathbf{x}) - h(\mathbf{x} \mid \mathbf{z}) \nonumber \\
		&= h(\mathbf{x}) + \mathbb{E}_{p^*(\mathbf{x}) q_{\phi}(\mathbf{z} \mid \mathbf{x})}\!\left[ \ln p(\mathbf{x} \mid \mathbf{z}) \right],
	\end{align}
	where $h(\mathbf{x}) = -\int p^*(\mathbf{x}) \ln p^*(\mathbf{x})\,\mathrm{d}\mathbf{x}$ is the differential entropy of the environmental stimulus process, and $p(\mathbf{x} \mid \mathbf{z})$ represents the true, intractable posterior decoding distribution.
	
	Let $p_{\theta}(\mathbf{x} \mid \mathbf{z})$ be an arbitrary parametrized variational decoding distribution. By the non-negativity of the Kullback--Leibler divergence $D_{\mathrm{KL}}(p(\mathbf{x} \mid \mathbf{z}) \,\|\, p_{\theta}(\mathbf{x} \mid \mathbf{z})) \ge 0$, we have:
	\begin{equation}
		\int p(\mathbf{x} \mid \mathbf{z}) \ln p(\mathbf{x} \mid \mathbf{z})\,\mathrm{d}\mathbf{x} \ge \int p(\mathbf{x} \mid \mathbf{z}) \ln p_{\theta}(\mathbf{x} \mid \mathbf{z})\,\mathrm{d}\mathbf{x}.
	\end{equation}
	Taking expectations over the joint distribution $p(\mathbf{x}, \mathbf{z}) = p^*(\mathbf{x}) q_{\phi}(\mathbf{z} \mid \mathbf{x})$ establishes the classical Barber--Agakov variational lower bound \cite{barberIMAlgorithmVariational}:
	\begin{equation}\label{eq:barberAgakovBound}
		I(\mathbf{x}; \mathbf{z}) \ge h(\mathbf{x}) + \mathbb{E}_{p^*(\mathbf{x}) q_{\phi}(\mathbf{z} \mid \mathbf{x})}\!\left[ \ln p_{\theta}(\mathbf{x} \mid \mathbf{z}) \right].
	\end{equation}
	
	\paragraph{Gaussian decoding likelihood and mean squared distortion.} We assume an isotropic Gaussian observation likelihood with parametrized conditional mean $\hat{\mathbf{x}}_{\theta}(\mathbf{z}) \in \mathbb{R}^{d \times T}$ and isotropic variance $\sigma^{2} \mathbf{I}_{dT}$:
	\begin{equation}
		p_{\theta}(\mathbf{x} \mid \mathbf{z}) = \left(2 \pi \sigma^{2}\right)^{-\frac{dT}{2}} \exp\left( -\frac{1}{2 \sigma^{2}} \|\mathbf{x} - \hat{\mathbf{x}}_{\theta}(\mathbf{z})\|_{2}^{2} \right).
	\end{equation}
	Evaluating the expected log-likelihood under this density yields:
	\begin{equation}
		\mathbb{E}_{p^*(\mathbf{x}) q_{\phi}(\mathbf{z} \mid \mathbf{x})}\!\left[ \ln p_{\theta}(\mathbf{x} \mid \mathbf{z}) \right] = -\frac{dT}{2} \ln(2 \pi \sigma^{2}) - \frac{1}{2\sigma^{2}} \mathbb{E}_{p^*(\mathbf{x}) q_{\phi}(\mathbf{z} \mid \mathbf{x})}\!\left[ \|\mathbf{x} - \hat{\mathbf{x}}_{\theta}(\mathbf{z})\|_{2}^{2} \right].
	\end{equation}
	Defining the expected per-element mean squared reconstruction error as $\mathrm{MSE}(\phi, \theta) \equiv \frac{1}{dT}\mathbb{E}[\|\mathbf{x} - \hat{\mathbf{x}}_\theta(\mathbf{z})\|_{2}^{2}]$, the variational bound in eq.~\eqref{eq:barberAgakovBound} becomes:
	\begin{equation}\label{eq:miGaussianBound}
		I(\mathbf{x}; \mathbf{z}) \ge h(\mathbf{x}) - \frac{dT}{2} \ln(2 \pi \sigma^{2}) - \frac{dT}{2\sigma^{2}} \mathrm{MSE}(\phi, \theta).
	\end{equation}
	
	\paragraph{Analytical maximization over decoding variance.} Because the variational bound holds for any choice of variance $\sigma^{2} > 0$, we find the tightest lower bound by differentiating the right-hand side of eq.~\eqref{eq:miGaussianBound} with respect to $\sigma^{2}$ and setting the derivative to zero:
	\begin{equation}
		\frac{\partial}{\partial \sigma^{2}}\left[ -\frac{dT}{2} \ln(2 \pi \sigma^{2}) - \frac{dT}{2\sigma^{2}} \mathrm{MSE}(\phi, \theta) \right] = -\frac{dT}{2\sigma^{2}} + \frac{dT}{2(\sigma^2)^2} \mathrm{MSE}(\phi, \theta) = 0.
	\end{equation}
	Solving for the optimal decoding variance yields:
	\begin{equation}
		\sigma^{*2} = \mathrm{MSE}(\phi, \theta).
	\end{equation}
	Substituting $\sigma^{*2}$ back into eq.~\eqref{eq:miGaussianBound}:
	\begin{align}
		I(\mathbf{x}; \mathbf{z}) &\ge h(\mathbf{x}) - \frac{dT}{2} \ln\left( 2 \pi \mathrm{MSE}(\phi, \theta) \right) - \frac{dT}{2} \nonumber \\
		&= h(\mathbf{x}) - \frac{dT}{2} \left[ \ln\left( 2 \pi \mathrm{MSE}(\phi, \theta) \right) + 1 \right] \nonumber \\
		&= h(\mathbf{x}) - \frac{dT}{2} \ln\left( 2 \pi e \mathrm{MSE}(\phi, \theta) \right).
	\end{align}
	Because $-\ln(2\pi e\mathrm{MSE})$ is a strictly monotonically decreasing function of $\mathrm{MSE}(\phi, \theta)$, minimizing the decoder mean squared reconstruction error directly maximizes the lower bound on trajectory mutual information between environmental stimuli and emitted spike trains.
	
	\subsection{Theoretical considerations for homeostasis and metabolic cost}\label{section:homeostasisTheory}
	
	The KL-divergence is a critical component of the overall ELBO loss, yet its biological origins are less obvious. Indeed, the imposed regularization is \emph{intuitively} a metabolic cost, in the sense that it penalizes excess divergence from an ``ideal'' homeostatic baseline; to this end, it has been explored rather rigorously by \citet{vafaiiMetabolicCostInformation2026}. Here, we elucidate an exact derivation of the KL-divergence functional form between a baseline firing rate $\lambda_{\text{ref}}$ and a sustained activity rate $\lambda$ from biological first principles. Remarkably, the Poisson KL-divergence over firing intensities emerges directly from the birth--death kinetics of homeostatic mRNA transcription without requiring any prior assumptions on neural spike-train statistics or model architecture. We further make use of stochastic thermodynamics to directly achieve a lower bound on the chemical energy expenditure required for maintaining a specific firing rate $\lambda$ away from a prescribed baseline rate $\lambda_{\text{ref}}$.
	
	\subsubsection{Deriving a chemical master equation for homeostasis}
	
	One of the main, at least outward-facing, goals of neural homeostasis is to maintain an average prescribed firing rate. Aside from metabolic reasons, the neural cell is incentivized to continuously adjust its internals so as to maintain an optimal regime for information processing (i.e., if a sensory neuron is thrust into an environment saturated with stimulus, it should reduce its gain to retain its dynamic range; vice versa holds in the opposite scenario). In general, there are multiple forms of neural homeostasis operating in tandem and across timescales \cite{turrigianoHomeostaticSynapticPlasticity2012, marderVariabilityCompensationHomeostasis2006, olearyCellTypesNetwork2014, mozolewskiRoleNuclearCa22021, toyoizumiGeneralizedBienenstockCooper2005}. In all cases, however, the neuron needs some signal that allows it to track its own average spiking activity over time. Overwhelmingly, this is well-attributed to intracellular calcium, which serves to track, or proxy, the firing rate over time \cite{olearyCellTypesNetwork2014}. A simple model is usually given by hybrid dynamics, where the action of spiking triggers a certain influx of calcium ions $\text{Ca}^{2+}$ into the cell according to:
	\begin{equation}
		[\text{Ca}^{2+}] \rightarrow [\text{Ca}^{2+}] + \eta,
	\end{equation}
	where $\eta$ represents the increase in calcium concentration from the given influx of $\text{Ca}^{2+}$ ions. During quiescent intervals, excess calcium is pumped out of the cell, leading the concentration to leak at a rate $\alpha$:
	\begin{equation}
		\frac{\mathrm{d}[\text{Ca}^{2+}]}{\mathrm{d}t} = -\alpha [\text{Ca}^{2+}].
	\end{equation}
	The problem of homeostasis is then to minimize the difference between the tracked average firing rate and the idealized (homeostatic) firing rate, which corresponds to its own baseline calcium level $[\text{Ca}^{2+}]_{\text{base}}$:
	\begin{equation}\label{eq:homeostaticTarget}
		\frac{1}{T} \int_{0}^{T} \left([\text{Ca}^{2+}] - [\text{Ca}^{2+}]_{\text{base}}\right)\,\mathrm{d}t = 0.
	\end{equation}
	
	It is unnecessary to fixate on the exact formulations presented as, taken abstractly, the fundamental principles remain the same: the calcium concentration is to be roughly proportional to the firing rate at a given moment $\lambda(t)$, and the firing rate in turn contributes to the influx of calcium ions affecting the concentration; an external mechanism which monitors $[\text{Ca}^{2+}]$ then subsequently modulates the firing rate via exerting some influence. Such a coupled system can now be generalized:
	\begin{align}
		\frac{\mathrm{d}[\text{Ca}^{2+}]}{\mathrm{d}t} &= -\alpha [\text{Ca}^{2+}] + \eta \lambda(m, t), \label{eq:calciumDynamics} \\
		\frac{\mathrm{d}m}{\mathrm{d}t} &= -\gamma m + \kappa [\text{Ca}^{2+}], \label{eq:mRNADynamics}
	\end{align}
	where eq.~\eqref{eq:mRNADynamics} is an external variable tracking the calcium concentration with time constant $\tau_{m} \equiv \gamma^{-1}$ (similarly we have $\tau_{c} \equiv \alpha^{-1}$ for eq.~\eqref{eq:calciumDynamics}) and $\lambda(m, t)$ is taken to be the spiking rate of the neuron as a function of both time and the modulator. We assume that $\lambda(m, t)$ is monotonically influenced in a manner inversely proportional to $m$. 
	
	In this case, the constant rate parameter $\kappa$ represents the rate of mRNA transcription per unit calcium concentration $[\text{Ca}^{2+}]$, and $\gamma$ denotes the transcript degradation rate. The underlying chemical processes can be expressed as:
	\begin{align}
		[\text{Ca}^{2+}] &\xlongrightarrow{\mathmakebox[3.5cm]{\alpha}} \varnothing \\
		\varnothing &\xlongrightarrow{\mathmakebox[3.5cm]{\eta \lambda(m, t)}} [\text{Ca}^{2+}] \\
		m &\xlongrightarrow{\mathmakebox[3.5cm]{\gamma}} \varnothing \\
		\varnothing &\xlongrightarrow{\mathmakebox[3.5cm]{\kappa [\text{Ca}^{2+}]}} m
	\end{align}
	Since $m \in \mathbb{N}_{0}$ counts physical mRNA molecules, the system forms a continuous-time discrete Markov chain with calcium-dependent birth rates. 
	
	\begin{figure}[htbp]
		\centering
		\resizebox{\linewidth}{!}{%
			\begin{tikzpicture}[
				node distance=2.2cm,
				>=stealth,
				thick,
				state/.style={inner sep=2pt},
				every node/.style={font=\normalsize}
				]
				
				\node[state] (s0) {$0$};
				\node[state] (s1) [right of=s0] {$1$};
				\node[state] (s2) [right of=s1] {$2$};
				\node[state] (dots1) [right of=s2] {$\dots$};
				\node[state] (sm) [right of=dots1] {$m$};
				\node[state] (sm1) [right of=sm] {$m + 1$};
				\node[state] (dots2) [right of=sm1] {$\dots$};
				
				\draw[->, transform canvas={yshift=3mm}] (s0) -- node[above] {$\kappa [\text{Ca}^{2+}]$} (s1);
				\draw[->, transform canvas={yshift=3mm}] (s1) -- node[above] {$\kappa [\text{Ca}^{2+}]$} (s2);
				\draw[->, transform canvas={yshift=3mm}] (s2) -- node[above] {$\kappa [\text{Ca}^{2+}]$} (dots1);
				\draw[->, transform canvas={yshift=3mm}] (dots1) -- node[above] {$\kappa [\text{Ca}^{2+}]$} (sm);
				\draw[->, transform canvas={yshift=3mm}] (sm) -- node[above] {$\kappa [\text{Ca}^{2+}]$} (sm1);
				\draw[->, transform canvas={yshift=3mm}] (sm1) -- node[above] {$\kappa [\text{Ca}^{2+}]$} (dots2);
				
				\draw[->, transform canvas={yshift=-3mm}] (s1) -- node[below] {$\gamma$} (s0);
				\draw[->, transform canvas={yshift=-3mm}] (s2) -- node[below] {$2 \gamma$} (s1);
				\draw[->, transform canvas={yshift=-3mm}] (dots1) -- node[below] {$3 \gamma$} (s2);
				\draw[->, transform canvas={yshift=-3mm}] (sm) -- node[below] {$m \gamma$} (dots1);
				\draw[->, transform canvas={yshift=-3mm}] (sm1) -- node[below] {$(m + 1)\gamma$} (sm);
				\draw[->, transform canvas={yshift=-3mm}] (dots2) -- node[below] {$(m + 2)\gamma$} (sm1);
			\end{tikzpicture}%
		}
		\caption{Markov chain state transition diagram for the mRNA count $m$.}
		\label{fig:chemicalMarkovChain}
	\end{figure}
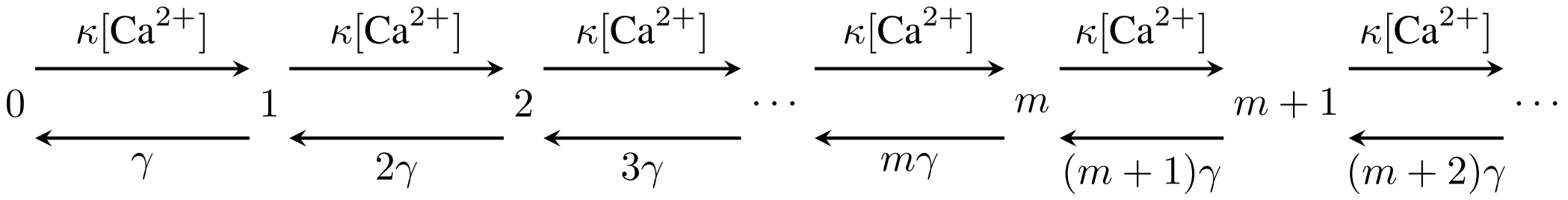
	
	Setting the continuous calcium concentration as $c \equiv [\text{Ca}^{2+}]$, the time evolution of the joint probability density $P(m, c, t)$ is governed by the continuity equation on the hybrid state space:
	\begin{equation}
		\frac{\partial P(m, c, t)}{\partial t} = - \nabla J.
	\end{equation}
	The total probability flux $J$ can be split into discrete (jumps in mRNA count) and continuous (changes in calcium concentration) parts, so that:
	\begin{equation}
		\frac{\partial P(m, c, t)}{\partial t} = -\frac{\partial J_{\text{cont}}(m, c, t)}{\partial c} + J_{\text{jump}}(m, c, t), \label{eq:continuityEquation}
	\end{equation}
	where $J_{\text{cont}}(m, c, t)$ represents the probability current carried along by continuous calcium dynamics at state $m$, and $J_{\text{jump}}(m, c, t)$ accounts for the net discrete probability transfer into state $m$ due to mRNA transcription and degradation events.
	
	The continuous probability flux $J_{\text{cont}}(m, c, t)$ is carried entirely by the deterministic calcium velocity field
	\begin{equation}\label{eq:calciumVelocityField}
		v(c, m, t) \equiv \frac{\mathrm{d}c}{\mathrm{d}t} = -\alpha c + \eta \lambda(m, t),
	\end{equation}
	yielding:
	\begin{equation}
		J_{\text{cont}}(m, c, t) = v(c, m, t) \, P(m, c, t). \label{eq:fluxContinuous}
	\end{equation}
	For discrete transitions in state $m$, we see by referencing the state-transition structure in Figure~\ref{fig:chemicalMarkovChain} that transcription occurs with rate $\kappa c$ and degrades with rate $(m + 1)\gamma$. The probability flux between states $m$ and $m + 1$ is hence
	\begin{equation}
		J_{m + 1}^{\text{jump}}(c, t) = \kappa c \, P(m, c, t) - (m + 1)\gamma \, P(m + 1, c, t). \label{eq:fluxJump}
	\end{equation}
	The net discrete jump flux entering state $m$ is given by the discrete difference:
	\begin{equation}
		J_{\text{jump}}(m, c, t) = J_{m}^{\text{jump}}(c, t) - J_{m + 1}^{\text{jump}}(c, t). \label{eq:totalFluxJump}
	\end{equation}
	Substituting eqs.~\eqref{eq:fluxContinuous}, \eqref{eq:fluxJump}, and \eqref{eq:totalFluxJump} into eq.~\eqref{eq:continuityEquation} yields the full chemical master equation (CME):
	\begin{multline}
		\frac{\partial P(m, c, t)}{\partial t} = -\frac{\partial}{\partial c}\left[ v(c, m, t) P(m, c, t) \right] + \kappa c \, P(m - 1, c, t) \\
		+ (m + 1)\gamma P(m + 1, c, t) - \left( \kappa c + m\gamma \right) P(m, c, t), \label{eq:chemicalMasterEquation}
	\end{multline}
	for $m \ge 0$, subject to the boundary condition $P(-1, c, t) \equiv 0$ and normalizability $P(m, \infty, t) = 0$.
	
	The marginal distribution of $m$, defined as $P(m, t) \equiv \int_{0}^{\infty} P(m, c, t) \, \mathrm{d}c$, is obtained by integrating eq.~\eqref{eq:chemicalMasterEquation} over all realizations $c \in [0, \infty)$:
	\begin{multline}
		\int_{0}^{\infty} \frac{\partial P(m, c, t)}{\partial t}\,\mathrm{d}c = \int_{0}^{\infty} -\frac{\partial}{\partial c}\left[ v(c, m, t) P(m, c, t) \right] \mathrm{d}c \\
		+ \int_{0}^{\infty} \Big( \kappa c \, P(m - 1, c, t) + (m + 1)\gamma P(m + 1, c, t) - \left( \kappa c + m\gamma \right) P(m, c, t) \Big) \mathrm{d}c.
	\end{multline}
	
	Starting with the advective term on the right-hand side, the spatial derivative integrated over $c$ evaluates directly via the Fundamental Theorem of Calculus:
	\begin{align}
		\int_{0}^{\infty} -\frac{\partial}{\partial c}\left[ v(c, m, t) P(m, c, t) \right] \mathrm{d}c 
		&= -\left[ v(c, m, t) P(m, c, t) \right]_{c = 0}^{c = \infty} \\
		&= -\lim_{c \to \infty} \left( v(c, m, t) P(m, c, t) \right) + v(0, m, t) P(m, 0, t) \\
		&= 0,
	\end{align}
	where the boundary terms vanish due to probability normalization at infinity ($P(m, \infty, t) = 0$) and at zero calcium ($P(m, 0, t) = 0$) on account of the inflow boundary $v(0, m, t) \ge 0$. Next, applying Leibniz's integral rule to the left-hand side converts the partial time derivative into a total time derivative:
	\begin{equation}
		\int_{0}^{\infty} \frac{\partial P(m, c, t)}{\partial t}\,\mathrm{d}c = \frac{\mathrm{d}}{\mathrm{d}t} \int_{0}^{\infty} P(m, c, t)\,\mathrm{d}c = \frac{\mathrm{d}P(m, t)}{\mathrm{d}t},
	\end{equation}
	so we are left with:
	\begin{multline}
		\frac{\mathrm{d}P(m, t)}{\mathrm{d}t} = \int_{0}^{\infty} \Big( \kappa c \, P(m - 1, c, t) \\ + (m + 1)\gamma P(m + 1, c, t) - \left( \kappa c + m\gamma \right) P(m, c, t) \Big) \mathrm{d}c.
	\end{multline}
	The degradation terms evaluate immediately:
	\begin{equation}
		\int_{0}^{\infty} (m + 1) \gamma P(m + 1, c, t)\,\mathrm{d}c = (m + 1) \gamma P(m + 1, t),
	\end{equation}
	and
	\begin{equation}
		-\int_{0}^{\infty} m \gamma P(m, c, t)\,\mathrm{d}c = -m \gamma P(m, t).
	\end{equation}
	Because the transcription rate $\kappa$ is a constant parameter, it factors directly out of the integral. Conditioning on state $m$ via $P(m, c, t) = P(c \mid m, t) P(m, t)$ yields:
	\begin{align}
		\int_{0}^{\infty} \kappa c \, P(m, c, t)\,\mathrm{d}c &= \kappa P(m, t) \int_{0}^{\infty} c \, P(c \mid m, t)\,\mathrm{d}c \\
		&= \kappa \, \mathbb{E}[c \mid m, t] P(m, t).
	\end{align}
	Noticing that, physiologically, calcium operates at a much quicker scale than homeostatic mRNA dynamics ($\tau_{c} \ll \tau_{m}$), we can consider a calcium variable to quickly settle in response to any given value of $m$. Consequently, we may perform an adiabatic elimination:
	\begin{equation}
		\mathbb{E}[c \mid m, t] \approx \mathbb{E}[c \mid m],
	\end{equation}
	and similarly for state $m - 1$:
	\begin{align}
		\int_{0}^{\infty} \kappa c \, P(m - 1, c, t)\,\mathrm{d}c &= \kappa P(m - 1, t) \int_{0}^{\infty} c \, P(c \mid m - 1, t)\,\mathrm{d}c \\
		&= \kappa \, \mathbb{E}[c \mid m - 1, t] P(m - 1, t) \\
		&\approx \kappa \, \mathbb{E}[c \mid m - 1] P(m - 1, t).
	\end{align}
	Defining the effective state-dependent transcription rate as $\kappa^*(m) \equiv \kappa \, \mathbb{E}[c \mid m]$ brings the marginal distribution to:
	\begin{equation}\label{eq:stateDependentCME}
		\frac{\mathrm{d}P(m, t)}{\mathrm{d}t} = \kappa^{*}(m - 1) P(m - 1, t) + (m + 1)\gamma P(m + 1, t) - \left( \kappa^{*}(m) + m\gamma \right) P(m, t).
	\end{equation}
	In general, eq.~\eqref{eq:stateDependentCME} does not admit a Poissonian stationary distribution on account of its state-dependent transcription rates $\kappa^*(m)$. However, homeostatic biophysical processes in real neurons do not depend on the singular mRNA microstate $m$ at every instant in time; instead, pools of mRNA contribute to inducing downstream protein production, which acts as the actual modulator of the cell's internals. Specifically, mRNA directs channel conductance through proteins $p$, whose accumulation is governed by:
	\begin{equation}
		\frac{\mathrm{d}p}{\mathrm{d}t} = -\iota p + \zeta m.
	\end{equation}
	Proteins operate at a much longer timescale than both calcium and mRNA dynamics, usually on the order of days against hours and seconds, respectively, meaning $\tau_{m} \ll \tau_{p}$ (taking $\tau_{p} \equiv \iota^{-1}$). Consequently, protein levels (for $p(0) = 0$) act as a continuous low-pass filter on $m$:
	\begin{equation}
		p(t) = \zeta \int_{0}^{t} e^{-\iota(t - s)} m(s)\,\mathrm{d}s.
	\end{equation}
	We update eq.~\eqref{eq:calciumDynamics} to reflect this architecture by letting the firing rate be modulated by the slow protein mediator, so that $\lambda(p, t)$ replaces $\lambda(m, t)$. Conditioned on the slow variable $p$, calcium reaches a steady-state expectation $c^{*}(p) \equiv \mathbb{E}[c \mid p]$. Thus, the direct microstate dependency on $m$ is removed, giving an effective transcription rate $\kappa^{*}(p) = \kappa c^{*}(p)$, and the CME reduces to:
	\begin{equation}\label{eq:stateFreeCME}
		\frac{\mathrm{d}P(m, t)}{\mathrm{d}t} = \kappa^{*}(p) P(m - 1, t) + (m + 1)\gamma P(m + 1, t) - \left( \kappa^{*}(p) + m\gamma \right) P(m, t).
	\end{equation}
	Equation~\eqref{eq:stateFreeCME} is an $M/M/\infty$ queue with birth rate $\kappa^{*}(p)$ and individual death rate $\gamma$, which possesses the known stationary solution given precisely by a Poisson distribution:
	\begin{equation}
		P(m) = \frac{1}{m!} \left( \frac{\kappa^{*}(p)}{\gamma} \right)^{m} \exp\left( -\frac{\kappa^{*}(p)}{\gamma} \right),
	\end{equation}
	with expected steady-state transcript count $\mathbb{E}[m] = \frac{\kappa^{*}(p)}{\gamma}$ and variance $\operatorname{Var}(m) = \frac{\kappa^{*}(p)}{\gamma}$. Moreover, for any process initialized at state $m(0) = 0$, the probability distribution at any instant $t$ remains Poissonian with time-varying mean $\mu(t)$:
	\begin{equation}
		P(m, t) = \frac{1}{m!} \left[ \mu(t) \right]^{m} e^{-\mu(t)}, \quad \text{where} \quad \mu(t) = \frac{\kappa^{*}(p)}{\gamma} \left( 1 - e^{-\gamma t} \right).
	\end{equation}
	
	\subsubsection{Determining the baseline homeostatic distribution}
	
	The autonomous homeostatic fixed point $(c^{*}, m^{*}, p^{*})$, where $c^{*} \equiv [\text{Ca}^{2+}]_{\text{base}}$, is determined by the intersection of the nullclines:
	\begin{align}
		0 &= -\alpha c^{*} + \eta \lambda(p^{*}), \label{eq:calciumNullcline} \\
		0 &= -\gamma m^{*} + \kappa c^{*}, \label{eq:mRNANullcline} \\
		0 &= -\iota p^{*} + \zeta m^{*}. \label{eq:proteinNullcline}
	\end{align}
	Solving eq.~\eqref{eq:calciumNullcline} gives the baseline calcium concentration:
	\begin{equation}\label{baselineCalcium}
		c^{*} = [\text{Ca}^{2+}]_{\text{base}} = \frac{\eta}{\alpha} \lambda(p^{*}).
	\end{equation}
	Substituting $c^{*}$ into eq.~\eqref{eq:mRNANullcline} yields the steady-state mRNA transcript expectation:
	\begin{equation}\label{baselineMRNA}
		m^{*} = \frac{\kappa c^{*}}{\gamma} = \frac{\kappa \eta}{\gamma \alpha} \lambda(p^{*}).
	\end{equation}
	From eq.~\eqref{eq:proteinNullcline}, the steady-state protein mediator level is:
	\begin{equation}\label{baselineProtein}
		p^{*} = \frac{\zeta}{\iota} m^{*} = \frac{\zeta \kappa \eta}{\iota \gamma \alpha} \lambda(p^{*}).
	\end{equation}
	We assume that evolution has tuned the biophysical parameters $\alpha$, $\gamma$, $\iota$, $\zeta$, and $\kappa$ to produce the appropriate point $(c^{*}, m^{*}, p^{*})$. Defining the resting reference firing rate as $\lambda^{*}_{\text{ref}} \equiv \lambda(p^{*})$:
	\begin{equation}\label{eq:baselineFiringRate}
		\lambda^*_{\text{ref}} = \frac{\alpha c^{*}}{\eta}.
	\end{equation}
	The baseline transcription rate is therefore:
	\begin{equation}\label{eq:baselineTranscriptionRate}
		\kappa^{*}_{\text{ref}} = \kappa c^{*} = \frac{\kappa \eta}{\alpha} \lambda^{*}_{\text{ref}},
	\end{equation}
	and the resting transcript expectation is $\mu^{*}_{\text{ref}} = \frac{\kappa^{*}_{\text{ref}}}{\gamma} = \frac{\kappa \eta}{\gamma \alpha}\lambda^{*}_{\text{ref}}$. This yields the resting homeostatic transcript distribution:
	\begin{equation}\label{eq:poissonRateDistribution}
		P(m \mid \lambda^{*}_{\text{ref}}) = \frac{1}{m!} \left( \frac{\kappa \eta}{\gamma \alpha} \lambda^{*}_{\text{ref}} \right)^{m} \exp\left( -\frac{\kappa \eta}{\gamma \alpha} \lambda^{*}_{\text{ref}} \right).
	\end{equation}
	
	When the neuron is driven to a sustained target firing rate $\lambda^{*} \neq \lambda^{*}_{\text{ref}}$, the transcript pool shifts to a driven Poisson distribution $P(m \mid \lambda^{*})$ with mean $\mu^{*} = \frac{\kappa \eta}{\gamma \alpha} \lambda^{*}$. The relative entropy between the driven state and the resting homeostatic baseline evaluates in closed form to:
	\begin{align}
		D_{\text{KL}}\!\left(P(m \mid \lambda^{*}) \,\|\, P(m \mid \lambda^{*}_{\text{ref}})\right) 
		&= \sum_{m = 0}^{\infty} P(m \mid \lambda^{*}) \ln\left( \frac{P(m \mid \lambda^{*})}{P(m \mid \lambda^{*}_{\text{ref}})} \right) \nonumber \\
		&= \mu^{*} \ln\left( \frac{\mu^{*}}{\mu^{*}_{\text{ref}}} \right) - (\mu^{*} - \mu^{*}_{\text{ref}}) \nonumber \\
		&= \frac{\kappa \eta}{\gamma \alpha} \left[ \lambda^{*} \ln\left( \frac{\lambda^{*}}{\lambda^{*}_{\text{ref}}} \right) - (\lambda^{*} - \lambda^{*}_{\text{ref}}) \right]. \label{eq:intensityDivergence}
	\end{align}
	It can also be noted that eq.~\eqref{eq:intensityDivergence} is the Bregman divergence on Poisson intensities generated by the strictly convex potential $\psi(x) = x \ln x - x$, and depends solely on the mean firing rates $\lambda^{*}$ and $\lambda^{*}_{\text{ref}}$ regardless of higher-order spike-train statistics.
	
	\subsubsection{Stochastic thermodynamics of homeostatic control}
	
	In stochastic thermodynamics, the entropy production rate $\dot{S}$ of a driven chemical reaction network can be decomposed into basal housekeeping and excess regulatory driving components:
	\begin{equation}\label{eq:entropyProductionDecomposition}
		\dot{S} = \dot{S}_{\text{hk}} + \dot{S}_{\text{ex}}.
	\end{equation}
	Here, $\dot{S}_{\text{hk}} \ge 0$ characterizes the housekeeping dissipation required to sustain the resting non-equilibrium steady-state (NESS), while $\dot{S}_{\text{ex}} \ge 0$ captures the excess dissipation incurred when externally driving and maintaining the system away from its resting baseline. Under isothermal conditions at physiological temperature $T_{\text{temp}}$, the chemical power expenditure is obtained by multiplying eq.~\eqref{eq:entropyProductionDecomposition} by $k_{B} T_{\text{temp}}$:
	\begin{equation}
		\dot{W}_{\text{tot}} = \dot{W}_{\text{hk}}^{\text{rest}} + \dot{W}_{\text{ex}},
	\end{equation}
	where $\dot{W}_{\text{hk}}^{\text{rest}} \equiv k_{B} T_{\text{temp}} \dot{S}_{\text{hk}}$ corresponds to resting housekeeping power, and $\dot{W}_{\text{ex}} \equiv k_{B} T_{\text{temp}} \dot{S}_{\text{ex}}$ represents excess regulatory power. In the context of the neuronal cell, driving an active neuron away from its homeostatic baseline to another time-varying firing rate $\lambda(t) \neq \lambda^{*}_{\text{ref}}$ necessarily requires spending excess power. Physically, this arises from the regulatory control exerted by the cell in order to counter deviations from the preferred state; with respect to our formulation, the regulatory control is the calcium-coupled transcription machinery for downstream conductance modifications.
	
	The altered activity shifts the instantaneous calcium-driven transcription initiation rate from its baseline $\kappa^{*}_{\text{ref}} = \frac{\kappa \eta}{\alpha}\lambda^{*}_{\text{ref}}$ to $\kappa^{*}(t) = \frac{\kappa \eta}{\alpha}\lambda(t)$. Because the transcript pool turns over continuously at degradation rate $\gamma$, sustaining a shifted distribution against spontaneous relaxation toward $P(m \mid \lambda^{*}_{\text{ref}})$ requires replenishing the state-space divergence of eq.~\eqref{eq:intensityDivergence} at turnover rate $\gamma$. By the generalized Second Law for non-equilibrium state maintenance \cite{espositoSecondLawLandauer2011, parrondoThermodynamicsInformation2015}, the excess non-equilibrium regulatory power $\dot{W}_{\text{ex}}(t)$ needed to sustain this driven rate is lower-bounded by the relative entropy rate $\dot{\mathcal{D}}_{\text{KL}}(t) \equiv \gamma D_{\text{KL}}\!\left(P(m \mid \lambda(t)) \,\parallel\, P(m \mid \lambda^{*}_{\text{ref}})\right)$:
	\begin{equation}\label{eq:excessPowerBound}
		\dot{W}_{\text{ex}}(t) \ge k_{B} T_{\text{temp}} \, \dot{\mathcal{D}}_{\text{KL}}(t) = k_{B} T_{\text{temp}} \, \xi \left[ \lambda(t) \ln\left( \frac{\lambda(t)}{\lambda^{*}_{\text{ref}}} \right) - \left( \lambda(t) - \lambda^{*}_{\text{ref}} \right) \right],
	\end{equation}
	where $\xi \equiv \frac{\kappa \eta}{\alpha}$ is the dimensionless biophysical coupling constant and 
	\begin{align}
		\dot{\mathcal{D}}_{\text{KL}}(t) 
		&= \kappa^{*}(t) \ln\left( \frac{\kappa^*(t)}{\kappa^{*}_{\text{ref}}} \right) - \left( \kappa^{*}(t) - \kappa^{*}_{\text{ref}} \right) \nonumber \\
		&= \frac{\kappa \eta}{\alpha} \left[ \lambda(t) \ln\left( \frac{\lambda(t)}{\lambda^{*}_{\text{ref}}} \right) - \left( \lambda(t) - \lambda^{*}_{\text{ref}} \right) \right].
	\end{align}
	At the homeostatic baseline $\lambda(t) = \lambda^{*}_{\text{ref}}$, the excess regulatory work vanishes, leaving $\dot{W}_{\text{tot}}(t) = \dot{W}_{\text{hk}}^{\text{rest}}$. Yet, even at rest ($\lambda^{*} = \lambda^{*}_{\text{ref}}$) where $\dot{W}_{\text{ex}} = 0$, the neuron must continuously expend basal housekeeping metabolic power $\dot{W}_{\text{hk}}^{\text{rest}}$ to maintain its NESS against passive calcium leak. At homeostasis with $\lambda(p^{*}, t) = \lambda^{*}_{\text{ref}}$, calcium influx and efflux achieve balance:
	\begin{equation}
		c^{*} = \frac{\eta}{\alpha} \lambda^{*}_{\text{ref}} \implies \alpha c^{*} = \eta \lambda^{*}_{\text{ref}},
	\end{equation}
	meaning that calcium ions entering the cell at rate $\eta \lambda^{*}_{\text{ref}}$ must be actively pumped out (predominantly via the plasma membrane $\text{Ca}^{2+}$-ATPase [PMCA] and $\text{Na}^+/\text{Ca}^{2+}$ exchangers \cite{olearyCellTypesNetwork2014, mozolewskiRoleNuclearCa22021}) against their electrochemical gradient at rate $\alpha c^{*}$. Absorbing the cytosolic volume into $c^{*}$ and $\eta$, the resting housekeeping power dissipation is proportional to the Gibbs free energy $\Delta G$ of active calcium transport:
	\begin{equation}\label{eq:housekeepingWork}
		\dot{W}_{\text{hk}}^{\text{rest}} = \Delta G \, \alpha c^{*} = \Delta G \, \eta \lambda^{*}_{\text{ref}}.
	\end{equation}
	Combining basal housekeeping and excess homeostatic control work (alongside any linear activity-dependent ion extrusion $\Delta G \, \eta (\lambda(t) - \lambda^*_{\text{ref}})$), the homeostatic regulatory power satisfies:
	\begin{equation}
		\dot{W}_{\text{tot}}(t) = \dot{W}_{\text{hk}}^{\text{rest}} + \dot{W}_{\text{ex}}(t) \ge \Delta G \, \eta \lambda^{*}_{\text{ref}} + k_B T_{\text{temp}} \, \xi \left[ \lambda(t) \ln\left( \frac{\lambda(t)}{\lambda^{*}_{\text{ref}}} \right) - \left( \lambda(t) - \lambda^{*}_{\text{ref}} \right) \right].
	\end{equation}
	Expectedly, the true housekeeping of a neuron incorporates far more metabolic processes than just calcium expulsion \cite{laughlinMetabolicCostNeural1998, hasenstaubMetabolicCostUnifying2010}, but for the purposes of this formulation, we consider eq.~\eqref{eq:housekeepingWork} to be a catch-all basal term.
	
	\paragraph{Thermodynamic grounding of the Poisson ELBO regularizer.}
	This stochastic thermodynamic formulation resolves the physical origin of the latent regularization penalty in the recurrent Poisson Variational Autoencoder. In our model, the neural population emits latent spike counts $\mathbf{z}_{t} \in \mathbb{N}_{0}^{n}$ at each discrete simulation step $t \in \{1, \dots, T\}$ corresponding to time bins of width $\Delta t$. The approximate posterior is parametrized directly by the dynamic bin-integrated rate vector $q_{\phi}(\mathbf{z}_{t} \mid \mathbf{x}_{\le t}, \mathbf{z}_{<t}) = \operatorname{Poisson}(\mathbf{z}_{t}; \boldsymbol{\lambda}_{t})$, while the resting homeostatic prior is given by $p(\mathbf{z}_{t}) = \operatorname{Poisson}(\mathbf{z}_{t}; \boldsymbol{\lambda}_{\text{ref}})$.
	
	Evaluating the step-wise KL-divergence between these two multivariate Poisson distributions yields:
	\begin{equation}\label{eq:directPoissonKL}
		D_{\text{KL}}\!\left(q_{\phi}(\mathbf{z}_{t} \mid \mathbf{x}_{\le t}, \mathbf{z}_{<t}) \,\|\, p(\mathbf{z}_{t})\right) = \sum_{i = 1}^{n} \left[ \lambda_{t, i} \ln\left(\frac{\lambda_{t, i}}{\lambda_{\text{ref}, i}}\right) - (\lambda_{t, i} - \lambda_{\text{ref}, i}) \right].
	\end{equation}
	Because the discrete simulation variables represent bin-integrated expectations ($\boldsymbol{\lambda}_t = \boldsymbol{\lambda}(t)\Delta t$ for continuous physical firing rate $\boldsymbol{\lambda}(t)$ in $\mathrm{Hz}$), the continuous trajectory entropy rate is $\dot{\mathcal{D}}_{\text{KL}}(t) = \frac{\xi}{\Delta t} D_{\text{KL}}\!\left(q_{\phi}(\mathbf{z}_{t} \mid \mathbf{x}_{\le t}, \mathbf{z}_{<t}) \,\|\, p(\mathbf{z}_{t})\right)$. Substituting this into eq.~\eqref{eq:excessPowerBound}, the instantaneous excess metabolic power required to drive the population away from resting homeostasis satisfies:
	\begin{equation}\label{eq:instantaneousPowerBound}
		\dot{W}_{\text{ex}}(t) \ge \frac{k_{B} T_{\text{temp}} \xi}{\Delta t} \, D_{\text{KL}}\!\left(q_{\phi}(\mathbf{z}_{t} \mid \mathbf{x}_{\le t}, \mathbf{z}_{<t}) \,\|\, p(\mathbf{z}_{t})\right) \propto \beta_{\text{KL}} D_{\text{KL}}\!\left(q_{\phi}(\mathbf{z}_{t} \mid \mathbf{x}_{\le t}, \mathbf{z}_{<t}) \,\|\, p(\mathbf{z}_{t})\right).
	\end{equation}
	Integrating this dissipation rate over an observation sequence of $T$ time steps (total duration $\mathcal{T} = T \Delta t$), the time increment $\Delta t$ cancels, yielding the cumulative excess thermodynamic work $\Delta W_{\text{ex}}$ expended by cellular machinery above basal housekeeping:
	\begin{align}\label{eq:cumulativeWorkBound}
		\Delta W_{\text{ex}} \equiv \int_{0}^{\mathcal{T}} \dot{W}_{\text{ex}}(t)\,\mathrm{d}t 
		&\ge \sum_{t = 1}^{T} \left( \frac{k_{B} T_{\text{temp}} \xi}{\Delta t} \, D_{\text{KL}}\!\left(q_{\phi}(\mathbf{z}_{t} \mid \mathbf{x}_{\le t}, \mathbf{z}_{<t}) \,\|\, p(\mathbf{z}_{t})\right) \right) \Delta t \\
		&= k_{B} T_{\text{temp}} \, \xi \sum_{t = 1}^{T} D_{\text{KL}}\!\left(q_{\phi}(\mathbf{z}_{t} \mid \mathbf{x}_{\le t}, \mathbf{z}_{<t}) \,\|\, p(\mathbf{z}_{t})\right).
	\end{align}
	Subsequently taking $\beta_{\text{KL}} \propto k_{B} T_{\text{temp}} \, \xi$ leads to
	\begin{equation}
		\Delta W_{\text{ex}} \propto \beta_{\text{KL}} \sum_{t = 1}^{T} D_{\text{KL}}\!\left(q_{\phi}(\mathbf{z}_{t} \mid \mathbf{x}_{\le t}, \mathbf{z}_{<t}) \,\|\, p(\mathbf{z}_{t})\right).
	\end{equation}
	Indeed, this is precisely equivalent to the total sequence-level KL-divergence penalty portion of the ELBO objective:
	\begin{equation}
		\mathcal{L}_{\text{ELBO}} = \sum_{t = 1}^{T} \mathbb{E}_{q_{\phi}}\!\left[\ln p_{\theta}(\mathbf{x}_{t} \mid \mathbf{z}_{\le t})\right] - \beta_{\text{KL}} \sum_{t = 1}^{T} \mathbb{E}_{q_{\phi}}\!\left[D_{\text{KL}}\!\left(q_{\phi}(\mathbf{z}_{t} \mid \mathbf{x}_{\le t}, \mathbf{z}_{<t}) \,\|\, p(\mathbf{z}_{t})\right)\right].
	\end{equation}
	Additionally, due to this correspondence, we can consider tuning $\beta_{\text{KL}}$ as setting the energetic price of non-equilibrium driving. Therefore, the Poisson ELBO objective provides a rigorous mathematical realization of the fundamental biological \emph{energy--accuracy trade-off} in sensory transduction \cite{lanEnergySpeedAccuracy2012}, balancing the mutual information of the sensory representation against the physical chemical work required to sustain it.
	
	\clearpage
	
	\section{Supplementary Results}
	
	\begin{figure}
		\centering
		\includegraphics[width=\textwidth]{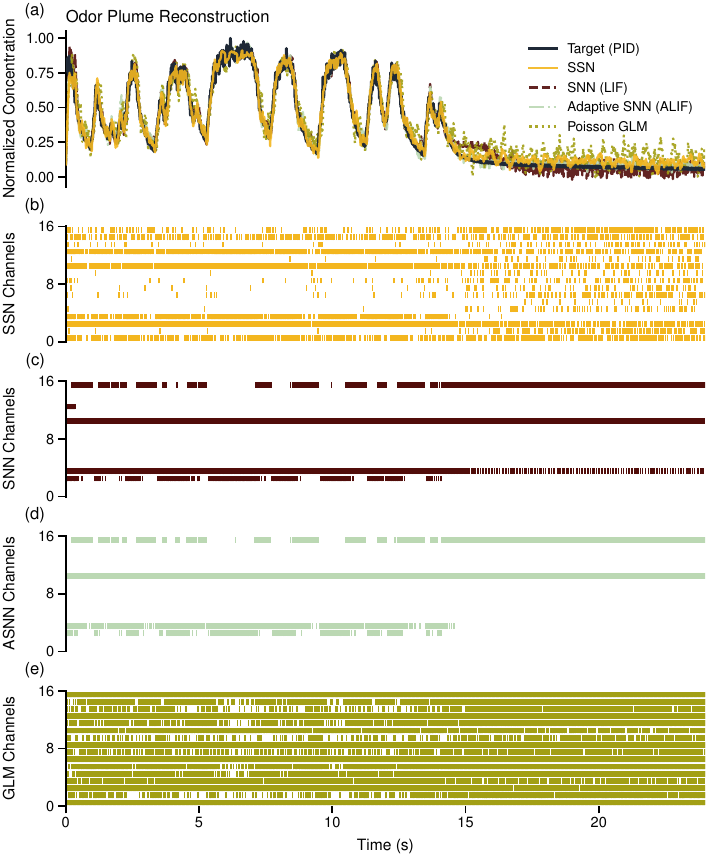}
		\caption{Full spike raster visualization for the odor plume reconstruction task. \textbf{(a)} Example reconstructions produced by each individual model. \textbf{(b)} Spike raster of SSN channel firing. \textbf{(c)} Spike raster of SNN model firing. Notably, it is overwhelmingly sparse and artificial. \textbf{(d)} Spike raster of ASNN channel firing. In similar fashion to the SNN, the ASNN also produces highly sparse, artificial looking activity. \textbf{(e)} Spike raster of GLM model firing. Unlike all other models presented, its firing activity is extraordinarily dense, far beyond a realistic degree.}
		\label{fig:plume-reconstruction-rasters}
	\end{figure}
	
\end{document}